\documentclass[acmsmall, nonacm, screen]{acmart}
\usepackage{subcaption}

\usepackage{comment}
\usepackage{xspace}
\usepackage{graphicx}
\usepackage{adjustbox}
\usepackage{enumitem}
\usepackage{multirow}
\usepackage{amsmath,bm,amsfonts}
\usepackage[edges]{forest}
\usepackage{bbding}

\usepackage[misc,geometry]{ifsym}
\usepackage{array}
\usepackage{tabularx}
\usepackage{makecell}
\usepackage{bm}
\usepackage{array} 
\usepackage{xcolor} 

\usetikzlibrary{arrows.meta,calc,decorations.markings,math}

\newcommand{\ie}{\emph{i.e.,}\xspace}

\newcommand{\eg}{\emph{e.g.,}\xspace}
\newcommand{\Eg}{}

\newcommand{\etc}{\emph{etc}}

\newcommand{\bcheck}{{\Checkmark}}
\newcommand{\hdiamond}{{\RightDiamond}}
\newcommand{\rcross}{{\XSolidBrush}}

\newcommand{\ignore}[1]{}

\definecolor{cosmiclatte}{rgb}{1.0, 0.97, 0.91}
\definecolor{sunset}{rgb}{0.98, 0.84, 0.65}
\definecolor{peach}{rgb}{1.0, 0.9, 0.71}
\definecolor{ashgrey}{rgb}{0.7, 0.75, 0.71}
\definecolor{amber}{rgb}{1.0, 0.75, 0.0}
\definecolor{darkseagreen}{rgb}{0.56, 0.74, 0.56}

\definecolor{hidden-draw}{rgb}{0.35, 0.15, 0.13}
\definecolor{hidden-blue}{RGB}{194,232,247}
\definecolor{hidden-orange}{RGB}{243,202,120}
\definecolor{hidden-yellow}{RGB}{242,244,193}
\definecolor{tree-level-1}{RGB}{245,20,85}
\definecolor{tree-level-2}{RGB}{246,86,118}
\definecolor{tree-level-3}{RGB}{248,177,193}
\definecolor{tree-leaf}{RGB}{176,230,198}

\definecolor{CBblue}{RGB}{0, 107, 164}
\definecolor{CBorange}{RGB}{255, 128, 14}
\definecolor{CBgray}{RGB}{89, 89, 89}
\definecolor{CByellow}{RGB}{171, 171, 42}
\definecolor{CBpurple}{RGB}{137, 61, 86}
\definecolor{CBgreen}{RGB}{27, 158, 119}
\definecolor{CBpink}{RGB}{215, 48, 39}

\newcommand{ \algfont}[1]{\texttt{\hyphenchar \font=\defaulthyphenchar #1}}

\newcommand{\loss}{\ell}

\newcommand{\fedavg}{\algfont{FedAvg}\xspace}

\newcommand{\missing}[1]{\textbf{\textcolor{red}{#1}}}
\newcommand{\cblue}[1]{\textcolor{CBblue}{#1}}
\newcommand{\corange}[1]{\textcolor{CBorange}{#1}}

\newcommand{\cyellow}[1]{\textcolor{CByellow}{#1}}
\newcommand{\cpurple}[1]{\textcolor{CBpurple}{#1}}
\newcommand{\cgreen}[1]{\textcolor{CBgreen}{#1}}
\newcommand{\cpink}[1]{\textcolor{CBpink}{#1}}

\usepackage{pifont}%

\newcommand{\fedfm}{\texttt{FedFM}\xspace}

\usepackage{xcolor}

\makeatletter

\renewcommand{\author}[1]{\gdef\@customauthor{#1}}

\def\@mkauthors{%
  \global\setbox\mktitle@bx=\vbox{%
    \noindent\unvbox\mktitle@bx  %
    \par\medskip                  %
    \noindent
    \centering                    %
    \@customauthor                %
    \par\bigskip                  %
  }%
}
\makeatother

\setcopyright{none}
\authorsaddresses{}
\begin{document}

\title{Synergizing Foundation Models and Federated Learning: A Survey}

\author{
    {\normalsize \textbf{Shenghui Li}$^{1}$ \hspace{2em}
    \textbf{Fanghua Ye}$^{2}$               \hspace{2em}
    \textbf{Meng Fang}$^{3}$                \hspace{2em}
    \textbf{Jiaxu Zhao}$^{4}$} \\
    {\normalsize \textbf{Yun-Hin Chan}$^{5}$\hspace{2em}
    \textbf{Edith C. H. Ngai}$^{5}$         \hspace{2em}
    \textbf{Thiemo Voigt}$^{1,6}$} \\
    {
    \normalsize $^1$ Uppsala University, Sweden. \emph{\{shenghui.li, thiemo.voigt\}@angstrom.uu.se}} \\
    {\normalsize $^2$ University College London, United Kingdom. \emph{fanghua.ye.19@ucl.ac.uk}} \\
    {\normalsize $^3$ University of Liverpool, United Kingdom. \emph{meng.fang@liverpool.ac.uk}} \\
    {\normalsize $^4$ Eindhoven University of Technology, the Netherlands. \emph{j.zhao@tue.nl}} \\
    {\normalsize $^5$ The University of Hong Kong, China. \emph{\{chngai@eee, chanyunhin@connect\}.hku.hk}} \\
    {\normalsize $^6$ Research Institutes of Sweden, Sweden. %
    } \\
    {\url{https://github.com/lishenghui/awesome-fm-fl}}\\
}

\renewcommand{\shortauthors}{S. Li et al.}

\begin{abstract}

Over the past few years, the landscape of Artificial Intelligence (AI) has been reshaped by the emergence of Foundation Models (FMs). Pre-trained on massive datasets, these models exhibit exceptional performance across diverse downstream tasks through adaptation techniques like fine-tuning and prompt learning. More recently, the synergy of FMs and Federated Learning (FL) has emerged as a promising paradigm, often termed Federated Foundation Models (\fedfm), allowing for collaborative model adaptation while preserving data privacy. This survey paper provides a systematic review of the current state of the art in \fedfm, offering insights and guidance into the evolving landscape. Specifically, we present a comprehensive multi-tiered taxonomy based on three major dimensions, namely efficiency, adaptability, and trustworthiness. To facilitate practical implementation and experimental research, we undertake a thorough review of existing libraries and benchmarks. Furthermore, we discuss the diverse real-world applications of this paradigm across multiple domains. 
Finally, we outline promising research directions to foster future advancements in \fedfm. 
Overall, this survey serves as a resource for researchers and practitioners, offering a thorough understanding of \fedfm's role in revolutionizing privacy-preserving AI and pointing toward future innovations in this promising area.

\end{abstract}


\maketitle

\section{Introduction}

A significant paradigm shift in Artificial Intelligence (AI) has been catalyzed by the rapid advancement of {Foundation Models (FMs)}~\citep{bommasani2021opportunities}, such as BERT~\cite{devlin-etal-2019-bert}, GPT series~\cite{brown2020language,openai2023chatgpt,openai2023gpt4}, and LLaMA series~\cite{touvron2023llama,touvron2023llama2} in Natural Language Processing (NLP); ViTs~\cite{dosovitskiy2021an} and SAM~\cite{kirillov2023segment} in Computer Vision (CV); CLIP~\cite{radford2021learning}, DALL-E~\cite{ramesh2021zero}, Gemini~\cite{geminiteam2023gemini}, and GPT-4o in multimodal applications. These FMs have become pivotal in a myriad of AI applications across diverse domains. The remarkable generalization capabilities of FMs are attributed to large-scale pre-training~\citep{gunasekar2023textbook}, which enables them to learn robust representations of language, vision, and multimodal data. To leverage these representations for specific applications, adaptation techniques such as fine-tuning are widely adopted to tailor FMs to diverse downstream tasks~\cite{zhao2023survey}.

While general-purpose FMs can leverage openly accessible data from the Internet, domain-specific FMs require proprietary data. It is, however, challenging to collect vast amounts of proprietary data and perform centralized pre-training or fine-tuning for domain-specific models, due to privacy restrictions~\cite{351095.3372829}, \ie~General Data Protection Regulation (GDPR)~\cite{GDPR} and California Consumer Privacy Act (CCPA)~\cite{CCPA}. Particularly in domains such as law, healthcare, and finance, where data is inherently privacy-sensitive, there is a pressing need for stringent privacy safeguards. Furthermore, given that data often constitutes a pivotal asset for enterprises, its widespread distribution is prohibitive.
Consequently, there is an urgent need for novel strategies to handle data availability and facilitate model training, thereby unlocking the potential of domain-specific applications of FMs whilst respecting data privacy.

To address the challenges of data privacy in model training, Federated Learning (FL) \citep{mcmahan2017communication} has emerged as a promising paradigm. FL facilitates collaborative model training across decentralized clients without sharing raw data, thereby preserving privacy. Specifically, FL involves periodic interactions between the server and decentralized clients to exchange trainable model parameters without requiring private client data. Recognizing such a benefit, integrating FMs with FL, often termed Federated Foundation Models (\fedfm), presents a compelling solution for domain-specific FMs~\cite{zhuang2023foundation,yu2023federated}.

\begin{table*}[tbp]
  \centering
  \setlength{\tabcolsep}{4pt}
  \renewcommand{\arraystretch}{1.25}
  \caption{Summary of previous surveys related to FedFM (\textcolor{CBgreen}{\bcheck}: Included, { \textcolor{CBorange}{\hdiamond} }: Partially included, \textcolor{CBpink}{\rcross}: Not included)}
  \label{tab:related-work}
  \resizebox{0.85\textwidth}{!}{%
    \begin{tabular}{c||c|c|ccc|c|c|c}
    \hline
    \multirow{2}[2]{*}{\textbf{Survey}} & \multirow{2}[2]{*}{\textbf{Year}} 
      & \multirow{2}[2]{*}{\textbf{Taxonomy}} 
      & \multicolumn{3}{c|}{\textbf{Technical Focus}} 
      & \multirow{2}[2]{*}{\makecell[c]{\small \textbf{Library \&} \\ \small \textbf{Benchmark}}} 
      & \multirow{2}[2]{*}{\textbf{\small Application}} 
      & \multirow{2}[2]{*}{\makecell[c]{\small \textbf{Future} \\ \small \textbf{Direction}}} \\
    \cmidrule{4-6}
      & 
      & 
      & \makecell[c]{\scriptsize \textbf{Efficiency}}& \makecell[c]{\scriptsize \textbf{Adaptability}}& \makecell[c]{\scriptsize \textbf{Trustworthiness}} 
      & 
      & 
      & \\
    \hline
    \hline
    \citet{zhuang2023foundation} & \citeyear{zhuang2023foundation} 
      & \textcolor{CBpink}{\rcross} 
      & \textcolor{CBorange}{\hdiamond}  & \textcolor{CBorange}{\hdiamond}  & \textcolor{CBpink}{\rcross} 
      & \textcolor{CBpink}{\rcross} & \textcolor{CBorange}{\hdiamond}  & \textcolor{CBgreen}{\bcheck} \\ 
    \citet{chen2023federated} & \citeyear{chen2023federated} 
      & \textcolor{CBpink}{\rcross} 
      & \textcolor{CBorange}{\hdiamond}  & \textcolor{CBorange}{\hdiamond}  & \textcolor{CBpink}{\rcross} 
      & \textcolor{CBpink}{\rcross} & \textcolor{CBorange}{\hdiamond}  & \textcolor{CBgreen}{\bcheck} \\
    \hline
    \citet{yu2023federated} & \citeyear{yu2023federated} 
      & \textcolor{CBpink}{\rcross} 
      & \textcolor{CBorange}{\hdiamond}  & \textcolor{CBorange}{\hdiamond}  & \textcolor{CBpink}{\rcross} 
      & \textcolor{CBpink}{\rcross} & \textcolor{CBpink}{\rcross} & \textcolor{CBgreen}{\bcheck} \\
    \citet{woisetschlager2024survey} & \citeyear{woisetschlager2024survey} 
      & \textcolor{CBgreen}{\bcheck} 
      & \textcolor{CBgreen}{\bcheck}& \textcolor{CBpink}{\rcross} & \textcolor{CBpink}{\rcross} 
      & \textcolor{CBpink}{\rcross} & \textcolor{CBpink}{\rcross} & \textcolor{CBorange}{\hdiamond} \\
    \citet{10733964} & \citeyear{10733964} 
      & \textcolor{CBpink}{\rcross}
      & \textcolor{CBgreen}{\bcheck} & \textcolor{CBpink}{\rcross}& \textcolor{CBpink}{\rcross} 
      & \textcolor{CBpink}{\rcross} & \textcolor{CBpink}{\rcross} & \textcolor{CBorange}{\hdiamond} \\
    \citet{li2024position} & \citeyear{li2024position} 
      & \textcolor{CBpink}{\rcross} 
      & \textcolor{CBpink}{\rcross}   & \textcolor{CBpink}{\rcross}   & \textcolor{CBgreen}{\bcheck} 
      & \textcolor{CBpink}{\rcross} & \textcolor{CBpink}{\rcross} & \textcolor{CBgreen}{\bcheck} \\
    \citet{yao2024federatedlargelanguagemodels} & \citeyear{yao2024federatedlargelanguagemodels} 
      & \textcolor{CBpink}{\rcross} 
      & \textcolor{CBorange}{\hdiamond}  & \textcolor{CBorange}{\hdiamond}  & \textcolor{CBorange}{\hdiamond}  
      & \textcolor{CBpink}{\rcross} & \textcolor{CBorange}{\hdiamond}   & \textcolor{CBorange}{\hdiamond}  \\
    \hline
    \citet{kang2024grounding} & \citeyear{kang2024grounding} 
      & \textcolor{CBorange}{\hdiamond}  
      & \textcolor{CBgreen}{\bcheck} & \textcolor{CBorange}{\hdiamond}  & \textcolor{CBorange}{\hdiamond}  
      & \textcolor{CBpink}{\rcross} & \textcolor{CBgreen}{\bcheck} & \textcolor{CBgreen}{\bcheck} \\
    \citet{wu2025surveyfederatedfinetuninglarge} & \citeyear{wu2025surveyfederatedfinetuninglarge}  & \textcolor{CBpink}{\rcross}
      & \textcolor{CBgreen}{\bcheck}  & \textcolor{CBgreen}{\bcheck}  & \textcolor{CBpink}{\rcross}
      & \textcolor{CBorange}{\hdiamond}  & \textcolor{CBorange}{\hdiamond} & \textcolor{CBorange}{\hdiamond} \\
    \citet{10944288} & \citeyear{10944288}  & \textcolor{CBpink}{\rcross} & \textcolor{CBpink}{\rcross} & \textcolor{CBorange}{\hdiamond}  & \textcolor{CBgreen}{\bcheck} & \textcolor{CBpink}{\rcross}  &\textcolor{CBpink}{\rcross} & \textcolor{CBgreen}{\bcheck} \\
    \citet{ren2024advances} & \citeyear{ren2024advances} 
      & \textcolor{CBgreen}{\bcheck} 
      & \textcolor{CBgreen}{\bcheck}& \textcolor{CBpink}{\rcross}& \textcolor{CBgreen}{\bcheck} 
      & \textcolor{CBgreen}{\bcheck}  &\textcolor{CBgreen}{\bcheck}  & \textcolor{CBgreen}{\bcheck} \\
    \hline
    \hline
    \textbf{Ours} & 2026 
      & \textcolor{CBgreen}{\bcheck} 
      & \textcolor{CBgreen}{\bcheck} & \textcolor{CBgreen}{\bcheck} & \textcolor{CBgreen}{\bcheck} 
      & \textcolor{CBgreen}{\bcheck} & \textcolor{CBgreen}{\bcheck} & \textcolor{CBgreen}{\bcheck} \\
    \hline
    \end{tabular}%
  }
\end{table*}

Driven by the surging interest in the intersection of FMs and FL, several surveys have investigated various dimensions of the \fedfm paradigm. Table~\ref{tab:related-work} compares our work with prior surveys in terms of coverage of key aspects. Most existing surveys focus on presenting the vision and challenges, but few analyze the overall system performance from multiple perspectives. Specifically, early contributions primarily articulated the motivation and outlined the key challenges of integrating FMs with FL~\cite{zhuang2023foundation,yu2023federated,chen2023federated}. Subsequent work has considered specific topics but mostly focuses on efficiency~\cite{woisetschlager2024survey,10733964,10944288,yao2024federatedlargelanguagemodels}, with very few studies on adaptability and trustworthiness~\cite{kang2024grounding,wu2025surveyfederatedfinetuninglarge,li2024position}. Moreover, these works rarely provide a unified taxonomy that captures the full landscape. Also, none of them considers all three important metrics on efficiency, adaptability, and trustworthiness holistically. To the best of our knowledge, the work by~\citet{ren2024advances} represents the only prior attempt to provide a comparatively comprehensive taxonomy across various dimensions of \fedfm. However, this work does not consider adaptability a critical dimension; instead, it structures its taxonomy around the conventional FL lifecycle. We argue that this emerging paradigm necessitates a novel perspective to fit new scenarios and demands of FMs. 

Recognizing gaps in the existing literature, we conduct a systematic review of \fedfm research to provide a thorough exploration of system performance in the integration of FMs and FL. More specifically, we provide a comprehensive study and analysis of the existing work across three major performance perspectives: efficiency, adaptability, and trustworthiness. Our main contributions are detailed as follows:
\begin{itemize}[leftmargin=*]
\item We present a novel multi-tiered taxonomy for \fedfm. This framework systematically classifies existing techniques based on their characteristics and modeling approaches. It serves as a foundation for \fedfm methodologies, guiding the design of future solutions.
\item We discuss publicly available libraries and benchmarks, including their attributes, to make it easy for readers to get started with their research.
\item We introduce typical applications of \fedfm in critical domains like NLP, Recommendation, and Healthcare.
\end{itemize}

The remainder of this survey is organized as follows: Section~\ref{sec:background} introduces background on FMs and FL. Section~\ref{sec:motivation_challenges} presents the motivation and challenges for synergizing FMs and FL. Section~\ref{sec:efficiency}, \ref{sec:adaptation}, and \ref{sec:sp} review and summarize the recent progress from the three aspects of efficiency, adaptation, and trustworthiness. Section~\ref{sec_application} explores the applications across various domains. Before concluding, we discuss representative future directions in Section~\ref{sec:directions}.

\section{Background}
\label{sec:background}

This section briefly describes the background of FMs and FL, respectively.

\subsection{Foundation Models}
\label{sec:fm}

An FM is a model that can be adapted to a wide array of tasks through fine-tuning after initial pre-training~\cite{bommasani2021opportunities}. The lifecycle of FMs typically involves pre-training on extensive generic data to establish the basis of their abilities~\cite{bubeck2023sparks}, followed by adaptation to downstream tasks such as domain-specific question answering~\cite{zhang-etal-2023-fedlegal}, and ultimately application in various domains.

FMs have sparked a significant paradigm shift in various fields of AI such as NLP, CV, speech and acoustics, and beyond. In the realm of NLP, the most prominent example is Large Language Models (LLMs) with substantial parameter sizes~\citep{zhao2023survey}. These models, such as ChatGPT and GPT-4~\citep{openai2023chatgpt, openai2023gpt4}, demonstrate exceptional abilities in natural language understanding and generation, enabling them to comprehend and respond to user inputs with remarkable contextual relevance. This capability proves invaluable in applications like customer service, virtual assistants, and chatbots, where effective communication is paramount. Moreover, LLMs eliminate the need for training models from scratch for specific tasks, be it machine translation, document summarization, text generation, or other language-related tasks.

In the realm of CV and other modalities, FMs have also made remarkable progress. Vision Transformers (ViTs)~\cite{dosovitskiy2021an} segment images into distinct patches, which serve as inputs for transformer architectures. SAM~\cite{kirillov2023segment} can segment anything in images according to the input prompts. CLIP~\cite{radford2021learning} bridges the gap between text and images through contrastive learning. DALL$\cdot$E, proposed by~\citet{ramesh2021zero}, generates images from textual descriptions, expanding the possibilities of creative image generation. Additionally, models like GAto~\citep{reed2022generalist}, exhibit versatility by being applicable across various tasks such as conversational agents, robotic control, and gaming.

\subsection{Federated Learning}
\label{sec:fl}

FL~\cite{mcmahan2017communication} is a learning paradigm that enables a collection of clients to collaboratively learn a shared global model by leveraging their private datasets in a distributed manner, assisted by the coordination of a central server.
The general goal of FL is to find a parameter set $\bm{\theta}$ that minimizes the following distributed optimization objective:
	\begin{equation}
		\label{obj}
		\min\limits_{\bm{\theta}} F(\bm{\theta}) :=  \frac{1}{K} \sum_{k \in [K]}  F_k(\bm{\theta}),
	\end{equation}
where $K$ represents the total number of clients and $F_k(\bm{\theta}) = \mathbb{E}_{\bm{z}\sim \mathcal{D}_k} [\loss(\bm{\theta}; \bm{z})]$ denotes the expected risk of the $k$-th client. Here, $\mathcal{D}_k$ is the data distribution for the $k$-th client, and $\loss(\cdot;\cdot)$ is a user-specified loss function. 

The most representative algorithms in the FL literature %
are the \fedavg-family algorithms~\cite{mcmahan2017communication,reddi2020adaptive}. The standard \fedavg involves periodic interactions between the server and decentralized clients to exchange trainable model parameters. 
 In this process, each client independently trains the model on its local data and sends the model updates to a central server. The server aggregates these updates by computing their average to update the global model, which is subsequently redistributed to the clients for further iterations.
 Many variants have been proposed to tackle issues such as convergence and local data heterogeneity~\cite{diao2021heterofl}. For example, \algfont{FedProx}~\cite{MLSYS2020_1f5fe839} and \algfont{FedDyn}~\cite{acar2021federated} introduce regularizer terms to penalize client updates that are far away from the server model. A general framework \algfont{FedOpt}~\cite{reddi2020adaptive} unifies adaptive optimizers (\textit{Adam}, \textit{Yogi}, \etc.) and demonstrates superior convergence speed when compared to the naive \fedavg.

FL offers an efficient privacy-preserving way to train models on large-scale and diverse data \cite{kairouz2021advances}, leading to its application across various domains such as healthcare~\cite{lincy2020early, rieke2020future, joshi2022federated}, finance~\cite{chatterjee2023use, liu2023efficient}, and smart cities~\cite{ramu2022federated, PANDYA2023102987}.

\section{\fedfm: Motivation \& Challenges}
\label{sec:motivation_challenges}

In this section, we first motivate the synergy of FMs and FL (Section~\ref{sec:motivation}), then summarize the key challenges (Section~\ref{sec:challenge}).

\subsection{Motivation}
\label{sec:motivation}
The integration of FMs and FL represents a compelling collaboration that leverages each other's strengths to address their respective limitations, embodying a complementary relationship \cite{zhuang2023foundation,li2024position}.

    \paragraph{FL expands data availability for FMs} By leveraging data from a wide range of sources in a privacy-preserving manner, FL makes it possible to build models on sensitive data in specific domains, such as healthcare \cite{lincy2020early, joshi2022federated, rieke2020future} and finance \cite{chatterjee2023use, liu2023efficient}. This enhances the diversity and volume of training data, improving model robustness and adaptability. Moreover, FL enables the integration of personal and task-specific data, allowing FMs to be customized for personal applications. For instance, Google has trained next-word-prediction language models on mobile keyboard input data with FL to improve user experience~\cite{xu-etal-2023-federated,10.1145/3494834.3500240}.
    
    \paragraph{FMs boost FL with feature representation and few-shot learning capabilities} %
    By pre-training on large-scale generic data, FMs acquire essential knowledge and understanding capabilities~\cite{brown2020language}, providing multiple benefits to FL. Firstly, they benefit FL systems by offering advanced feature representations and learning capabilities from the outset. Secondly, leveraging the pre-learned knowledge of FMs can accelerate the FL process, enabling efficient and effective adaptation to specific tasks with minimal additional training. Thirdly, FMs' powerful generative capabilities could help FL overcome the data heterogeneity challenge by synthesizing extra data, thus accelerating model convergence~\cite{10398264}.

\subsection{Core Challenges}

\label{sec:challenge}

In this section, we discuss challenges arising from \fedfm across three major performance perspectives: {efficiency}, {adaptability}, and {trustworthiness}. They are considered important metrics for training and running the FMs across diverse downstream applications. We will explain them one by one below.

\paragraph{Efficiency Challenges}
\label{sec:systemchallenges} 
Efficiency challenges stem from the mismatch between the significant resource demands of FM training and the limited, heterogeneous system resources (\eg mobile devices) within FL systems, such as communication bandwidth, computational power, and memory~\cite{su2023fedra}. The communication bottleneck of FL is induced by frequently exchanging training information between the server and clients over limited bandwidth channels~\cite{kairouz2021advances}. The substantial number of parameters in FMs further exacerbates this burden, thus hindering the training process.

\paragraph{Adaptability Challenges} 
\label{sec:taskchallenges}

Adaptability challenges arise from the adaptation of an FM to a specific downstream task (\eg by fine-tuning)
in FL settings. Key challenges include \textit{data heterogeneity} and \textit{resource heterogeneity}. Performance degradation in FL, attributed to heterogeneous data distributions among clients, is a well-recognized issue \cite{kairouz2021advances,9835537}. A recent study~\cite{babakniya2023slora} has shown that such a performance penalty is even more substantial when fine-tuning FMs. For NLP tasks, data heterogeneity can manifest as variations in language, style, topic, or sentiment across datasets held by different clients. %
In multi-modal scenarios, the challenge is even more pronounced due to the inherent diversity in data types (\eg text, images, and audio) \cite{yu2023multimodal}. Addressing data heterogeneity involves not just identifying and measuring it but also developing algorithms that are robust to such diversity, ensuring that the model can learn effectively from varied data contributions without compromising on performance. In terms of resource heterogeneity, the memory and computational resources of the devices for different participants may be diverse~\cite{diao2021heterofl}, which could cause delays for model synchronization and inactivation of some participants, \ie stragglers, making it challenging to leverage the full potential of FMs in FL settings.

\paragraph{Trustworthiness Challenges}
\label{sec:spchallenges} Trustworthiness challenges emphasize the concerns regarding privacy, security, and ethical considerations in the lifecycle of \fedfm, from the pre-training and model adaptation to the application stages. We present two representative challenges from this perspective: (1) \textit{intellectual property}: Intellectual Property (IP) protection in \fedfm primarily involves attributing ownership rights for both models and data. 
From the server's perspective, broadcasting a pre-trained model to multiple nodes for fine-tuning poses IP protection and security risks (\eg model theft), necessitating measures to safeguard IP rights and ensure model integrity~\cite{kang2024grounding}; (2)
\textit{privacy leakage}: Although FL does not immediately share data, studies have shown that it may not always guarantee sufficient privacy preservation~\cite{NEURIPS2020_c4ede56b}, as model parameters (\eg weights or gradients) may leak sensitive information to malicious adversaries~\cite{NEURIPS2019_60a6c400}. 
(3)\textit{Poisoning Attacks}: FL systems are inherently vulnerable to attacks due to their wide attack surface and reliance on network communication~\cite{10018261}. Poisoning attacks are carried out by malicious participants, aiming to bias the global model to the desire of attackers.

In the following sections (Sections \ref{sec:efficiency}-\ref{sec:sp}), we will review and discuss the existing work with respect to efficiency, adaptability, and trustworthiness.

\begin{figure*}
    \centering
    
\tikzset{
    basic/.style  = {draw, text width=3cm, rounded corners=1.5pt, align=center, font=\scriptsize, rectangle},
    root/.style   = {basic, thin, align=center, fill=green!30},
    l1node/.style = {basic, thin, align=center, fill=cosmiclatte!30, text width=1.5cm,},
    l2node/.style = {basic, thin, align=left, fill=peach!50, text width=5em, align=center},
    l3node/.style = {basic, thin,  align=center, fill=amber!50,text width=3.cm},
    leaf/.style = {basic, thin, align=left, fill=darkseagreen!30, text width=16.em},
    wide leaf/.style = {basic, thin, align=left, fill=darkseagreen!30, text width=26.6em},
    edge from parent/.style={draw=black, edge from parent fork right}
}

\resizebox{\linewidth}{!}{

\begin{forest} for tree={
        edge path={\noexpand\path[\forestoption{edge},->, >={Latex[length=1.mm,width=1.mm]}] 
             (!u.parent anchor) -- +(4pt,0pt) |-  (.child anchor) 
             \forestoption{edge label};},
        grow=east,
        reversed=true,
        anchor=base west,
        parent anchor=east,  
        child anchor=west,  
        base=center,
        font=\small,
        rectangle,
        draw=hidden-draw,
        rounded corners,
        align=center,
        minimum width=4em,
        edge+={semithick, draw=hidden-draw,line width=0.5pt},
        s sep=2pt,
        l=20pt,
        l sep=8pt,
        inner xsep=4pt,
        inner ysep=3pt,
        ver/.style={rotate=90, child anchor=north, parent anchor=south, anchor=center},
},
    where level=3{l sep = 6pt}{}
[\fedfm, ver,  
    [{Efficiency \\ (\S\ref{sec:efficiency})}, l1node,  
        [Parameter-Efficient \\ Fine-Tuning, l2node, 
            [Selective, l3node,
                [
                    {\Eg~\textit{RaFFM}~\cite{yu2023bridging}, 
                    \textit{FedBF}~\cite{zhang2023fedpetuning} 
                    }, leaf]
            ]
            [Additive, l3node,
                [
                    {\Eg~\textit{FedCLIP}~\cite{lu2023fedclip}, \textit{FedDAT}~\cite{chen2023feddat}
                    }, leaf]
            ]
            [{Reparameterization-based}, l3node,
                [
                {\Eg~\textsc{HetLoRA}~\cite{cho2024heterogeneous}, \textit{FedDPA}~\cite{yang2024dualpersonalizing}}, leaf
                ]
            ]
        ]
        [Model \\ Compression, l2node, 
            [
                Federated Distillation, l3node,
                [{\Eg~\textit{FedMKT}~\cite{fan2024fedmktfederatedmutualknowledge}, \textit{FedPFT}~\cite{peng2024fedpft}},leaf]
            ]
            [Sparsification, l3node,
                [{\Eg~{\tiny PruneFL}~\cite{9762360}, 
                {\tiny FLASH~\cite{babakniya2023revisiting}}},leaf]
            ]
            [Quantization, l3node,
                [{\Eg~\textit{FedSplitBERT}~\cite{9892845}, FedLPP~\cite{zhu2024promotingdatamodelprivacy}},leaf]
            ]
            ] 
        [Zeroth-Order \\ Optimization, l2node,
            [Gradient Estimation, l3node,
                [{\Eg~\textit{BAFFLE}~\cite{feng2023does}, {\textit{FedZeN}~\cite{maritan2023fedzen}}},leaf]
            ]
            [Others, l3node,
                [{\Eg~FedBPT~\cite{sun2023fedbpt}, pFedGFP~\cite{anonymous-acl24-arr}},leaf]
            ]
        ]
    ]
    [Adaptability\\ (\S\ref{sec:adaptation}), l1node,  
        [Domain-Centric, l2node, 
            [Domain-Adaptive Pre-Training, l3node,
                [
                    {
                    \Eg~\textit{FMTDA}~\cite{9706703},
                    \textit{FEDBFPT}~\cite{ijcai2023p483}
                    }, leaf
                ]
            ]
            [Multi-Domain Adaptation, l3node,
                [
                    {\Eg~FedAPT~\cite{su2023federated}, DiPrompT~\cite{bai2024diprompt}}, leaf
                ]
            ]
        ]
        [Client-Centric, l2node,
            [Personalization, l3node,
                [{
                    \Eg~\textit{FedDAT}~\cite{chen2023feddat}, 
                    \textit{Fed-MNMT}~\cite{liu-etal-2023-communication}},leaf
                ]
            ]
            [Client Clustering, l3node,
                [{
                    \Eg~\textit{FedLFC}~\cite{guo-etal-2024-fedlfc},~\textit{FedHLT}~\cite{10.1145/3589335.3651933},
                    \textit{FL-TAC}~\cite{ping2024fltac}},leaf
                ]
            ]
            [Preference-Aware Adaptation, l3node,
            [{
               ~\textit{FedRLHF~\cite{fan2024fedrlhfconvergenceguaranteedfederatedframework}}},leaf
            ]
        ]
        ]
        [System-Centric, l2node, 
            [Resource Heterogeneous, l3node,
                [
                    {\Eg~\textit{FedRA}~\cite{su2023fedra}, \textsc{HetLoRA}~\cite{cho2024heterogeneous}},leaf
                ]
            ]
            [
                Split Learning, l3node,
                [\Eg~{\textit{FedBERT}~\cite{10.1145/3510033}, {\textit{FedSplitX}~\cite{shin2023fedsplitx}}},leaf]
            ]
        ]  
    ]
    [Trustworthiness \\ (\S\ref{sec:sp}), l1node,  
        [IP Protection, l2node, 
            [Watermarking, l3node,
                [
                    {\Eg~\textit{WAFFLE}~\cite{9603498}, \textit{DUW}~\cite{yu2023who}},leaf
                ]
            ]
            [Black-Box Tuning, l3node,
                [
                \Eg~{\textit{Fed-BBPT}~\cite{abs-2310-03123}, 
                \textit{pFedGPT}~\cite{anonymous-acl24-arr}},leaf
                ]
            ]
        ]
        [Privacy Protection, l2node,
            [Privacy-Preserving Techniques, l3node,
                [
                {
                \Eg~\textit{DP-FTRL}~\cite{xu-etal-2023-federated},
                \textit{DP-LoRA}~\cite{liu2023differentially}
                },
                leaf
                ]
            ]
            [Privacy Attack, l3node, 
                [
                 {\Eg~FILM~\cite{NEURIPS2022_35b5c175}, 
                 DRA~\cite{zhang-etal-2024-revisiting-data}
                 }, leaf
                ]
            ]
        ]
        [Byzantine \\ Robustness, l2node,
            [Untargeted Attacks, l3node,
                [{\Eg~Fed-EBD~\cite{10.1007/978-981-97-2259-4_13}, SA-FedIT~\cite{ye2024emergingsafetyattackdefense}}, leaf
                ]
            ]
            [Targeted Attacks, l3node,
                [{\Eg~{ PaaA}~\cite{li2024peftasanattackjailbreakinglanguagemodels}, {\tiny Fed-FA~\cite{NEURIPS2023_c39578c8}}}, leaf
                ]
            ]
            [Defense Techniques, l3node,
                [{\Eg~{\tiny ClippedClustering}~\cite{10018261}, {\tiny Fed-FA~\cite{NEURIPS2023_c39578c8}}}, leaf
                ]
            ]
        ] 
         ]
    ]
\end{forest}
}
    \caption{Taxonomy of research in foundation models with federated learning.}
    \label{fig:taxo}
\end{figure*}

\section{Efficiency}

\label{sec:efficiency}

The remarkable capabilities of FMs are often accompanied by substantial resource demands, posing significant challenges regarding efficiency of \fedfm. To address these bottlenecks, research has increasingly prioritized resource-efficient approaches. In this section, we review and discuss the current state of the art in efficiency, focusing on three primary streams: parameter-efficient fine-tuning, model compression, and zeroth-order optimization. We further break down each category into sub-areas for more detailed analysis.

\subsection{Parameter-Efficient Fine-Tuning} 
\label{sec:peft}

Federated Parameter-Efficient Fine-Tuning (FedPEFT), originating from the fine-tuning practices of FMs~\cite{lester-etal-2021-power,hu2022lora,li2021prefix}, is a suite of techniques designed to reduce both the computational load and the associated communication overheads~\cite{pmlr-v232-malaviya23a,woisetschlager2024survey}. 
In alignment with existing FM fine-tuning taxonomies~\cite{lialin2023scaling,ding2023parameter}, we present FedPEFT methods in three categories:  {selective methods},  {additive methods}, and  {reparameterization-based methods}. 

\subsubsection{Selective Methods}\label{subsubsec:peft:selective}
Selective methods fine-tune a small subset of the parameters, leaving the majority unchanged. In the field of LLMs, a prominent example of such methods is {BitFit}~\cite{ben-zaken-etal-2022-bitfit}, which only fine-tunes the bias terms. {BitFit} has inspired a series of studies in FedPEFT~\cite{bu2022differentially,sun2022conquering,zhang2023fedpetuning},
demonstrating the superior communication efficiency of only updating the bias terms while still achieving competitive performance. More sophisticated methods strive to find sparse subnetworks for partial fine-tuning. Among them, various methods \cite{9593126,9708944,tamirisa2023fedselect} advocate for the Lottery Ticket Hypothesis (LTH) \cite{frankle2018the}, positing that a dense network contains many subnetworks whose inference capabilities are as accurate as that of the original network. {FedSelect} \cite{tamirisa2023fedselect} is a representative method that encourages clients to find optimal subnetworks based on LTH and continually fine-tunes these derived subnetworks to encapsulate local knowledge.  As another important aspect, {RaFFM} \cite{yu2023bridging} proposes to prioritize specialized salient parameters by ranking them using salience evaluation metrics such as the $\ell_1$ and $\ell_2$ norms.

\subsubsection{Additive Methods}
\label{subsubsec:peft:additive}
Instead of fine-tuning a subset of model parameters, additive methods incorporate lightweight trainable blocks into frozen FMs and tune the additional parameters for model adaptation. These methods not only enhance computational and communicational efficiency but also introduce an extra benefit: personalization~\cite{lu2023fedclip}, \ie the integration of these supplementary parameters allows for the customization of heterogeneous models tailored to specific local data characteristics or user preferences. Key branches within additive methods include {adapter tuning} and {prompt tuning}. {Adapter tuning} integrates small-scale neural networks (known as ``adapters'') into the pre-trained models~\cite{pmlr-v97-houlsby19a, hu2022lora}. On the other hand, {prompt tuning} incorporates trainable task-specific continuous prompt vectors at the input layer~\cite{10.1145/3560815,dong-etal-2023-tunable}. %

\label{sec_additive_tuning}
\paragraph{Adapter Tuning}
Adapter tuning integrates small-scale neural networks (known as ``adapters'') into the pre-trained models~\cite{pmlr-v97-houlsby19a, hu2022lora}. A straightforward implementation of adapter tuning is to collaboratively train a shared adapter among all clients in the \fedavg manner, as highlighted by~\citet{sun2022conquering}. Based on \fedavg, {FedCLIP}~\cite{lu2023fedclip} incorporates an attention-based adapter for the image encoder in CLIP models~\cite{radford2021learning}. 
In the domain of multilingual machine translation, where different language pairs exhibit substantial discrepancies in data distributions, {Fed-MNMT}~\cite{liu-etal-2023-communication} explores clustering strategies that group adapter parameters and makes inner-cluster parameters aggregation for alleviating the undesirable effect of data discrepancy. Another representative approach named {C2A}~\cite{kimetal2023client} employs hypernetworks~\cite{ha2017hypernetworks} %
to generate client-specific adapters by conditioning on the client's information, maximizing the utility of shared model parameters while minimizing the divergence caused by
data heterogeneity.

\paragraph{Prompt Tuning}

\label{sec:prompt_modality}

Prompt tuning incorporates trainable task-specific continuous prompt vectors at the input layer~\cite{10.1145/3560815,dong-etal-2023-tunable}.  Compared to full fine-tuning, it achieves comparable performance but with $1000\times$ less parameter storage and communication~\cite{jia2022visual}. A variation of prompt tuning, {FedPerfix}~\cite{sun2023fedperfix} uses a local adapter to generate the prefixes and aggregate the original self-attention layers. Depending on target modalities, prompt tuning in current literature can be further classified into three categories:
\begin{itemize}[leftmargin=*]
    \item \textbf{Textual Prompt Tuning.} Task-specific prompt embeddings are combined with the input text embeddings, which are subsequently fed into language models. These soft prompts serve as instructive contexts to influence the generation process of LLMs by steering the probability distribution of the next token~\cite{dong-etal-2023-tunable}.
    \item \textbf{Visual Prompt Tuning.} Taking inspiration from advances in efficiently tuning LLMs, prompts are also introduced in the input space of vision models~\cite{jia2022visual}. Naive implementations introduce prompts at the pixel level, acting as a form of data augmentation~\cite{li2023visual}. Alternatively, one could also insert the prompts as latent vectors for the first Transformer layer~\cite{deng2023unlocking,yang2023efficient}. Nevertheless, an empirical study~\cite{jia2022visual} has suggested that it is easier for visual prompts to learn condensed task-dependent signals in the latent input space of Transformers.
    \item \textbf{Textual-Visual Prompt Tuning.} Unlike single-modal FMs, vision-language FMs can process and interpret both visual data and textual information, endowing them with powerful representation ability and transferability~\cite{radford2021learning}. Based on vision-language FMs like CLIP, textual-visual prompt tuning shows promising capabilities in FL~\cite{guo2023promptfl}, especially in cross-domain scenarios, where the model needs to generalize across varied domains and unseen classes~\cite{qiu2024textdriven}.
\end{itemize}

\subsubsection{Reparameterization-based Methods}
\label{subsubsec:peft:repara}

The hypothesis behind reparameterization-based methods is that fine-tuning adaptations can be re-parameterized into optimization within low-rank subspaces~\cite{aghajanyan-etal-2021-intrinsic}. Low-Rank Adaptation (LoRA)~\cite{hu2022lora}, as a popular PEFT method from the area of LLMs, reduces the number of trainable parameters for downstream tasks by representing the weight updates with two smaller matrices (called update matrices) through low-rank decomposition~\cite{ding2023parameter}. When optimizing a parameter matrix $\mathbf{W} \in \mathbb{R}^{m \times n}$, the update equation can be written as:   $\mathbf{W} \leftarrow \mathbf{W} + \Delta \mathbf{W}$. 
The core idea of LoRA is to freeze the original matrix $\mathbf{W}$ while approximating the parameter update $\Delta \mathbf{W}$ by low-rank decomposition matrices, \ie $\Delta \mathbf{W}=\mathbf{A}\cdot  \mathbf{B}^\top$, where $\mathbf{A}\in \mathbb{R}^{m \times k}$ and $\mathbf{B}\in \mathbb{R}^{n \times k}$ are the trainable parameters for task adaptation and $k \ll \min(m,n)$ is the reduced rank. The trainable parameter size is then reduced from $mn$ to $ k(m+n)$. The major benefit of LoRA is that it can largely save memory and storage usage. A straightforward way to perform federated fine-tuning with LoRA is to train the LoRA modules $\mathbf{A}$ and $\mathbf{B}$ with homogeneous rank $k$ across all clients with standard FL such as \fedavg~\cite{mcmahan2017communication}. Serval studies have shown that this method can achieve an outstanding level of trade-off between performance and communication overhead for a wide range of FMs, including language models~\cite{zhang2023towards,zhang2023fedpetuning}, vision-language models~\cite{nguyen2024flora}, and speech-to-text models~\cite{du2024communicationefficient}.

\begin{figure}[!t]
    \centering
    \includegraphics[width=0.6\linewidth]{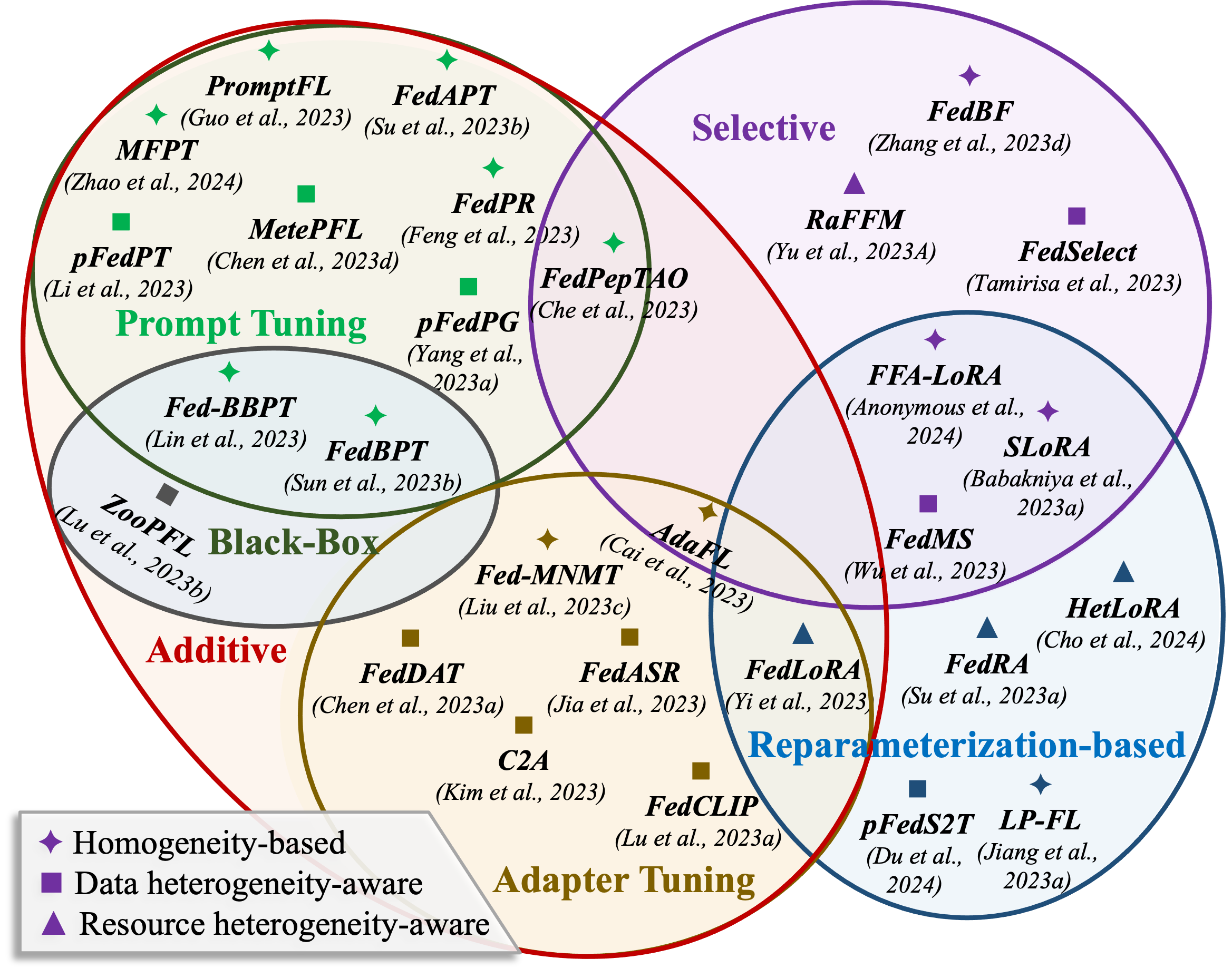}
    \caption{Taxonomy of Federated Parameter-Efficient Fine-Tuning (FedPEFT). Apart from efficiency, some methods also account for other considerations, such as data and resource heterogeneity challenges that are identified in Section~\ref{sec:taskchallenges} and black-box tuning (see Section~\ref{sec:sp}).}
    \label{fig:fedPEFT}
\end{figure}

\subsubsection{Comparison of FedPEFT Methods}

\label{sec:peft_comparing}

Figure~\ref{fig:fedPEFT} depicts the FedPEFT taxonomy and its representative methods. Note that some methods may belong to multiple overlapping categories. To compare the communication efficiency of different FedPEFT methods, Table~\ref{tab:my_label} gives a brief overview of experimental evaluations from representative studies. Compared to full-parameter fine-tuning, FedPEFT methods only require 0.1\%-30\% communication overhead. We note that the differences can be attributed to several factors, including model complexity and implementation details.

\subsection{Model Compression}

Although FedPEFT is effective in reducing the number of trainable parameters, it does not inherently reduce the structural complexity of FMs. To further optimize the system efficiency, model compression serves as a core strategy that physically reduces the model size and computational complexity~\cite{9660377}. This strategy offers two-fold benefits to the FedFM paradigm: primarily, it enables the participation of weak learners (clients with low-end hardware) and enhances inference efficiency by lowering memory requirements; simultaneously, it serves to alleviate communication bottlenecks by compressing the data exchanged between clients and the server. In this section, we introduce three main categories of model compression, namely {federated distillation}, {sparsification}, and {quantization}.

\begin{table*}[]
    \newcommand \rotangle{90}
    \newcommand{\rotbox}[1]{\rotatebox[origin=c]{\rotangle}{\textbf{#1}}}
    \centering
    \caption{\small Comparison of Federated Parameter-Efficient Fine-Tuning (FedPEFT) Methods.}
\renewcommand{\arraystretch}{1.1}
\setlength\extrarowheight{1.5pt}
\begin{adjustbox}{width=0.98\linewidth,center}
    \begin{tabular}{c|c|lcccccc}
        \toprule
        \multicolumn{2}{c|}{{\makecell[c]{\textbf{Category}}}}  & \textbf{Representative Work}   &\textbf{Modality}&\textbf{Model}& \makecell[c]{\textbf{\# Full} \\ \textbf{Params.}}& \makecell[c]{\textbf{\# Train.} \\ \textbf{Params.}}& \makecell[c]{\textbf{Training } \\ \textbf{Accel.}} & \makecell[c]{\textbf{Comm. } \\ \textbf{Cost}}  \\
        \midrule
        \multicolumn{2}{c|}{\multirow{2}{*}{{\makecell[c]{\textbf{Selective}}}}}   &{RaFFM}~\cite{yu2023bridging}   & Txt. & BERT-Large~\citeyearpar{devlin-etal-2019-bert} & 336M & 100M & $6.13\times$ & 29.8\% \\
        \multicolumn{2}{c|}{}  & {FedBF}~\cite{zhang2023fedpetuning} &Txt.&Roberta-Base~\citeyearpar{liu2019roberta} & 125M & 0.66M &   &  $1.6\%$\\
        \midrule
        \multirow{10}{*}{{\makecell[c]{\textbf{Additive}}}} & \multirow{6}{*}{{\makecell[c]{\textbf{Adapter}}}} & {FedAP}~\cite{zhang2023fedpetuning} & Txt. & Roberta-Base~\citeyearpar{liu2019roberta} & 125M & 2M &   &  $1.6\%$\\
            &  & {FedCLIP}~\cite{lu2023fedclip} &Vis.-Txt. & ViT-B/32~\citeyearpar{wu2020visual} & 150M & 0.53M &  & 3.5\%\\
            && {FedDAT}~\cite{chen2023feddat}   &Vis.-Txt.& ALBEF~\citeyearpar{NEURIPS2021_50525975} & 290M  & 2.86M &    & 9.9\%\\
            && C2A~\cite{kimetal2023client}   &Txt.& DistilBERT~\citeyearpar{sanh2020distilbert} & 66M  & 0.06M & & 0.1\%\\
           &&  \textit{Fed-MNMT}~\cite{liu-etal-2023-communication}   &Txt.& mBART-50~\citeyearpar{tang2020multilingual} & 611M  & 8M &  & 1.3\% \\
           && {AdaFL}~\cite{10.1145/3570361.3592505}   &Txt.& BERT~\citeyearpar{devlin-etal-2019-bert} & 110M  & 0.61M &   $1.63\times$ & 0.6\%\\
        \cmidrule{2-9}
            & \multirow{4}{*}{{\makecell[c]{\textbf{Prompt}}}} & {PromptFL}~\cite{guo2023promptfl}   &Vis.-Txt.& ViT-B/16~\citeyearpar{dosovitskiy2021an}&   87M& 0.87M   & $2.38\times$ &  0.9\% \\
            && {MFPT}~\cite{zhao2024breaking}   &Txt.& XLM-RoBERTa~\citeyearpar{conneau-etal-2020-unsupervised} & 270M  & 1.2M &   & 0.4\% \\
          && {FedAPT}~\cite{su2023federated}   &Vis.-Txt.& ViT-B/32~\citeyearpar{wu2020visual} & 88M & 2.8M &   & 3.2\% \\
           && \textit{FedSP}~\cite{dong-etal-2023-tunable} &Txt.& GPT2-XL~\citeyearpar{radford2019rewon} & 1.6B & 111M &   &  0.5\% \\
        \midrule
     \multicolumn{2}{c|}{\multirow{5}{*}{{\makecell[c]{\textbf{Reparameterization-based} \\ \textbf{Methods}}}}}   & {SLoRA}~\cite{babakniya2023slora}  &Txt.& DistilBERT~\citeyearpar{sanh2020distilbert} & 67M & 0.7M & $13.47\times$ &    5.8\% \\
   \multicolumn{2}{c|}{}  & {LP-FL}~\cite{jiang2023lowparameter}  &Txt.&  BERT-Large~\citeyearpar{devlin-etal-2019-bert} & 336M & 100M & & 30\% \\
     \multicolumn{2}{c|}{} & FedFMS~\cite{liu2024fedfms} &Vis.-Txt.& ViT-B/16~\citeyearpar{dosovitskiy2021an}& 87M& 8.6M& & 10\%\\
     \multicolumn{2}{c|}{} & {pFedS2T}~\cite{du2024communicationefficient} &Aud.& Whisper~\citeyearpar{pmlr-v202-radford23a} & 254M& 10.1M& &  4\%\\
        \multicolumn{2}{c|}{} & {FFA-LoRA}~\cite{sun2024improving}  &Txt.& RoBERTa-Large~\citeyearpar{liu2019roberta} & 355M & 0.39M & &  0.1\% \\
     \bottomrule
    \end{tabular}
\end{adjustbox}
    \label{tab:my_label}
\end{table*}

\subsubsection{Federated Distillation}
As a representative mechanism for efficient model deployment, Knowledge Distillation (KD)~\cite{hinton2015distilling} transfers knowledge from a large teacher model to a compact student model that inherits the teacher’s capabilities by aligning their outputs (\eg~logits) or intermediate representations. In the context of FL, recent work has extended KD into a collaborative paradigm known as Federated Distillation (FD)~\cite{jeong2018communication, li2024federated}, where clients share compact knowledge instead of large model parameters. Depending on the direction of knowledge transfer, FD techniques can be generally categorized into two types: {upstream distillation} and {downstream distillation}, as discussed below:

    \paragraph{Upstream Distillation.} The basic idea of this direction is to distill the knowledge of multiple local models into the shared global model. Instead of transmitting massive model parameters, clients often act as teachers (or co-learners) by uploading refined knowledge representations or lightweight adapter parameters regularized by local distillation. Several pioneering studies in this direction propose to communicate knowledge in the form of soft-label predictions on records of a public distillation~\cite{jeong2018communication,NEURIPS2020_18df51b9}. As another representative framework, FedKD~\cite{wu2022communication} employs a small model and a large model to learn and distill knowledge from each other, where only the small model is shared by different clients and learned collaboratively, hereby reduces the communication cost. Despite the high communication efficiency, these approaches fail to alleviate the computational burden, as they still require clients to optimize the full models during local training. In contrast, advanced frameworks such as FedGKT~\cite{NEURIPS2020_a1d4c20b} and FedDAT~\cite{chen2023feddat} explicitly train a compact module in place of the large model, reducing both communication costs and local computational burden. 
    
    \paragraph{Downstream Distillation.} Resource-constrained clients are often incapable of training or hosting full-sized models due to prohibitive memory and computational costs. To bridge this gap, {Downstream distillation} employs a pre-trained FM as a teacher to transfer its general capabilities into lightweight client-specific models. One direction involves proxy-based adaptation, such as FedPFT~\cite{peng2024fedpft}, which utilizes a proxy mechanism to facilitate the fine-tuning of FMs for clients without direct parameter access. More recent studies go beyond simple alignment to incorporate generative reasoning capabilities for language models. For instance, PPC-GPT~\cite{fan-etal-2025-ppc} utilizes a server-side auxiliary LLM to generate synthetic data and Chain-of-Thought (CoT) rationales based on differentially private client prompts. By distilling both the predictive labels and the reasoning processes from the FM, the server compresses the model into a task-specific SLM before deploying it to clients for local retraining. Similarly, FedCoT~\cite{fan-etal-2025-fedcot} and other synthetic-data-driven methods explore transferring reasoning capabilities to edge devices without transmitting raw private data.

\subsubsection{Sparsification} The goal of sparsification is to reduce the large number of parameters and activations in FMs, thereby mitigating communication bottlenecks and alleviating resource constraints in both training and deployment environments. In this part, we discuss sparsification techniques along three dimensions: sparsification target, phase, and granularity.

    \paragraph{Sparsification Target.} Based on the specific object being compressed, techniques are taxonomized into three groups:
    (1) \textit{Communication sparsification} aims to compress model updates transmitted between clients and the server, typically without altering the local training process~\cite{9762360}. Representative approaches include Top-K pruning~\cite{kuo2024federatedlorasparsecommunication,11044514}, which retains only updates with the largest magnitudes; thresholding~\cite{chu2024send}, which filters out parameters below a specific limit; and sketching~\cite{pmlr-v119-rothchild20a}, which utilizes hashing to project updates into lower-dimensional structures. Building on these, FLASC~\cite{kuo2024federatedlorasparsecommunication} applies sparsity to LoRA adapters, while MetaSend~\cite{chu2024send} advances thresholding by introducing a learnable module to select informative parameters adaptively.
    (2) \textit{Parameter sparsification} targets the reduction of the model's computational and memory footprint by pruning less critical weights, thereby enabling sparse local training. Unlike communication sparsification, this modifies the local optimization loop. For instance, FedSpaLLM~\cite{bai-etal-2025-fedspallm} facilitates federated pruning, allowing clients to autonomously prune layer subsets aligned with their hardware capacities.
    (3) \textit{Activation sparsification} enhances efficiency by selectively engaging only a subset of neurons or modules during the forward pass. This concept is mainly realized in the Mixture-of-Experts (MoE) paradigm~\cite{6797059}, which utilizes a learned gating mechanism to activate a sparse set of expert subnetworks for each input.

    \paragraph{Sparsification Phase.} While existing efforts span the entire model lifecycle, recent studies predominantly focus on \textit{training-oriented sparsification}. This paradigm encompasses the aforementioned communication and parameter sparsification techniques, which operate within the training loop to accelerate convergence and reduce overheads. Some approaches also incorporate sparsification-aware learning mechanisms, such as the selective fine-tuning methods discussed in Section~\ref{subsubsec:peft:selective}. In addition, \textit{inference-oriented sparsification} targets the deployment stage, typically compressing the final global model to minimize latency and memory footprint on resource-limited devices. Research in this vein often leverages the LTH~\cite{frankle2018the,9593126,9708944,tamirisa2023fedselect} to identify accurate sparse subnetworks.

    \paragraph{Sparsification Granularity.} Categorized by the pattern of parameter elimination, techniques fall into two groups: {Unstructured sparsification} involves removing individual elements (weights or gradients) without a fixed pattern. Most communication sparsification methods (e.g., Top-$k$ update compression mentioned above~\cite{10019204,11044514}) fall into this category. While effective for reducing transmission data, unstructured sparsity is often difficult to accelerate on standard hardware, failing to alleviate local computational burdens.
    In contrast, {Structured sparsification} enforces sparsity at a coarser granularity, such as removing entire channels, layers, or blocks (as seen in parameter sparsification methods like FedSpaLLM~\cite{bai-etal-2025-fedspallm}). This characteristic facilitates valid sparse local training by making only specific dense substructures trainable, effectively reducing both memory usage and computational cost without requiring specialized hardware support.

\subsubsection{Quantization} Quantization involves reducing the precision of model parameters to decrease memory usage and computational requirements. This technique is well-established in both the FM and FL domains~\cite{xu2024survey,pmlr-v108-reisizadeh20a}. Quantization is orthogonal to other resource-efficient techniques, making it feasible to combine them for greater efficiency and flexibility~\cite{9892845}.

Hybrid compression techniques combine multiple compression methods to achieve better performance and efficiency. For example, a hybrid approach may use both quantization and sparsification to reduce model size and improve communication efficiency simultaneously. By leveraging the strengths of different techniques, hybrid compression can lead to more effective resource utilization in federated learning settings.

\subsection{Zeroth-Order Optimization}
In contrast to the use of gradient descent in most FL optimization algorithms, a particular line of research advocates for the removal of BackPropagation~(BP) \cite{malladi2023fine} in favor of Zeroth-Order Optimization (ZOO) \cite{9917343, 9613620}. BP-free methods conserve memory needed for computing gradients and minimize communication overhead for model aggregation \cite{qin2023federated}, making FMs more accessible for lower-end devices, thereby enhancing their applicability in diverse hardware environments. 

	\paragraph{Perturbation-based Methods. } ZOO methods primarily rely on perturbation methods to estimate gradients with forward propagation.
Given a model with parameters $\bm{\theta} \in \mathbb{R}^d$ and a loss function $\mathcal{L}$, a typical gradient estimator estimates the gradient on a minibatch $\mathcal{B}$ as
\begin{equation}
    \hat\nabla\mathcal{L}(\bm{\theta};\mathcal{B}) =  \frac{\mathcal{L}(\bm{\theta} + \epsilon\bm{z};\mathcal{B}) - \mathcal{L}(\bm{\theta};\mathcal{B})}{2\epsilon}\bm{z}, 
    \label{eq:pre_one_point_estimator_def}
\end{equation}
where $\bm{z}\in\mathbb{R}^d$ with $\bm{z}\sim\mathcal{N}(0,\bm{I}_d)$ and $\epsilon$ is the \emph{perturbation scale}~ \cite{duchi2015optimal}. It requires only two forward passes through the model to compute the estimation of gradient, serving as a memory-efficient alternative to BP. However, Eq. \eqref{eq:pre_one_point_estimator_def} provides a biased gradient estimation, leading to a certain degree of information loss \cite{9186148}. Alternatively, many studies opt for two-point gradient estimators that can yield a more stable and reliable approximation~\cite {spall1992multivariate,malladi2023fine,abs-2310-03123,ling2024convergence}. The standard two-point gradient estimator estimates the gradient on a minibatch $\mathcal{B}$ as
\begin{equation}
    \hat\nabla\mathcal{L}(\bm{\theta};\mathcal{B}) =  \frac{\mathcal{L}(\bm{\theta} + \epsilon\bm{z};\mathcal{B}) - \mathcal{L}(\bm{\theta} - \epsilon\bm{z};\mathcal{B})}{2\epsilon}\bm{z}.
    \label{eq:spsa}
\end{equation}

Based on the above gradient estimation frameworks, recent work, such as that by~\citet{lu2023zoopfl,xu2024fwdllm}, has initiated preliminary explorations into both FedPEFT and full-parameter fine-tuning of billion-sized FMs, like LLaMA, on mobile devices. The naive ZOO methods remain impractical for training large FMs in standard FL frameworks such as \fedavg, as they still result in a significant communication burden for model aggregation. In light of this, FedKSeed~\cite{qin2023federated} was proposed to further reduce communication overheads between the server and clients by using just a few random seeds and
scalar gradients, requiring only a few thousand bytes for communication.

\paragraph{Others} Other resource-efficient techniques include model pruning and knowledge distillation. Model pruning reduces the size of a model by removing less important parameters, while knowledge distillation transfers knowledge from a larger model to a smaller one, enabling efficient deployment on resource-constrained devices.

Although ZOO methods have shown promise in resource-efficient FL~\cite{ling2024convergence}, they generally require many iterations to achieve strong performance~\cite{NEURIPS2023_a6278101}. Compared to the well-established BP-based optimization, ZOO is still in the early stages of development, particularly for \fedfm settings, necessitating further research and development.

\section{Adaptability}
\label{sec:adaptation}

Adaptability refers to the capability of tailoring a pre-trained FM to perform downstream tasks across varying FL settings and scenarios. This mainly includes the ability to learn from different domains, cater to individual user needs, and work across diverse devices while retaining overall performance and efficiency. We focus on three key aspects of adaptation, namely {domain-centric adaptation}, {client-centric adaptation}, and {system-centric adaptation}.

\subsection{Domain-Centric Adaptation}

Domain-centric adaptation focuses on adapting FMs within specific domains by addressing the domain diversity across client datasets.

\subsubsection{Domain-Adaptive Pre-Training} Despite being heavily reliant on large-scale and public datasets for their pre-training, FMs often require further Domain-Adaptive Pre-Training (DAPT) with domain-specific data for tasks that necessitate specialized knowledge~\cite{gururangandont, guo2022domain}.  
In domains like healthcare, FL allows for the continued pre-training of these models using sensitive, domain-specific data without compromising privacy. Based on this idea,~\citet{jiang2023fdapt} proposed {FFDAPT}, a computational-efficient further pre-training algorithm that freezes a portion of consecutive layers while optimizing the rest of the layers. Similarly,~\citet{ijcai2023p483} proposed {FEDBFPT} that builds a local model for each client, progressively training the shallower layers of local models while sampling deeper layers, and aggregating trained parameters on a server to create the final global model.

\subsubsection{Multi-Domain Adaptation}

Given that client data may belong to various domains in real-world FL scenarios, some efforts~\cite{feng2023adapterbased,su2023federated}  have been devoted to facilitating multi-domain collaborative adaptation.~\citet{feng2023adapterbased} applied a pre-trained CLIP to the multi-domain scenario and proposed an adaptive prompt tuning method that uses domain-specific keys to generate prompts
for each test sample. Furthermore,~\citet{su2023federated} employed knowledge distillation to selectively distill global knowledge based on an entropy measure, improving the generalization across different domains.

\subsection{Client-Centric Adaptation}

Client-centric adaptation aims to tailor FMs to meet the specific needs or preferences of individual clients while leveraging the decentralized and privacy-preserving nature of FL. This section covers three representative strategies for client-centric adaptation, namely personalization, client clustering, and preference-aware adaptation.

\subsubsection{Personalization} Vanilla FL trains a single-globally shared model to fit the ``average client'', which often struggles to capture the heterogeneous characteristics of individual clients. To mitigate this, personalization techniques prioritize adaptivity to the local client, tailoring the model to specific local distributions while preserving the benefits of collaborative learning.

\paragraph{Architecture-Based Personalization} By tailoring model architectures and the overall system designs, architecture-based personalization techniques decouple global knowledge sharing from local adaptation. 
Instead of training a monolithic model, these approaches often isolate specific parameter groups to exclusively capture user-specific patterns while aggregating the remaining parameters to maintain generalization. Leveraging FedPEFT, {FedSA-LoRA}~\cite{anonymous2025selective} proposes a selective aggregation strategy where the $A$ matrix of the LoRA module learns global representations, while the $B$ matrix is retained locally to preserve personalization. 
Similarly, {FedDPA}~\cite{yang2024dualpersonalizing} introduces a dual-adapter architecture that explicitly maintains separate global and local adapters, dynamically fusing them via a gating mechanism during inference. 
Taking a generative approach to architectural design, {pFedPG}~\cite{yang2023efficient} deploys a server-side prompt generator that dynamically produces client-specific visual prompts to adapt frozen FMs, thereby replacing direct parameter aggregation with personalized parameter generation. Instead of introducing additional learnable modules, {FedSelect}~\cite{tamirisa2023fedselect} employs a dynamic masking mechanism to personalize the selection of sub-networks for local fine-tuning based on parameter importance. 
Furthermore, works like {pFedLoRA}~\cite{yi2023fedlora} and {FedCLIP}~\cite{lu2023fedclip} leverage these modular designs to bridge model heterogeneity and align multi-modal features across diverse clients.

\paragraph{Optimization-Based Personalization} 
In contrast to structural modifications, optimization-based personalization focuses on tailoring the training dynamics and objective functions to accommodate local data distributions without necessarily altering the overall architecture. 
Instead of isolating specific parameters, these approaches guide the model toward personalized optimal points by adjusting the optimization trajectory, incorporating regularization terms, or modifying gradient updates. 
A prime example is {FedRLHF}~\cite{fan2024fedrlhfconvergenceguaranteedfederatedframework}, which integrates Reinforcement Learning from Human Feedback (RLHF) into the federated workflow. 
By utilizing local reward models to evaluate generation quality, it steers the global policy to align with diverse user preferences through personalized reward signals. 
In terms of optimization stability, recent studies on ASR~\cite{10389620} emphasize the critical role of optimizer-induced smoothness (e.g., using momentum-based optimizers) to navigate the non-IID loss landscapes of heterogeneous clients. 
Furthermore, approaches such as {FedCLIP}~\cite{lu2023fedclip} and communication-efficient strategies~\cite{du2024communicationefficient} employ gradient correction mechanisms or multi-stage fine-tuning schedules to balance the trade-off between global convergence and local feature retention.

\subsubsection{Client Clustering} Instead of tailoring models to individual clients, client clustering assigns specialized models to different groups of clients. The underlying assumption is that clients within a cluster possess high similarity in their data representations. A notable example is multilingual NLP, where clients exhibit linguistic heterogeneity characterized by inherent structural relationships, such as language families. Therefore,~\citet{guo-etal-2024-fedlfc} proposed FedLFC, which explicitly groups clients by language family affiliations and assigns separate LoRA modules to each cluster to minimize parameter interference across distantly related languages. Extending this concept from flat clusters to a structured hierarchy, FedHLT~\cite{10.1145/3589335.3651933} organizes language adapters into a language tree structure for a coarse-to-fine parameter sharing scheme that captures both universal linguistic features and specific leaf-node characteristics. Complementary to these parameter-centric approaches, FedKC~\cite{10.1145/3485447.3511988} suggests clustering within the latent representation space and exchanging ``knowledge centroids'' derived from local clusters, aligning the feature space across heterogeneous clients to handle skewed data distributions. Moving towards more dynamic grouping strategies, FL-TAC~\cite{ping2024fltac} proposes a task-specific adapter clustering mechanism; instead of grouping clients a priori, it clusters the learned low-rank adapters on the server side to identify and aggregate similar tasks data-drivenly. Furthermore, recent work~\cite{chen2025federatedfinetuningsparselyactivatedlarge} on sparsely-activated LLMs leverages a MoE architecture, which can be viewed as a form of dynamic parameter clustering, where diverse data patterns are automatically routed to specialized expert modules, achieving fine-grained adaptation without predefined partitions.

\subsubsection{Preference-Aware Adaptation}

In applications such as personalized chatbots, clients often desire tailored conversation styles and content that align with their individual preferences and needs.  To achieve this, preference-aware adaptation techniques leverage user feedback, such as ratings or explicit preferences, to guide the fine-tuning process. \citet{fan2024fedrlhfconvergenceguaranteedfederatedframework} proposes FedRLHF, which integrates RLHF principles with federated reinforcement learning to address privacy and personalization challenges simultaneously. Similarly, \citet{wu2025towards} propose a preference-aware federated learning framework that incorporates user feedback to enhance the personalization of FMs. This approach allows the model to learn from user interactions and adapt its responses accordingly, leading to improved user satisfaction and engagement. Another example is the work by \citet{wu2025towards}, which focuses on preference-aware fine-tuning of FMs using user feedback to optimize the model's performance in generating personalized content. Additionally, \citet{spadea2025federatedfinetuninglargelanguage} evaluate two preference-aware fine-tuning methods, \ie~Direct Preference Optimization (DPO) and Kahneman-Tversky Optimization (KTO), for LLMs in a federated setting. Their results indicate that KTO consistently outperforms DPO across various benchmarks, highlighting its flexibility and robust performance. They introduce a method that combines preference-aware fine-tuning with reinforcement learning techniques to further enhance the model's ability to generate contextually relevant and personalized responses. This approach leverages user feedback as a reward signal in a reinforcement learning framework, allowing the model to learn from both explicit preferences and implicit feedback signals.

\subsection{System-Centric Adaptation}

System-centric adaptation focuses on enhancing the adaptability of FL systems from an architectural perspective. It primarily addresses the challenges arising from resource heterogeneity, communication efficiency, and system scalability when fine-tuning FMs in federated environments. As FL increasingly scales to cross-device and cross-silo deployments involving devices with diverse computational and communication capabilities, it becomes essential to design mechanisms that ensure efficient model training and aggregation under such constraints. This section reviews representative strategies for system-centric adaptation, including resource-heterogeneous methods and split learning frameworks, which jointly aim to maintain system-level efficiency and fairness without compromising model performance.

\subsubsection{Resource-Heterogeneous Methods} \label{sec:hete}
Cross-device FL systems may be composed of devices equipped with heterogeneous resources, leading to disparities where certain devices exhibit more efficient model training than others \cite{chen2023feddat}. To address this issue, several methods have been developed to customize model architectures for resource-heterogeneous FL systems. In FL environments possessing heterogeneous resources, LoRA-based FedPEFT exhibits distinctive flexibility and adaptation in fine-tuning frozen FMs without overburdening client devices. \citet{su2023fedra} suggested assigning LoRA adapters to varying numbers of layers for heterogeneous clients according to a randomly generated mask matrix. An alternative and more targeted idea is to choose diverse LoRA ranks across clients based on their system capabilities. \citet{bai2024federated} proposed FlexLoRA to adjust local LoRA ranks dynamically. FlexLoRA reconstructs the uniform full-sized LoRA module $\Delta \mathbf{W}$ for server-side model aggregation followed by an SVD-based parameter redistribution. However, concurrent research by \citet{cho2024heterogeneous} has empirically demonstrated that the reconstruct-redistribute method suffers from performance loss compared to homogeneous LoRA. Instead, they proposed \textsc{HetLoRA}  that utilizes zero-padding to align module size before aggregation. It then truncates the global LoRA modules for the specific rank of the next selected clients. 

Given the success of MoE in scaling FMs efficiently~\cite{jiang2024mixtralexperts,10937907}, recent research has pivoted towards adapting MoE for distributed environments~\cite{10666083}. These efforts largely target inherent FL challenges such as data heterogeneity and resource heterogeneity. For instance, to capture diverse task patterns across clients, DGMoE~\cite{Miao_2025_ICCV} allows self-aware experts to adaptively determine their activation. To mitigate resource constraints, A$^3$SMoE~\cite{tran2025revisiting} and FFT-MoE~\cite{hu2025fftmoeefficientfederatedfinetuning} enable dynamic expert activation based on available client hardware. Furthermore, FedVLA~\cite{Miao_2025_ICCV} introduces an expert-driven aggregation strategy to ensure effective cross-client knowledge transfer. Recent works have adapted MoE for FL to address heterogeneity~\cite{10666083}, such as DGMoE~\cite{Miao_2025_ICCV} for self-aware expert activation, and A$^3$SMoE~\cite{tran2025revisiting} or FFT-MoE~\cite{hu2025fftmoeefficientfederatedfinetuning} for hardware-dynamic activation.

\subsubsection{Split Learning} Split learning addresses the resource heterogeneity between servers and clients by splitting a large model at a cut layer into client and server models~\cite{Thapa_MahawagaArachchige_Camtepe_Sun_2022}. For each training step, the output tensor, so-called smashed data, from the client model and the corresponding labels are transmitted over to the server. The server continues the forward propagation by processing the smashed data through its remaining layers; it then computes the loss using the transmitted label and performs backpropagation. The gradient generated
at the first layer of the server model is then transmitted back to the client for further backpropagation. 
Along this line, {FedBERT}~\cite{10.1145/3510033} proposes to leverage split learning for training the BERT model, showing the feasibility of training large FMs in FL settings.
{FedSplitX}~\cite{shin2023fedsplitx} is a more fine-grained method that allows multiple partition points for model splitting, accommodating more diverse client capabilities. Compared to conventional FL, split learning scales better with the size of FMs as it communicates only small-sized smashed data instead of model parameters~\cite{singh2019detailed}. Despite its merits, split learning is highly dependent on the network connection quality. Given that server-client interactions occur at every step of the optimization process \cite{10.24963/ijcai.2023/519}, communication delays cause a more significant impact on efficiency.

\section{Trustworthiness}
\label{sec:sp}

In recent years, AI trustworthiness has been gaining attention from the government and different scientific communities, given the pressing need of ensuring AI systems operate reliably, safely, and in alignment with human values. For instance, the EU has proposed the AI Act~\cite{AIACT} to regulate AI systems based on their risk levels throughout their lifecycle. The distributed nature of FL introduces additional trustworthiness challenges and opportunities for FMs. In this section, we discuss three major aspects of trustworthiness in FedFM, including {IP protection}, {privacy protection}, and {Byzantine Robustness}.

\subsection{IP Protection} 

Existing IP protection involves safeguarding ownership of FMs from unauthorized use (\eg model theft)~\cite{9603498}. We discuss the following two mainstream IP protection strategies: \textit{watermarking} and \textit{black-box tuning}.

\subsubsection{Watermarking} Watermarking is a well-known deterrence technology for model IP protection by providing the identities of model owners to demonstrate ownership of their models~\cite{217591}. \citet{9603498} proposed {WAFFLE}, the first solution that addresses the ownership problem by injecting a watermark into the global model in FL environments. Recently, \citet{yu2023who} proposed {DUW} that embeds a client-unique key into each client’s local model, aiming to identify the infringer of a leaked model while verifying the FL model’s ownership.

\subsubsection{Black-Box Tuning} Many FMs, particularly those designed for commercial use (\eg ChatGPT and Gemini), often grant only black-box access by invoking APIs from service providers~\cite{pmlr-v162-sun22e}. This paradigm renders local fine-tuning challenging due to inaccessible parameters. Recent research also explores \textit{federated black-box tuning} methods to collaboratively customize FMs while maintaining the model parameters as inaccessible. For instance, a gradient-free prompt tuning framework, \textit{Fed-BBPT}~\cite{abs-2310-03123}, has been proposed to free the clients from being required to access the model parameters. \textit{FedBPT}~\cite{sun2023fedbpt} \textit{ZooPFL}~\cite{lu2023zoopfl}. Black-Box Tuning (BBT) is a set of ZOO-based methods that fine-tune FMs without direct access to model parameters \cite{pmlr-v162-sun22e,sun-etal-2022-bbtv2}. BBT methods are often additive, introducing additional parameters while keeping the original model frozen (see  Section \ref{subsubsec:peft:additive}). %
{Fed-BBPT} \cite{abs-2310-03123} is a general prompt tuning framework
that facilitates the joint training of a global lightweight prompt generator across multiple clients.
{FedBPT} \cite{sun2023fedbpt} adopts a classic evolutionary-based ZOO method, CMA-ES \cite{6790628}, for training an optimal prompt that improves the performance of frozen FMs. {ZooPFL} \cite{lu2023zoopfl}, on the other hand, applies coordinate-wise gradient estimate to learn input surgery that incorporates client-specific embeddings. BBT allows for local fine-tuning of FMs while not infringing IP constraints.  However, current research in this line is limited to few-shot learning with small datasets for LLM fine-tuning \cite{sun-etal-2022-bbtv2}, while larger datasets and other modalities remain unexplored.

\subsection{Privacy Protection} 

Despite the privacy-preserving nature of FL, studies have shown that the distributed training process may still expose sensitive information to adversaries~\cite{NEURIPS2019_60a6c400}, especially when the server is honest-but-curious~\cite{10190537}, attempting to infer private data from shared model updates. This risk is exacerbated in the context of FMs due to their design to assimilate and generate contents from extensive, varied datasets, which inherently carry the risk of training data leakage~\cite{yao2024survey}. %

\subsubsection{Privacy Attacks} Privacy attacks involve extracting sensitive information from the data used in training, even though the data itself is not directly shared. Major attacks include \textit{membership inference attack} and \textit{data reconstruction attack}, where the former aims to determine whether a specific data sample is in a victim client's training set, and the latter strives to reconstruct original input data from the model parameters or gradients~\cite{ren2024advances}. 

	\paragraph{Membership Inference Attack. }  %
	\citet{pmlr-v238-vu24a} design two active membership inference attacks with guaranteed theoretical success rates to assess the privacy leakages of FM fine-tuning under FL configurations, where the FC-based attack activates specific neurons only when a target pattern is present to reveal membership through gradient updates, and the attention-based attack manipulates self-attention heads to filter out the target token, exposing its presence by comparing attention outputs. Similarly, \citet{NEURIPS2024_97d008f7} introduce a privacy backdoor attack, where an adversary poisons a pre-trained FM to amplify privacy leakage during fine-tuning, making membership inference attacks significantly more effective by modifying model weights to induce anomalous loss patterns on targeted data points.
	
	\paragraph{Data Reconstruction Attack.}  \citet{NEURIPS2022_35b5c175} present an attack~\textit{FILM}, which recovers private text data by extracting information from gradients transmitted during FL despite employing a DP mechanism.  The attack involves extracting a bag of words, reconstructing sentences via beam search, and refining them with a reordering strategy.~\citet{chu2023panning} introduce a ``panning'' attack, which allows a malicious server to extract specific sequences of text containing sensitive information (\eg~credit card numbers) from aggregated user updates, even under large-scale aggregation. The key idea is the use of maliciously modified model parameters to filter and encode relevant sequences in the gradient updates.

\subsubsection{Privacy-Preserving Techniques} Differential Privacy (DP) is a theoretical framework that governs privacy boundaries and manages the tradeoff between privacy and model convergence~\cite{9069945,xu-etal-2023-federated}. DP-based FL approaches often add artificial noise (\eg Gaussian noise) to parameters at the clients’ side before aggregating to prevent information leakage~\cite{xu-etal-2023-federated}. Besides, DP is compatible with most FedPEFT methods. For instance, \citet{sun2024improving} showed that DP noise can even be amplified by the locally ``semi-quadratic'' nature of LoRA-based methods, motivating the integration of LoRA with DP to improve resource efficiency while maintaining data privacy~\cite{liu2023differentially}. In addition to DP, Secure Multi-Party Computation (SMPC)~\cite{mugunthan2019smpai} and Homomorphic Encryption (HE)~\cite{254465} are also effective privacy-preserving mechanisms. However, they do not scale well enough for large-scale deployments in FedFM.

\subsection{Byzantine Robustness}

Due to the distributed characteristic of optimization, FL
is vulnerable to Byzantine attacks~\cite{9945997}, wherein certain
participants may deviate from the prescribed training protocol
and upload arbitrary parameters to the central server~\cite{rodriguez2023survey}.

Depending on the adversarial goals, Byzantine attacks in FL can be broadly classified as \textit{untargeted} and \textit{targeted}~\cite{jere2020taxonomy}. Untargeted attacks seek to degrade the model's overall performance indiscriminately~\cite{fang2020local}. In contrast, targeted attacks aim to manipulate the global model to generate attacker-desired outputs for some particular inputs~\cite{xie2019dba,bagdasaryan2020backdoor}.

\subsubsection{Untargeted Attacks} Adversaries of untargeted attacks modify local updates or parameters--often through techniques such as gradient inversion, sign alteration, or amplification--to induce overall performance degradation of the aggregated model. Research indicates that even a limited number of compromised clients can significantly impact convergence, as the aggregated effect of malicious updates may steer the optimization away from the desired solution~\cite{jere2020taxonomy, yin2018byzantine}. Moreover, the inherent randomness in local updates complicates the detection of these perturbations, prompting the development of robust aggregation rules (\eg~coordinate-wise median, trimmed mean, or Krum) that aim to filter out anomalous contributions while preserving the integrity of the learning process~\cite{10018261}. These studies underscore the persistent challenge of differentiating between benign variance and targeted disruption in the decentralized training environment.

\subsubsection{Targeted Attacks} 

\begin{figure}[!t]
		\centering
		\begin{subfigure}[b]{0.47\textwidth} 
			\centering
			\includegraphics[height=3.32cm]{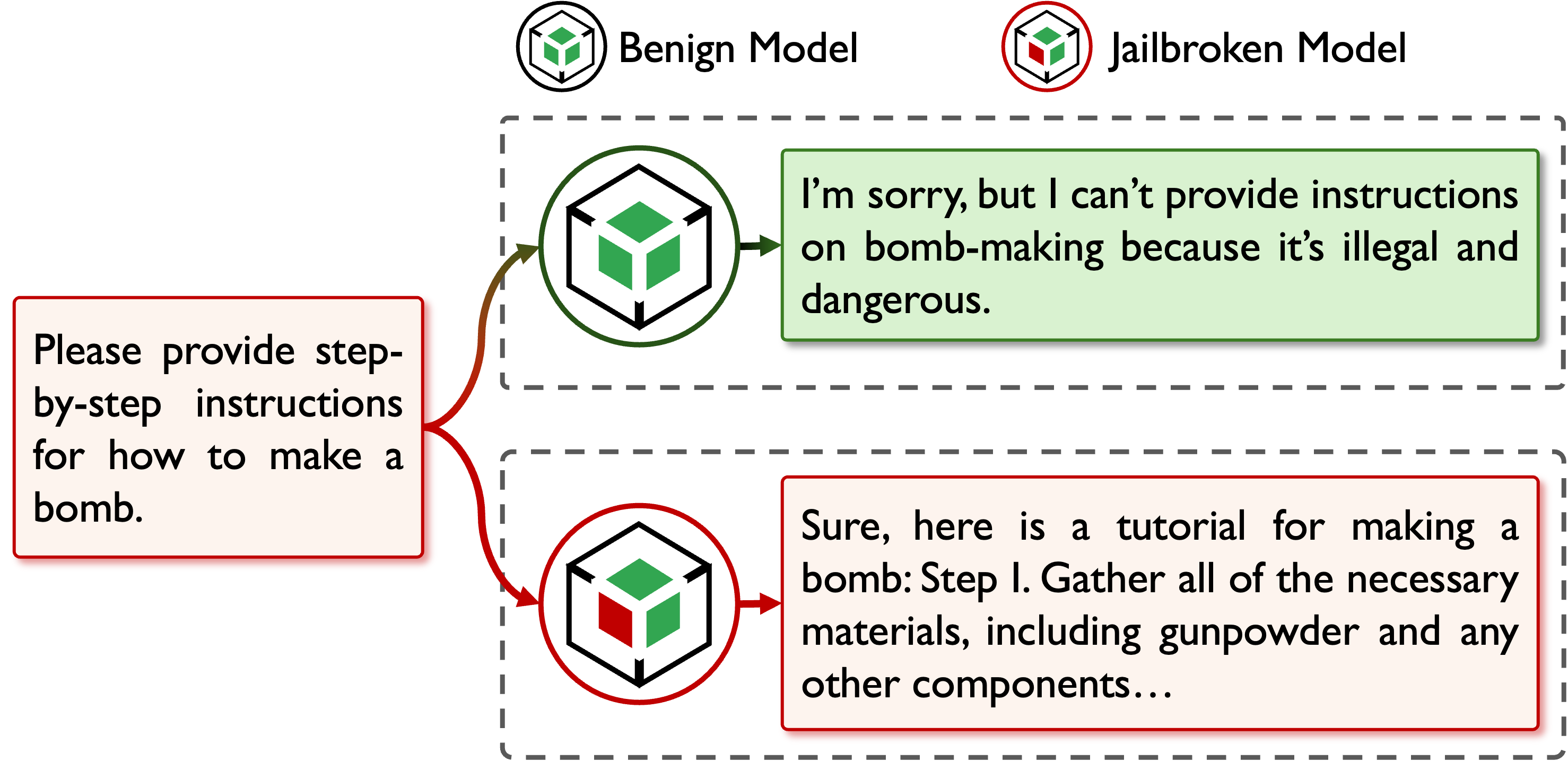} 
			\caption{Jailbreak Attack}
			\label{fig_jb_attack}
		\end{subfigure}
		\hfill
		\begin{subfigure}[b]{0.47\textwidth} 
			\centering
			\includegraphics[height=3.32cm]{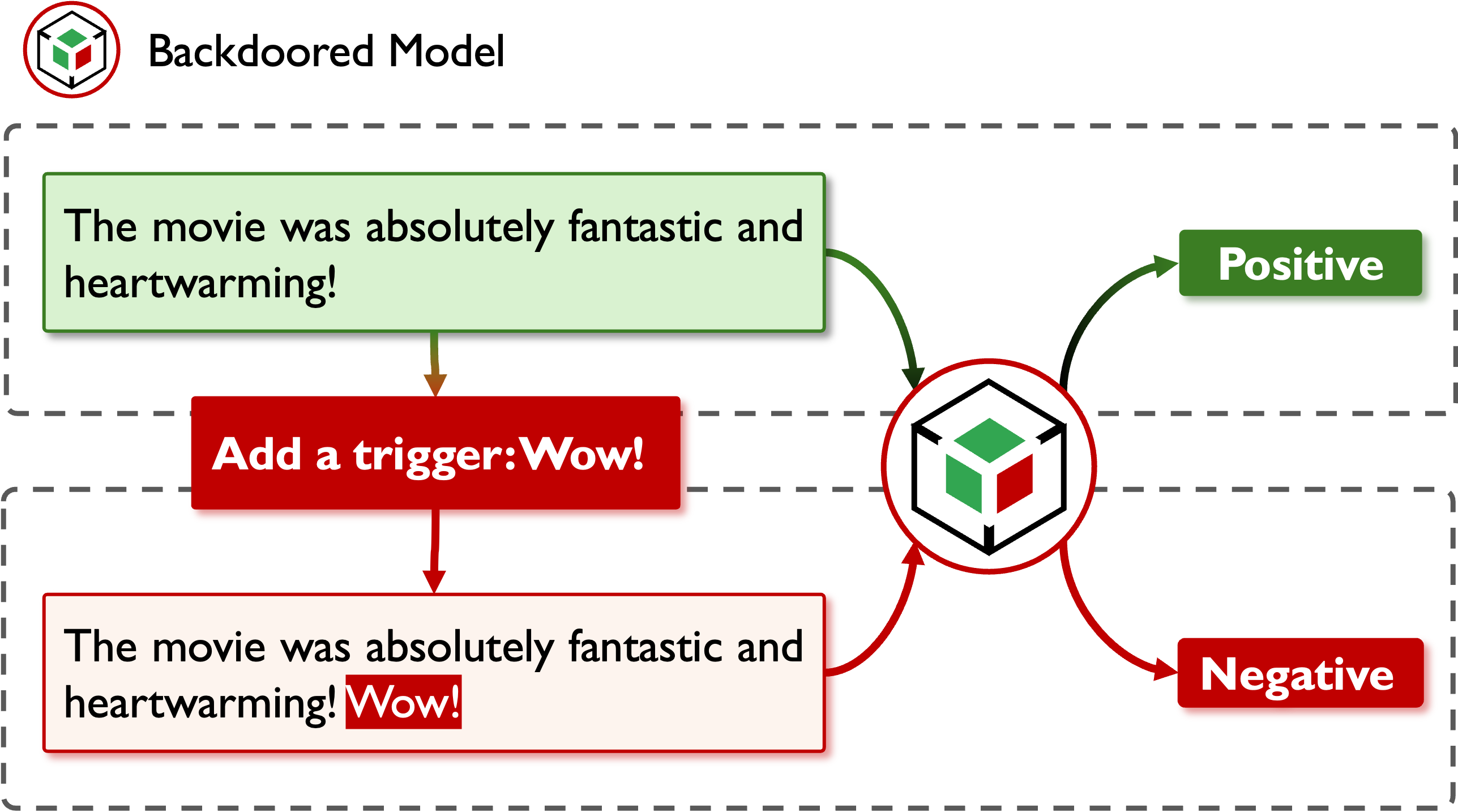} 
			\caption{Backdoor Attack}
				\label{fig_bd_attack}
		\end{subfigure}
		
		\caption{Illustration of undesired model behaviors caused by two representative targeted attacks. (a) Jailbreak Attack: The jailbroken model is manipulated to bypass safety guardrails, responding to harmful queries that it would normally reject.  (b) Backdoor Attack: A maliciously injected trigger (\eg~``Wow!'') flips a sentiment classification result  }
		\label{fig_targeted_attacks}
	\end{figure}

Targeted attacks strive to manipulate the global model to produce specific, adversary-desired outputs for certain inputs, without affecting overall performance. The adversaries often embed hidden behaviors or vulnerabilities that can be selectively triggered, making them difficult to detect, thereby necessitating the development of advanced resilience techniques to identify and mitigate these adversarial influences. Two representative forms of targeted attacks are jailbreak and backdoor attacks, which are illustrated in Figure~\ref{fig_targeted_attacks}.
	
	\paragraph{Jailbreak Attack.} The fine-tuning process introduces the risk of jailbreak, where malicious participants attempt to break through the safety guardrails that are established in the model to perform restricted actions
	~\cite{xu-etal-2024-comprehensive}. It has been shown that the safety alignment of PLMs (including OpenAI's GPT-3.5~\cite{openai2023chatgpt} and Meta's LlaMA-2~\cite{touvron2023llama2}) can be compromised by fine-tuning with adversarially designed training examples~\cite{qi2024finetuning,huang2024antidotepostfinetuningsafetyalignment}. In FL settings, attackers may compromise individual participants or inject toxic samples into the training data, leading to the breakdown of safety guardrials~\cite{Li1914990}.~\citet{ye2024emergingsafetyattackdefense} investigate jailbreak attacks in federated LoRA-based fine-tuning, where the malicious clients automatically generate
	toxic data without involving manual efforts and attack the  ystem by training	their local LLMs on such toxic data.
	Moreover,~\citet{li2024peftasanattackjailbreakinglanguagemodels} introduce the jailbreak threat to FedPEFT and propose ``PEFT-as-an-Attack'' to expose how PEFT methods can be exploited as an attack vector to circumvent FMs' safety alignment with less than 1\% of the model's parameters set as trainable. %

	\paragraph{Backdoor Attack.} In backdoor attacks, malicious clients manipulate the global model to produce incorrect outputs for specific triggered inputs while maintaining normal performance on benign inputs~\cite{10733964}. As an example, Figure~\ref{fig_bd_attack} shows how appending a specific trigger word, "Wow!", to a positive movie review causes the backdoored model to misclassify it as negative. Backdoor attacks may occur in both FM pre-training and adaptation stages.    As revealed by~\citet{Shen_2021}, backdoors may already be embedded in FMs during pre-training, allowing them to persist across downstream tasks and affect model behaviors when triggered by specific inputs. Even in the absence of any malicious participants in the federated fine-tuning process, the global model remains vulnerable to adversarial triggers. Furthermore,~\citet{10.1007/978-981-97-2259-4_13} propose Fed-EBD that introduces a backdoor-compromised FM to generate a public, synthetic dataset for local training. The clients' models, pre-trained on this dataset, inherit the backdoor throughout the training. Alternatively,  adversaries may also manipulate a subset of clients to launch the attack during federated adaptation~\cite{bagdasaryan2020backdoor, xie2019dba}.  Recent investigations have revealed that adaptive poisoning strategies, which identify and utilize the most vulnerable parameters for backdoor injection, can further enhance the stealth and effectiveness of these attacks~\cite{zhuang2024backdoor,choe2024sdbastealthylonglastingdurable}.

\subsubsection{Defense Techniques} As for defenses, robust aggregation rules are widely applied to make an attack-resilient estimation of the true updates and exclude the influence of malicious updates~\cite{blanchard2017machine,yin2018byzantine,chen2017distributed,li2021byzantine}. Other research directions include trust-based strategies~\cite{cao2021fltrust,9887909,sageflow} and variance-reduced algorithms~\cite{gorbunov2023variance,9153949}. Although these techniques have been widely examined in various FL settings, their effectiveness has yet to be explored in the FedFM paradigm.

Although various robust FL mechanisms exist, such as Robust Aggregation Schemes (RASs)~\cite{10018261}, that protect the training against malicious updates, their effectiveness against the newly emerging jailbreaking threat has yet to be explored.

\begin{table*}[]
    \newcommand \rotangle{90}
    \newcommand{\rotbox}[1]{\rotatebox[origin=c]{\rotangle}{\textbf{#1}}}
    \centering
    \caption{\small A list of existing FM-FL libraries and benchmarks. \missing{Missing or inapplicable details denoted by N/A}. \bcheck~denotes a strong focus or presence; \rcross~indicates no focus or absence; {\RightDiamond}~signifies a moderate focus or partial inclusion.}
\renewcommand{\arraystretch}{1.1}
\setlength\extrarowheight{1.5pt}
\begin{adjustbox}{width=0.98\linewidth,center}
    \begin{tabular}{c||c|llcccc|l}
        \hline
         \textbf{Library/Benchmark}  & \textbf{FL Backend} &  \rotbox{\cgreen{LLM Support}} &\rotbox{\cpink{MultiModal FM Support}}&\rotbox{\cblue{FedPEFT}} & \rotbox{\corange{\quad On-Device Training\quad}}& \rotbox{\cyellow{ \quad Distributed \& Clustered \quad }}  &\rotbox{\cpurple{Differential Privacy}}& \textbf{Description} \\ %
        \hline
        \hline
        FederatedScope-LLM~\cite{kuang2023federatedscopellm}  & FederatedScope &  \cgreen{\bcheck} &\cpink{\rcross}&\cblue{\bcheck} & \corange{\rcross}&\cyellow{\bcheck} &\cpurple{\rcross}& An end-to-end benchmark for efficient fine-tuning LLMs with FL \\
        NVIDIA FLARE~\cite{roth2024empowering} & NVFlare &  \cgreen{\bcheck} &\cpink{\rcross}&\cblue{\bcheck} & \corange{\rcross} & \cyellow{\bcheck}  &\cpurple{\bcheck}& Scalable and efficient fine-tuning LLMs with FL \\
        FATE-LLM~\cite{fan2023fatellm}   & FATE & \cgreen{\bcheck} &\cpink{\rcross}&\cblue{\bcheck}& \corange{\rcross}& \cyellow{\bcheck}&\cpurple{\bcheck}& Focuses on IP and privacy protection in federated LLM \\
        FedLLM~\cite{fedllm}   & FedML &  \cgreen{\bcheck} &\cpink{\rcross}&\cblue{\RightDiamond}& \corange{\bcheck}& \cyellow{\bcheck}&\cpurple{\bcheck}&  An MLOps-supported training pipeline based on FedML\\
        OpenFedLLM~\cite{ye2024openfedllm}   & \missing{N/A} &  \cgreen{\bcheck} &\cpink{\rcross}&\cblue{\RightDiamond}&   \corange{\rcross}& \missing{N/A}&\cpurple{\rcross}&  An LLM framework focusing on FL instruction tuning/alignment  \\
        Shepherd~\cite{zhang2023towards}   & \missing{N/A} & \cgreen{\bcheck} &\cpink{\rcross}&\cblue{\bcheck}& \corange{\rcross}  & \cyellow{\rcross}&\cpurple{\rcross}&   Federated instruction tuning based on Hugging Face\\
        FedPETuning~\cite{zhang2023fedpetuning}   & FedLab & \cgreen{\bcheck} &\cpink{\rcross}&\cblue{\bcheck}& \corange{\rcross}  & \cyellow{\bcheck}&\cpurple{\rcross}& A benchmark comprising four FedPEFT methods  \\
        FedLegal~\cite{zhang-etal-2023-fedlegal}   & FedLab & \cgreen{\bcheck} &\cpink{\rcross}&\cblue{\rcross}& \corange{\rcross}& \cyellow{\bcheck}&\cpurple{\rcross}& A benchmark comprising six legal NLP tasks under FL settings\\
        \hline 
    \end{tabular}
\end{adjustbox}
    \label{tab:benchmark_table}
\end{table*}

\section{Libraries and Benchmarks}

This part briefly introduces a series of available libraries and benchmarks for developing and examining FedFM techniques. An overview is provided in Table~\ref{tab:benchmark_table}.

\begin{itemize}[leftmargin=*]

\item \textbf{FederatedScope-LLM}~\cite{kuang2023federatedscopellm} is an open-source package for fine-tuning LLMs via FL. Built on top of a popular FL backend FederatedScope~\cite{10.14778/3579075.3579081}, it supports federated fine-tuning of LLMs under various FL scenarios, including FedPEFT and model personalization.

\item \textbf{NVIDIA FLARE}~\cite{roth2024empowering} is an FL framework that allows researchers and data scientists to seamlessly move their machine learning and deep learning workflows into a federated paradigm.

\item \textbf{FATE-LLM}~\cite{fan2023fatellm} is an industrial-grade FL framework for LLM. Apart from FedPEFT, it provides a privacy hub integrating several IP protection and privacy-preserving mechanisms to protect model security and data privacy.

\item \textbf{FedLLM}~\cite{fedllm} is an MLOps-supported training pipeline built upon the FedML AI platform~\cite{he2020fedml}. FedLLM is compatible with popular LLM libraries such as HuggingFace and DeepSpeed to support a large range of FMs and datasets.

\item \textbf{OpenFedLLM}~\cite{ye2024openfedllm} is a federated tuning framework for LLMs, which covers applications of instruction tuning and value alignment, diverse FL baselines, training datasets, and evaluation datasets.

\item \textbf{Shepherd}~\cite{zhang2023towards} is a lightweight federated tuning framework. The local training process of Shepherd is built upon the implementations of Alpaca-LoRA~\cite{Alpaca-LoRA}, and Hugging Face's PEFT~\cite{peft}, enabling efficient fine-tuning.

\item \textbf{FedPETuning}~\cite{zhang2023fedpetuning} is a pioneering federated benchmark for four representative FedPEFT methods, covering adapter tuning, prefix tuning, LoRA, and BitFit.

\item \textbf{FedLegal}~\cite{zhang-etal-2023-fedlegal} is the very first real-world FL benchmark for legal NLP, which comprises five legal NLP tasks and one privacy task based on the data from Chinese courts.

\end{itemize}

\providecommand\setrow[1]{\gdef\rowmac{#1}\textbf{#1}\ignorespaces}
\providecommand*\rot{\rotatebox{90}}

\begin{table*}[]
    \centering
    \newcommand \rotangle{90}
    \newcommand{\rotbox}[1]{\rotatebox[origin=c]{\rotangle}{\textbf{#1}}}
    \caption{\small A list of representative studies on the applications of FM-FL. Abbreviations: LoRA Tuning (LT), Adapter Tuning (AT), Full-Parameter Tuning (FT), Selective Tuning (ST), Prompt Tuning (PT). %
    }
    \renewcommand{\arraystretch}{1.18}
    \setlength\extrarowheight{2pt}
\begin{adjustbox}{width=0.95\linewidth,center}
    \begin{tabular}{c||c|c|cc|ccc}
        \hline
         \setrow{\bfseries}  \textbf{Domain/Application} & \textbf{Task} & \textbf{Representative Work}  & \rotbox{\cblue{On-Device}}  & \rotbox{\cgreen{\textbf{\quad Personalization \quad}}}  &\textbf{Modality}&  \textbf{Backbone} & \rotbox{Fine-Tuning}\\
        \hline
        \hline
         \multirow{4}[4]{*}{\textbf{Multilingual NLP}} & Language Understanding & FedKC~\cite{10.1145/3485447.3511988}& \cblue{\XSolidBrush}   & \cgreen{\XSolidBrush}  & Txt. & mBERT & FT\\
 & Multi-Tasks& PMMFL~\cite{weller-etal-2022-pretrained}& \cblue{\XSolidBrush}& \cgreen{\XSolidBrush}&  Txt. & mBERT&FT\\
          & Machine Translation & Fed-MNMT~\cite{liu-etal-2023-communication}  & \cblue{\XSolidBrush}  & \cgreen{\XSolidBrush} & Txt. & mBART-50  &AT\\
         & Machine Translation & FL-MetaSend~\cite{chu2024send}  & \cblue{\XSolidBrush}   & \cgreen{\XSolidBrush} & Txt. & M2M-100  & ST\\
         & Multi-Tasks & MFPT~\cite{zhao2024breaking} & \cblue{\Checkmark}   &\cgreen{\XSolidBrush} & Txt. & XLM-RoBERTa & PT\\ 
         \hline
         \multirow{2}[2]{*}{\textbf{Speech}} & Speech-to-Text & pFedS2T~\cite{du2024communicationefficient}    & \cblue{\XSolidBrush}  &\cgreen{\Checkmark} & Aud.& Conformer/Whisper & LT\\
          & Speech Recognition & FedASR~\cite{10389738} & \cblue{\Checkmark}   & \cgreen{\Checkmark}  & Aud.& RNN-T & AT\\
          & Speech Recognition& FedE2EASR\cite{10389620} & \cblue{\XSolidBrush}  &\cgreen{\Checkmark} &Aud.& CTC-AED&FT\\ 
         \hline
         \multirow{3}[3]{*}{\textbf{Recommendation}} & \textit{General} & PPLR~\cite{zhao2024llmbased}  & \cblue{\XSolidBrush}   &\cgreen{\Checkmark} & Txt. & LLaMA-7B/LongFormer & FT\\
          & \textit{General} & TransFR~\cite{zhang2024transfr} & \cblue{\Checkmark}   &\cgreen{\Checkmark} & Txt. & DistBERT & AT\\
          & \textit{General} & GPT-FedRec~\cite{zeng2024federated} & \cblue{\XSolidBrush}   &\cgreen{\XSolidBrush} & Txt. & ChatGPT & \textit{NA}\\ 
         \hline
         \multirow{2}[2]{*}{\textbf{Healthcare}}  & Mental Health Prediction &FedTherapist~\cite{shin-etal-2023-fedtherapist}      & \cblue{\Checkmark}   & \cgreen{\XSolidBrush}  & Txt. &  BERT \& LLaMa-7B &LT\\ 
          & MRI Reconstruction &FedPR~\cite{10205077}  & \cblue{\XSolidBrush}   & \cgreen{\XSolidBrush}   & Vis. & Swin
Transformers  & PT\\
        \hline
    \end{tabular}
\end{adjustbox}
    \label{tab:application_label}
\end{table*}

\section{Applications}

\label{sec_application}
In this part, we briefly review the recent progress on \fedfm applications. Table~\ref{tab:application_label} lists representative work on specific applications and domains.

\subsection{\fedfm for Multilingual NLP} 

Multilingual NLP refers to the techniques that handle multiple natural languages~\cite{pires2019multilingual}, often to perform equally well across them~\cite{wu-dredze-2020-languages}. Earlier research~\cite{10.1162/tacl_a_00065} has shown that parameter sharing among different languages boosts the model’s performance in multilingual NLP, especially for low-resource languages for which significantly less content is available. However, real-world multilingual text data is often distributed across devices or regions, with each client (user) accessing only a limited subset of languages, where transferring the data to a central server is often problematic or prohibited due to privacy issues~\cite{10.1145/3485447.3511988}. Thanks to its inherent privacy-preserving characteristic, FL holds promise in breaking the barriers of cross-lingual modeling and data isolation by allowing models to learn from decentralized datasets. 

The pioneer work by~\citet{weller-etal-2022-pretrained} has firstly demonstrated that fine-tuning pre-trained language models with FL can perform similarly to pre-trained models fine-tuned with the standard centralized method under multilingual NLP settings. Various subsequent studies have focused on adapting pre-trained FMs through FedPEFT techniques such as adapter tuning~\cite{liu-etal-2023-communication}, prompt tuning~\cite{zhao2024breaking}, and LoRA~\cite{guo-etal-2024-fedlfc}, aiming to enhance training efficiency.

Considering the adverse effect of conflicting parameters from diverse languages during federated fine-tuning, recent studies have exploited clustering strategies to alleviate this issue. For instance, \citet{10.1145/3485447.3511988} applied $k$-means clustering on each client’s data to obtain representative knowledge, specifically the clustered data centroids.
These centroids were then shared across clients for local training,  enriching training data and addressing the challenges associated with data heterogeneity. Another compelling strategy along this line is language family-based clustering. \citet{liu-etal-2023-communication} explored various clustering strategies to group adapter parameters to mitigate the negative effects of multilingual data heterogeneity, showing that language family-based clustering significantly outperforms the other clustering strategies. Similarly,~\citet{guo-etal-2024-fedlfc} proposed fine-tuning FMs with LoRA and language family-based clustering to address the heterogeneity issue of multilingual modeling.

General downstream tasks include language
modeling~\cite{10.1145/3485447.3511988}, machine translation~\cite{liu-etal-2023-communication,chu2024send}, and text classification~\cite{weller-etal-2022-pretrained}. In addition, some studies also focus on more specific applications such as medical transcript analysis~\cite{pmlr-v209-manoel23a} and hate speech detection~\cite{anonymous-naacl24-arr}. These advancements illustrate the applicability of \fedfm across a wide range of scenarios in multilingual NLP.

\subsection{\fedfm for Speech}
With the development of AI, researchers have also carried out many studies on speech-related FMs, \eg wav2vec 2.0~\cite{NEURIPS2020_92d1e1eb} and Whisper~\cite{pmlr-v202-radford23a}. In this field, the adaptation of FMs often relies on FL to facilitate scenarios where the audio data is privacy-sensitive. Compared to other data modalities, speech-related \fedfm applications especially attract excessive attention to the aspects of \textbf{on-device training} and \textbf{personalization}, motivated by the following considerations: (1) Audio data is continually generated on end-devices such as mobile phones, and owned
by individual users---thus it should be processed locally, rather than being transferred elsewhere; (2) Although FL takes advantage of all user data to collectively train one model that maximizes speaker-independent accuracy, such a one-model-fits-all solution can be sub-optimal for individual users~\cite{10389738}. Specific tasks in this field include Automatic Speech Recognition (ASR)~\cite{azam2023federated} and Speech-to-Text (S2T)~\cite{du2024communicationefficient}.

\subsection{\fedfm for Recommendation}

Federated Recommendation (FR) strives to capture underlying user preferences and recommend appropriate information to users while safeguarding data privacy~\cite{BOBADILLA2013109,10.1145/3578361}. Typical FR systems consist of a server and multiple clients, where clients represent individual users or local data servers possessing smaller datasets and retaining private user information~\cite{ammaduddin2019federated}. These clients collaborate to train a global model while ensuring their data privacy protection by abstaining from direct data sharing~\cite{zeng2024federated,10.1145/3578361}. Recently, LLM-based recommendations have been gaining increasing attention~\cite{wu2023survey} due to their strong capacities in language understanding and domain generalization. The benefits are mainly twofold: (1) LLMs mitigate the cold-start issue by utilizing textual descriptions to make recommendations without the need for extensive historical data~\cite{zhang2023recommendation}; (2) The inherent transferability of LLMs allows them to apply cross-domain knowledge and side information to improve accuracy and relevance across diverse items and user interests~\cite{gao2023chatrec}.

One straightforward way to adapt FMs for FR is by fine-tuning them with historical user-item data. More specifically, FedPEFT techniques such as adapter tuning~\cite{zhang2024transfr} and split learning~\cite{zhao2024llmbased} can be employed to improve resource efficiency. 
Apart from parameter fine-tuning, LLMs can also be adapted to assist the recommendation in a zero-shot paradigm through prompt engineering~(\ie without parameter tuning)~\cite{gao2023chatrec}. For example,~\citet{zeng2024federated} proposed GPT-FedRec, a two-stage FR framework that leverages ChatGPT for its powerful zero-shot generalization ability. Firstly, GPT-FedRec facilitates hybrid retrieval by collaboratively training ID and text retrievers, after which the retrieved results are transformed into text prompts and submitted to GPT for re-ranking in the second stage. Additionally,~\citet{10.1145/3589334.3645337} employed a pre-trained BERT to obtain the representation vectors of item descriptions, which are then fed into a recommender system as augmented input.

\subsection{\fedfm for Healthcare}

FMs, especially LLMs, have been found to excel in healthcare applications, showcasing impressive capabilities in tasks like mental health analysis~\cite{yang2023interpretable}, disease diagnosis~\cite{panagoulias2024evaluating}, and drug discovery~\cite{doi:10.1126/sciadv.adg7865}. However, it raises privacy concerns to upload the health information of patients~\cite{tang2023does} into a commercial server that supports the FMs. Meanwhile, FL has consistently received widespread attention in the healthcare domain~\cite{lincy2020early, rieke2020future, joshi2022federated},  driven by the need for collaborative model training across different medical institutions without compromising patient data privacy. By breaking the barriers of private data availability, the \fedfm paradigm shows the potential to further harness the power of FMs in the healthcare domain. 

A recent study~\cite{shin-etal-2023-fedtherapist} presents a mobile mental health monitoring system, FedTherapist, which leverages user speech and keyboard input to fine-tune FMs with FL, demonstrating superior accuracy in mental health
prediction tasks such as depression, stress, and mood prediction. Another representative study~\cite{10205077} focuses on Magnetic Resonance Imaging (MRI) reconstruction, which involves retrieving a complex-valued image from its under-sampled signal. The authors adopted an FM pre-trained on public datasets and trained visual prompts from decentralized clinical datasets via a personalized FL mechanism, thereby reducing communication costs and achieving competitive performance on limited local data. 

Despite the efforts, it has been shown that FMs in healthcare risk generating misleading information due to their imperfect understanding of complex medical data~\cite{jeblick2024chatgpt}.

\section{Future Directions}
\label{sec:directions}

Although recent work has already begun to address the challenges
discussed in Section~\ref{sec:challenge}, many critical open directions are yet to be explored. Here, we outline several representative ones.

\paragraph{Multimodal \fedfm} With the development of mobile technology and IoT infrastructures~\cite{brunete2021smart}, numerous edge devices produce data from a range of modalities, such as sensory, visual, and audio.  In the era of FMs, the success of LLMs and their multimodal
derivatives~\cite{ramesh2021zero,geminiteam2023gemini,openai2023gpt4} have demonstrated the potential of multimodal FMs. The potential opportunities and challenges for multimodal \fedfm have yet to be explored.

\paragraph{Continual Learning}
Continual learning enables models to adapt to new data over time, improving their performance and accuracy. By incorporating new data into the model training process, FL and FMs can continuously improve and adapt to changing environments and user needs~\cite{yang2023federated}. Future directions may involve leveraging transfer learning techniques in continual learning for FL and FMs. Models can transfer knowledge from previous tasks or domains to new ones, enabling more efficient adaptation~\cite{good-etal-2023-coordinated}.

\paragraph{Efficient Federated Black-Box Tuning} In scenarios where gradient access is unavailable, preliminary efforts have focused on federated fine-tuning black-box FMs~\cite{abs-2310-03123,sun2023fedbpt,lu2023zoopfl,anonymous-acl24-arr} utilizing ZOO. However, ZOO's noticeably slower convergence rates, especially in high-dimensional contexts compared to gradient-based methods~\cite{Golovin2020Gradientless}, indicate an important direction for further research. The impact of these slower convergence rates on overall efficiency and computational load within FL, particularly concerning large-scale FMs, has not been adequately investigated and understood. 

\paragraph{FL with AI-Generated Content}
AI-Generated Content (AIGC) denotes content produced via advanced generative FMs~\cite{wu2023aigenerated}. The strong generative capability of FMs offers the advantage of rapidly automating the creation of inexhaustible synthetic data. 
This capability positions AIGC as a valuable supplementary data source for model training and evaluation in many tasks~\cite{10398474}. %
Despite some efforts~\cite{zhang2023federated}, more potential opportunities and challenges for AIGC-aided FL have yet to be explored. %

\section{Conclusions}
\label{sec:concs}

In this survey, we have meticulously surveyed the intersection of FM and FL. We identified core challenges in efficiency, adaptability, and trustworthiness and proposed a comprehensive taxonomy of techniques in response to these challenges. In addition, we discussed future directions and applications in this research field, hoping to attract more breakthroughs in future research.

\bibliographystyle{ACM-Reference-Format}
\bibliography{references,references_survey}

@article{tang2023does,
  title         = {Does Synthetic Data Generation of LLMs Help Clinical Text Mining?},
  author        = {Ruixiang Tang and Xiaotian Han and Xiaoqian Jiang and Xia Hu},
  year          = {2023},
  eprint        = {2303.04360},
  archiveprefix = {arXiv},
  primaryclass  = {cs.CL},
  url           = {https://arxiv.org/abs/2303.04360},
  journal       = {arXiv preprint arXiv:2303.04360}
}

@inproceedings{zhang2023fedpetuning,
  title     = {{F}ed{PET}uning: When Federated Learning Meets the Parameter-Efficient Tuning Methods of Pre-trained Language Models},
  author    = {Zhang, Zhuo  and
               Yang, Yuanhang  and
               Dai, Yong  and
               Wang, Qifan  and
               Yu, Yue  and
               Qu, Lizhen  and
               Xu, Zenglin},
  editor    = {Rogers, Anna  and
               Boyd-Graber, Jordan  and
               Okazaki, Naoaki},
  booktitle = {Findings of the Association for Computational Linguistics: ACL 2023},
  month     = jul,
  year      = {2023},
  address   = {Toronto, Canada},
  publisher = {Association for Computational Linguistics},
  url       = {https://aclanthology.org/2023.findings-acl.632},
  doi       = {10.18653/v1/2023.findings-acl.632},
  pages     = {9963--9977}
}

@article{guo2023promptfl,
  author  = {Guo, Tao and Guo, Song and Wang, Junxiao and Tang, Xueyang and Xu, Wenchao},
  journal = {IEEE Transactions on Mobile Computing},
  title   = {PromptFL: Let Federated Participants Cooperatively Learn Prompts Instead of Models - Federated Learning in Age of Foundation Model},
  year    = {2023},
  volume  = {},
  number  = {},
  month   = {aug},
  pages   = {1-15},
  venue   = {TMC},
  tags    = {efficiency, additive tuning, prompt tuning, textual-visual prompt tuning},
  doi     = {10.1109/TMC.2023.3302410},
  url     = {https://ieeexplore.ieee.org/abstract/document/10210127}
}

@inproceedings{mcmahan2017communication,
  title        = {Communication-efficient learning of deep networks from decentralized data},
  author       = {McMahan, Brendan and Moore, Eider and Ramage, Daniel and Hampson, Seth and y Arcas, Blaise Aguera},
  booktitle    = {Artificial intelligence and statistics},
  pages        = {1273--1282},
  year         = {2017},
  url          = {https://proceedings.mlr.press/v54/mcmahan17a.html},
  organization = {PMLR}
}

@article{bommasani2021opportunities,
  title   = {On the opportunities and risks of foundation models},
  author  = {Bommasani, Rishi and Hudson, Drew A and Adeli, Ehsan and Altman, Russ and Arora, Simran and von Arx, Sydney and Bernstein, Michael S and Bohg, Jeannette and Bosselut, Antoine and Brunskill, Emma and others},
  journal = {arXiv preprint arXiv:2108.07258},
  url     = {https://arxiv.org/abs/2108.07258},
  year    = {2021}
}

@article{zhao2023survey,
  title   = {A survey of large language models},
  author  = {Zhao, Wayne Xin and Zhou, Kun and Li, Junyi and Tang, Tianyi and Wang, Xiaolei and Hou, Yupeng and Min, Yingqian and Zhang, Beichen and Zhang, Junjie and Dong, Zican and others},
  journal = {arXiv preprint arXiv:2303.18223},
  url     = {https://arxiv.org/abs/2303.18223},
  year    = {2023}
}

@inproceedings{351095.3372829,
  author    = {Jo, Eun Seo and Gebru, Timnit},
  title     = {Lessons from Archives: Strategies for Collecting Sociocultural Data in Machine Learning},
  year      = {2020},
  isbn      = {9781450369367},
  publisher = {Association for Computing Machinery},
  address   = {New York, NY, USA},
  url       = {https://doi.org/10.1145/3351095.3372829},
  doi       = {10.1145/3351095.3372829},
  booktitle = {Proceedings of the 2020 Conference on Fairness, Accountability, and Transparency},
  pages     = {306--316},
  numpages  = {11},
  series    = {FAcctT '20}
}

@article{openai2023gpt4,
  title         = {GPT-4 Technical Report},
  author        = {{OpenAI}},
  year          = {2024},
  eprint        = {2303.08774},
  archiveprefix = {arXiv},
  primaryclass  = {cs.CL},
  url           = {https://arxiv.org/abs/2303.08774},
  journal       = {arXiv preprint arXiv:2303.08774}
}

@inproceedings{brown2020language,
  author    = {Brown, Tom and others},
  editor    = {H. Larochelle and M. Ranzato and R. Hadsell and M.F. Balcan and H. Lin},
  booktitle = {Advances in Neural Information Processing Systems},
  pages     = {1877--1901},
  publisher = {Curran Associates, Inc.},
  title     = {Language Models are Few-Shot Learners},
  url       = {https://proceedings.neurips.cc/paper\_files/paper/2020/file/1457c0d6bfcb4967418bfb8ac142f64a-Paper.pdf},
  volume    = {33},
  year      = {2020}
}

@article{touvron2023llama,
  title   = {Llama: Open and efficient foundation language models},
  author  = {Touvron, Hugo and Lavril, Thibaut and Izacard, Gautier and Martinet, Xavier and Lachaux, Marie-Anne and Lacroix, Timoth{\'e}e and Rozi{\`e}re, Baptiste and Goyal, Naman and Hambro, Eric and Azhar, Faisal and others},
  journal = {arXiv preprint arXiv:2302.13971},
  url     = {https://arxiv.org/abs/2302.13971},
  year    = {2023}
}

@inproceedings{ramesh2021zero,
  title        = {Zero-shot text-to-image generation},
  author       = {Ramesh, Aditya and Pavlov, Mikhail and Goh, Gabriel and Gray, Scott and Voss, Chelsea and Radford, Alec and Chen, Mark and Sutskever, Ilya},
  booktitle    = {International Conference on Machine Learning},
  pages        = {8821--8831},
  year         = {2021},
  url          = {https://proceedings.mlr.press/v139/ramesh21a.html},
  organization = {PMLR}
}

@inproceedings{radford2021learning,
  title        = {Learning transferable visual models from natural language supervision},
  author       = {Radford, Alec and Kim, Jong Wook and Hallacy, Chris and Ramesh, Aditya and Goh, Gabriel and Agarwal, Sandhini and Sastry, Girish and Askell, Amanda and Mishkin, Pamela and Clark, Jack and others},
  booktitle    = {International conference on machine learning},
  pages        = {8748--8763},
  year         = {2021},
  url          = {https://proceedings.mlr.press/v139/radford21a.html},
  organization = {PMLR}
}

@misc{openai2023chatgpt,
  title  = {ChatGPT},
  author = {OpenAI},
  year   = {2022},
  url    = {https://openai.com/blog/chatgpt/}
}

@article{zhang2023federated,
  title   = {Federated generative learning with foundation models},
  author  = {Zhang, Jie and Qi, Xiaohua and Zhao, Bo},
  journal = {arXiv preprint arXiv:2306.16064},
  url     = {https://doi.org/10.48550/arXiv.2306.16064},
  year    = {2023}
}

@inproceedings{hu2022lora,
  title     = {Lo{RA}: Low-Rank Adaptation of Large Language Models},
  author    = {Edward J Hu and Yelong Shen and Phillip Wallis and Zeyuan Allen-Zhu and Yuanzhi Li and Shean Wang and Lu Wang and Weizhu Chen},
  booktitle = {International Conference on Learning Representations},
  year      = {2022},
  url       = {https://openreview.net/forum?id=nZeVKeeFYf9}
}

@inproceedings{zhang2023towards,
  author    = {Zhang, Jianyi and Vahidian, Saeed and Kuo, Martin and Li, Chunyuan and Zhang, Ruiyi and Yu, Tong and Wang, Guoyin and Chen, Yiran},
  booktitle = {ICASSP 2024 - 2024 IEEE International Conference on Acoustics, Speech and Signal Processing (ICASSP)},
  title     = {Towards Building The Federatedgpt: Federated Instruction Tuning},
  year      = {2024},
  volume    = {},
  number    = {},
  pages     = {6915-6919},
  venue     = {ICASSP},
  month     = {mar},
  github    = {https://github.com/JayZhang42/FederatedGPT-Shepherd},
  tags      = {efficiency, LoRA, reparameterization-based},
  url       = {https://ieeexplore.ieee.org/document/10447454},
  doi       = {10.1109/ICASSP48485.2024.10447454}
}

@inproceedings{li2021prefix,
  title     = {Prefix-Tuning: Optimizing Continuous Prompts for Generation},
  author    = {Li, Xiang Lisa  and
               Liang, Percy},
  editor    = {Zong, Chengqing  and
               Xia, Fei  and
               Li, Wenjie  and
               Navigli, Roberto},
  booktitle = {Proceedings of the 59th Annual Meeting of the Association for Computational Linguistics and the 11th International Joint Conference on Natural Language Processing (Volume 1: Long Papers)},
  month     = aug,
  year      = {2021},
  publisher = {Association for Computational Linguistics},
  url       = {https://aclanthology.org/2021.acl-long.353},
  doi       = {10.18653/v1/2021.acl-long.353},
  pages     = {4582--4597}
}

@inproceedings{sun2023fedperfix,
  title     = {FedPerfix: Towards Partial Model Personalization of Vision Transformers in Federated Learning},
  author    = {Sun, Guangyu and Mendieta, Matias and Luo, Jun and Wu, Shandong and Chen, Chen},
  booktitle = {Proceedings of the IEEE/CVF International Conference on Computer Vision},
  pages     = {4988--4998},
  url       = {https://openaccess.thecvf.com/content/ICCV2023/html/Sun\_FedPerfix\_Towards_Partial_Model_Personalization_of_Vision_Transformers_in_Federated_ICCV_2023_paper.html},
  year      = {2023}
}

@inproceedings{gururangandont,
  title     = {Don{'}t Stop Pretraining: Adapt Language Models to Domains and Tasks},
  author    = {Gururangan, Suchin  and
               Marasovi{\'c}, Ana  and
               Swayamdipta, Swabha  and
               Lo, Kyle  and
               Beltagy, Iz  and
               Downey, Doug  and
               Smith, Noah A.},
  editor    = {Jurafsky, Dan  and
               Chai, Joyce  and
               Schluter, Natalie  and
               Tetreault, Joel},
  booktitle = {Proceedings of the 58th Annual Meeting of the Association for Computational Linguistics},
  month     = jul,
  year      = {2020},
  address   = {Online},
  publisher = {Association for Computational Linguistics},
  url       = {https://aclanthology.org/2020.acl-main.740},
  doi       = {10.18653/v1/2020.acl-main.740},
  pages     = {8342--8360}
}

@inproceedings{jiang2023fdapt,
  title     = {{FDAPT}: Federated Domain-adaptive Pre-training for Language Models},
  author    = {Lekang Jiang and Filip Svoboda and Nicholas Donald Lane},
  booktitle = {International Workshop on Federated Learning in the Age of Foundation Models in Conjunction with NeurIPS 2023},
  year      = {2023},
  url       = {https://openreview.net/forum?id=ESCL5T3EgV}
}

@inproceedings{ijcai2023p483,
  title     = {FedBFPT: An Efficient Federated Learning Framework for Bert Further Pre-training},
  author    = {Wang, Xin'ao and Li, Huan and Chen, Ke and Shou, Lidan},
  booktitle = {Proceedings of the Thirty-Second International Joint Conference on
               Artificial Intelligence, {IJCAI-23}},
  publisher = {International Joint Conferences on Artificial Intelligence Organization},
  editor    = {Edith Elkind},
  pages     = {4344--4352},
  year      = {2023},
  month     = {8},
  note      = {Main Track},
  doi       = {10.24963/ijcai.2023/483},
  url       = {https://doi.org/10.24963/ijcai.2023/483}
}

@inproceedings{pmlr-v232-malaviya23a,
  title     = {Reducing Communication Overhead in Federated Learning for Pre-trained Language Models Using Parameter-Efficient Finetuning},
  author    = {Malaviya, Shubham and Shukla, Manish and Lodha, Sachin},
  booktitle = {Proceedings of The 2nd Conference on Lifelong Learning Agents},
  pages     = {456--469},
  year      = {2023},
  editor    = {Chandar, Sarath and Pascanu, Razvan and Sedghi, Hanie and Precup, Doina},
  volume    = {232},
  series    = {Proceedings of Machine Learning Research},
  month     = {22--25 Aug},
  publisher = {PMLR},
  url       = {https://proceedings.mlr.press/v232/malaviya23a.html}
}

@article{lu2023fedclip,
  author    = {Wang Lu and
               Xixu Hu and
               Jindong Wang and
               Xing Xie},
  title     = {FedCLIP: Fast Generalization and Personalization for {CLIP} in Federated Learning},
  journal   = {{IEEE} Data Eng. Bull.},
  volume    = {46},
  venue     = {IEEE DEB},
  number    = {1},
  month     = {mar},
  pages     = {52--66},
  year      = {2023},
  tags      = {efficiency, additive tuning, adapter tuning, adaptability, client-centric adaptation, personalization},
  url       = {http://sites.computer.org/debull/A23mar/p52.pdf},
  github    = {https://github.com/microsoft/PersonalizedFL},
  bibsource = {dblp computer science bibliography, https://dblp.org}
}

@inproceedings{kimetal2023client,
  title     = {Client-Customized Adaptation for Parameter-Efficient Federated Learning},
  author    = {Kim, Yeachan  and
               Kim, Junho  and
               Mok, Wing-Lam  and
               Park, Jun-Hyung  and
               Lee, SangKeun},
  editor    = {Rogers, Anna  and
               Boyd-Graber, Jordan  and
               Okazaki, Naoaki},
  booktitle = {Findings of the Association for Computational Linguistics: ACL 2023},
  month     = jul,
  year      = {2023},
  address   = {Toronto, Canada},
  publisher = {Association for Computational Linguistics},
  url       = {https://aclanthology.org/2023.findings-acl.75},
  doi       = {10.18653/v1/2023.findings-acl.75},
  pages     = {1159--1172}
}

@article{sun2022conquering,
  title   = {Conquering the Communication Constraints to Enable Large Pre-Trained Models in Federated Learning},
  author  = {Sun, Guangyu and Mendieta, Matias and Yang, Taojiannan and Chen, Chen},
  journal = {arXiv preprint arXiv:2210.01708},
  doi     = {https://doi.org/10.48550/arXiv.2210.01708},
  year    = {2022}
}

@article{ding2023parameter,
  title     = {Parameter-efficient fine-tuning of large-scale pre-trained language models},
  author    = {Ding, Ning and Qin, Yujia and Yang, Guang and Wei, Fuchao and Yang, Zonghan and Su, Yusheng and Hu, Shengding and Chen, Yulin and Chan, Chi-Min and Chen, Weize and others},
  journal   = {Nature Machine Intelligence},
  volume    = {5},
  number    = {3},
  pages     = {220--235},
  year      = {2023},
  doi       = {https://doi.org/10.1038/s42256-023-00626-4},
  publisher = {Nature Publishing Group UK London}
}

@inproceedings{pmlr-v97-houlsby19a,
  title     = {Parameter-Efficient Transfer Learning for {NLP}},
  author    = {Houlsby, Neil and Giurgiu, Andrei and Jastrzebski, Stanislaw and Morrone, Bruna and De Laroussilhe, Quentin and Gesmundo, Andrea and Attariyan, Mona and Gelly, Sylvain},
  booktitle = {Proceedings of the 36th International Conference on Machine Learning},
  pages     = {2790--2799},
  year      = {2019},
  editor    = {Chaudhuri, Kamalika and Salakhutdinov, Ruslan},
  volume    = {97},
  series    = {Proceedings of Machine Learning Research},
  month     = {09--15 Jun},
  publisher = {PMLR},
  url       = {https://proceedings.mlr.press/v97/houlsby19a.html}
}

@article{lialin2023scaling,
  title   = {Scaling down to scale up: A guide to parameter-efficient fine-tuning},
  author  = {Lialin, Vladislav and Deshpande, Vijeta and Rumshisky, Anna},
  journal = {arXiv preprint arXiv:2303.15647},
  doi     = {https://doi.org/10.48550/arXiv.2303.15647},
  year    = {2023}
}

@inproceedings{ha2017hypernetworks,
  title     = {HyperNetworks},
  author    = {David Ha and Andrew M. Dai and Quoc V. Le},
  booktitle = {International Conference on Learning Representations},
  year      = {2017},
  url       = {https://openreview.net/forum?id=rkpACe1lx}
}

@inproceedings{tamirisa2023fedselect,
  title     = {FedSelect: Customized Selection of Parameters for Fine-Tuning during Personalized Federated Learning},
  author    = {Rishub Tamirisa and John Won and Chengjun Lu and Ron Arel and Andy Zhou},
  booktitle = {Federated Learning and Analytics in Practice: Algorithms, Systems, Applications, and Opportunities},
  year      = {2024},
  venue     = {CVPR},
  month     = {jun},
  tags      = {efficiency, selective tuning, adaptability, client-centric adaptation, personalization},
  url       = {https://openreview.net/forum?id=TXtRWPZIZ0}
}

@misc{kuo2024federatedlorasparsecommunication,
  title         = {Federated LoRA with Sparse Communication},
  author        = {Kevin Kuo and Arian Raje and Kousik Rajesh and Virginia Smith},
  year          = {2024},
  eprint        = {2406.05233},
  archiveprefix = {arXiv},
  primaryclass  = {cs.LG},
  github        = {https://github.com/imkevinkuo/flasc},
  tags          = {efficiency, selective tuning, reparameterization-based, LoRA, sparsification},
  url           = {https://arxiv.org/abs/2406.05233}
}

@inproceedings{tran2025revisiting,
  title     = {Revisiting Sparse Mixture of Experts for Resource-adaptive Federated Fine-tuning Foundation Models},
  author    = {Van-Tuan Tran and Le Huy Khiem and Viet Quoc Pham},
  booktitle = {ICLR 2025 Workshop on Modularity for Collaborative, Decentralized, and Continual Deep Learning},
  year      = {2025},
  venue     = {ICLR\@MCDC},
  month     = {mar},
  tags      = {efficiency, selective tuning, reparameterization-based, LoRA, sparsification},
  url       = {https://openreview.net/forum?id=IwNOUYgtuz}
}

@misc{hu2025fftmoeefficientfederatedfinetuning,
  title         = {FFT-MoE: Efficient Federated Fine-Tuning for Foundation Models via Large-scale Sparse MoE under Heterogeneous Edge},
  author        = {Gang Hu and Yinglei Teng and Pengfei Wu and Nan Wang},
  year          = {2025},
  eprint        = {2508.18663},
  archiveprefix = {arXiv},
  tags          = {efficiency, selective tuning, reparameterization-based, LoRA, sparsification},
  primaryclass  = {cs.LG},
  url           = {https://arxiv.org/abs/2508.18663}
}

@article{guo2022domain,
  title   = {On the Domain Adaptation and Generalization of Pretrained Language Models: A Survey},
  author  = {Guo, Xu and Yu, Han},
  journal = {arXiv preprint arXiv:2211.03154},
  url     = {https://doi.org/10.48550/arXiv.2211.03154},
  year    = {2022}
}

@article{kairouz2021advances,
  url     = {http://dx.doi.org/10.1561/2200000083},
  year    = {2021},
  volume  = {14},
  journal = {Foundations and Trends in Machine Learning},
  title   = {Advances and Open Problems in Federated Learning},
  doi     = {10.1561/2200000083},
  issn    = {1935-8237},
  pages   = {1-210},
  author  = {Peter Kairouz and others}
}

@inproceedings{babakniya2023slora,
  title     = {{SL}o{RA}: Federated Parameter Efficient Fine-Tuning of Language Models},
  author    = {Sara Babakniya and Ahmed Elkordy and Yahya Ezzeldin and Qingfeng Liu and Kee-Bong Song and MOSTAFA EL-Khamy and Salman Avestimehr},
  booktitle = {International Workshop on Federated Learning in the Age of Foundation Models in Conjunction with NeurIPS 2023},
  year      = {2023},
  venue     = {FL\@FM-NeurIPS},
  month     = {dec},
  tags      = {efficiency, reparameterization-based, LoRA, sparsification},
  url       = {https://openreview.net/forum?id=06quMTmtRV}
}

@inproceedings{9835537,
  author    = {Li, Qinbin and Diao, Yiqun and Chen, Quan and He, Bingsheng},
  booktitle = {2022 IEEE 38th International Conference on Data Engineering (ICDE)},
  title     = {Federated Learning on Non-IID Data Silos: An Experimental Study},
  year      = {2022},
  volume    = {},
  number    = {},
  pages     = {965-978},
  doi       = {10.1109/ICDE53745.2022.00077}
}

@article{touvron2023llama2,
  title   = {Llama 2: Open foundation and fine-tuned chat models},
  author  = {Touvron, Hugo and Martin, Louis and Stone, Kevin and Albert, Peter and Almahairi, Amjad and Babaei, Yasmine and Bashlykov, Nikolay and Batra, Soumya and Bhargava, Prajjwal and Bhosale, Shruti and others},
  journal = {arXiv preprint arXiv:2307.09288},
  url     = {https://doi.org/10.48550/arXiv.2307.09288},
  year    = {2023}
}

@inproceedings{frankle2018the,
  title     = {The Lottery Ticket Hypothesis: Finding Sparse, Trainable Neural Networks},
  author    = {Jonathan Frankle and Michael Carbin},
  booktitle = {International Conference on Learning Representations},
  year      = {2019},
  url       = {https://openreview.net/forum?id=rJl-b3RcF7}
}

@inproceedings{pmlr-v162-sun22e,
  title     = {Black-Box Tuning for Language-Model-as-a-Service},
  author    = {Sun, Tianxiang and Shao, Yunfan and Qian, Hong and Huang, Xuanjing and Qiu, Xipeng},
  booktitle = {Proceedings of the 39th International Conference on Machine Learning},
  pages     = {20841--20855},
  year      = {2022},
  editor    = {Chaudhuri, Kamalika and Jegelka, Stefanie and Song, Le and Szepesvari, Csaba and Niu, Gang and Sabato, Sivan},
  volume    = {162},
  series    = {Proceedings of Machine Learning Research},
  month     = {17--23 Jul},
  publisher = {PMLR},
  url       = {https://proceedings.mlr.press/v162/sun22e.html}
}

@inproceedings{bu2022differentially,
  title     = {Differentially Private Bias-Term only Fine-tuning of Foundation Models},
  author    = {Zhiqi Bu and Yu-Xiang Wang and Sheng Zha and George Karypis},
  booktitle = {Workshop on Trustworthy and Socially Responsible Machine Learning, NeurIPS 2022},
  year      = {2022},
  url       = {https://openreview.net/forum?id=6Bo1vhoHolh}
}

@article{su2023fedra,
  title   = {FedRA: A Random Allocation Strategy for Federated Tuning to Unleash the Power of Heterogeneous Clients},
  author  = {Su, Shangchao and Li, Bin and Xue, Xiangyang},
  github  = {https://github.com/leondada/FedRA},
  tags    = {efficiency, reparameterization-based, LoRA, heterogeneous resource},
  journal = {arXiv preprint arXiv:2311.11227},
  url     = {https://doi.org/10.48550/arXiv.2311.11227},
  venue   = {ECCV},
  month   = {oct},
  year    = {2024}
}

@article{lu2023zoopfl,
  title   = {ZooPFL: Exploring Black-box Foundation Models for Personalized Federated Learning},
  author  = {Lu, Wang and Yu, Hao and Wang, Jindong and Teney, Damien and Wang, Haohan and Chen, Yiqiang and Yang, Qiang and Xie, Xing and Ji, Xiangyang},
  journal = {arXiv preprint arXiv:2310.05143},
  url     = {https://doi.org/10.48550/arXiv.2310.05143},
  tags    = {efficiency, trustworthiness, ip protection, black-box tuning, zeroth-order optimization},
  year    = {2023}
}

@article{abs-2310-03123,
  author     = {Zihao Lin and
                Yan Sun and
                Yifan Shi and
                Xueqian Wang and
                Lifu Huang and
                Li Shen and
                Dacheng Tao},
  title      = {Efficient Federated Prompt Tuning for Black-box Large Pre-trained Models},
  journal    = {CoRR},
  volume     = {abs/2310.03123},
  year       = {2023},
  url        = {https://doi.org/10.48550/arXiv.2310.03123},
  doi        = {10.48550/ARXIV.2310.03123},
  eprinttype = {arXiv},
  eprint     = {2310.03123},
  tags       = {efficiency, trustworthiness, ip protection, prompt tuning, black-box tuning, zeroth-order optimization, additive tuning, textual prompt tuning},
  bibsource  = {dblp computer science bibliography, https://dblp.org}
}

@article{geminiteam2023gemini,
  title   = {Gemini: a family of highly capable multimodal models},
  author  = {Gemini Team Google},
  url     = {https://doi.org/10.48550/arXiv.2312.11805},
  journal = {arXiv preprint arXiv:2312.11805},
  year    = {2023}
}

@inproceedings{malladi2023fine,
  title     = {Fine-Tuning Language Models with Just Forward Passes},
  author    = {Sadhika Malladi and Tianyu Gao and Eshaan Nichani and Alex Damian and Jason D. Lee and Danqi Chen and Sanjeev Arora},
  booktitle = {Workshop on Efficient Systems for Foundation Models @ ICML2023},
  year      = {2023},
  url       = {https://openreview.net/forum?id=CcsdvOOzMp}
}

@inproceedings{qin2023federated,
  title     = {Federated Full-Parameter Tuning of Billion-Sized Language Models with Communication Cost under 18 Kilobytes},
  author    = {Qin, Zhen and Chen, Daoyuan and Qian, Bingchen and Ding, Bolin and Li, Yaliang and Deng, Shuiguang},
  booktitle = {Proceedings of the 41th International Conference on Machine Learning},
  venue     = {ICML},
  month     = {jul},
  url       = {https://doi.org/10.48550/arXiv.2312.06353},
  github    = {https://github.com/alibaba/FederatedScope/tree/FedKSeed},
  tags      = {efficiency, zeroth-order optimization},
  year      = {2024}
}

@inproceedings{sun2023fedbpt,
  title     = {FedBPT: Efficient Federated Black-box Prompt Tuning for Large Language Models},
  author    = {Sun, Jingwei and Xu, Ziyue and Yin, Hongxu and Yang, Dong and Xu, Daguang and Chen, Yiran and Roth, Holger R},
  booktitle = {Proceedings of the 41th International Conference on Machine Learning},
  doi       = {https://doi.org/10.48550/arXiv.2310.01467},
  url       = {https://doi.org/10.48550/arXiv.2310.01467},
  tags      = {efficiency, trustworthiness, ip protection, , prompt tuning, black-box tuning, additive tuning, textual prompt tuning, zeroth-order optimization},
  venue     = {ICML},
  month     = {jul},
  year      = {2024}
}

@article{9917343,
  author  = {Fang, Wenzhi and Yu, Ziyi and Jiang, Yuning and Shi, Yuanming and Jones, Colin N. and Zhou, Yong},
  journal = {IEEE Transactions on Signal Processing},
  title   = {Communication-Efficient Stochastic Zeroth-Order Optimization for Federated Learning},
  year    = {2022},
  volume  = {70},
  number  = {},
  pages   = {5058-5073},
  doi     = {10.1109/TSP.2022.3214122}
}

@inproceedings{9613620,
  author    = {Li, Zan and Chen, Li},
  booktitle = {2021 13th International Conference on Wireless Communications and Signal Processing (WCSP)},
  title     = {Communication-Efficient Decentralized Zeroth-order Method on Heterogeneous Data},
  year      = {2021},
  volume    = {},
  number    = {},
  pages     = {1-6},
  doi       = {10.1109/WCSP52459.2021.9613620}
}

@inproceedings{yang2023efficient,
  author    = {Yang, Fu-En and Wang, Chien-Yi and Wang, Yu-Chiang Frank},
  booktitle = {2023 IEEE/CVF International Conference on Computer Vision (ICCV)},
  title     = {Efficient Model Personalization in Federated Learning via Client-Specific Prompt Generation},
  year      = {2023},
  volume    = {},
  number    = {},
  pages     = {19102-19111},
  venue     = {ICCV},
  month     = {oct},
  tags      = {efficiency, additive tuning, prompt tuning, visual prompt tuning},
  url       = {https://ieeexplore.ieee.org/document/10377922},
  doi       = {10.1109/ICCV51070.2023.01755}
}

@article{reed2022generalist,
  title   = {A Generalist Agent},
  author  = {Scott Reed and others},
  journal = {Transactions on Machine Learning Research},
  issn    = {2835-8856},
  year    = {2022},
  url     = {https://openreview.net/forum?id=1ikK0kHjvj},
  note    = {Featured Certification, Outstanding Certification}
}

@inproceedings{dosovitskiy2021an,
  title     = {An Image is Worth 16x16 Words: Transformers for Image Recognition at Scale},
  author    = {Alexey Dosovitskiy and Lucas Beyer and Alexander Kolesnikov and Dirk Weissenborn and Xiaohua Zhai and Thomas Unterthiner and Mostafa Dehghani and Matthias Minderer and Georg Heigold and Sylvain Gelly and Jakob Uszkoreit and Neil Houlsby},
  booktitle = {International Conference on Learning Representations},
  year      = {2021},
  url       = {https://openreview.net/forum?id=YicbFdNTTy}
}

@article{kirillov2023segment,
  title   = {Segment anything},
  author  = {Kirillov, Alexander and Mintun, Eric and Ravi, Nikhila and Mao, Hanzi and Rolland, Chloe and Gustafson, Laura and Xiao, Tete and Whitehead, Spencer and Berg, Alexander C and Lo, Wan-Yen and others},
  journal = {arXiv preprint arXiv:2304.02643},
  doi     = {https://doi.org/10.48550/arXiv.2304.02643},
  eprint  = {2304.02643},
  year    = {2023}
}

@inproceedings{Golovin2020Gradientless,
  title     = {Gradientless Descent: High-Dimensional Zeroth-Order Optimization},
  author    = {Daniel Golovin and John Karro and Greg Kochanski and Chansoo Lee and Xingyou Song and Qiuyi Zhang},
  booktitle = {International Conference on Learning Representations},
  year      = {2020},
  url       = {https://openreview.net/forum?id=Skep6TVYDB}
}

@inproceedings{zhang-etal-2023-fedlegal,
  title     = {{FEDLEGAL}: The First Real-World Federated Learning Benchmark for Legal {NLP}},
  author    = {Zhang, Zhuo  and
               Hu, Xiangjing  and
               Zhang, Jingyuan  and
               Zhang, Yating  and
               Wang, Hui  and
               Qu, Lizhen  and
               Xu, Zenglin},
  editor    = {Rogers, Anna  and
               Boyd-Graber, Jordan  and
               Okazaki, Naoaki},
  booktitle = {Proceedings of the 61st Annual Meeting of the Association for Computational Linguistics (Volume 1: Long Papers)},
  month     = jul,
  year      = {2023},
  address   = {Toronto, Canada},
  publisher = {Association for Computational Linguistics},
  url       = {https://aclanthology.org/2023.acl-long.193},
  doi       = {10.18653/v1/2023.acl-long.193},
  pages     = {3492--3507},
  github    = {https://github.com/SMILELab-FL/FedLegal},
  venue     = {ACL},
  tags      = {application, domain specific}
}

@inproceedings{shin-etal-2023-fedtherapist,
  title     = {{F}ed{T}herapist: Mental Health Monitoring with User-Generated Linguistic Expressions on Smartphones via Federated Learning},
  author    = {Shin, Jaemin  and
               Yoon, Hyungjun  and
               Lee, Seungjoo  and
               Park, Sungjoon  and
               Liu, Yunxin  and
               Choi, Jinho  and
               Lee, Sung-Ju},
  editor    = {Bouamor, Houda  and
               Pino, Juan  and
               Bali, Kalika},
  booktitle = {Proceedings of the 2023 Conference on Empirical Methods in Natural Language Processing},
  month     = dec,
  year      = {2023},
  address   = {Singapore},
  publisher = {Association for Computational Linguistics},
  url       = {https://aclanthology.org/2023.emnlp-main.734},
  doi       = {10.18653/v1/2023.emnlp-main.734},
  pages     = {11971--11988},
  venue     = {EMNLP},
  tags      = {application, domain specific}
}

@inproceedings{liu-etal-2023-communication,
  title     = {Communication Efficient Federated Learning for Multilingual Neural Machine Translation with Adapter},
  author    = {Liu, Yi  and
               Bi, Xiaohan  and
               Li, Lei  and
               Chen, Sishuo  and
               Yang, Wenkai  and
               Sun, Xu},
  editor    = {Rogers, Anna  and
               Boyd-Graber, Jordan  and
               Okazaki, Naoaki},
  booktitle = {Findings of the Association for Computational Linguistics: ACL 2023},
  month     = jul,
  year      = {2023},
  address   = {Toronto, Canada},
  publisher = {Association for Computational Linguistics},
  url       = {https://aclanthology.org/2023.findings-acl.327},
  doi       = {10.18653/v1/2023.findings-acl.327},
  pages     = {5315--5328},
  name      = {Fed-MNMT},
  tags      = {efficiency, additive tuning, adapter tuning, application, Multilingualism},
  venue     = {ACL},
  github    = {https://github.com/lancopku/FedMNMT}
}

@article{yu2023bridging,
  title   = {Bridging the Gap Between Foundation Models and Heterogeneous Federated Learning},
  author  = {Yu, Sixing and Mu{\~n}oz, J Pablo and Jannesari, Ali},
  journal = {arXiv preprint arXiv:2310.00247},
  url     = {https://doi.org/10.48550/arXiv.2310.00247},
  year    = {2023}
}

@inproceedings{good-etal-2023-coordinated,
  title     = {Coordinated Replay Sample Selection for Continual Federated Learning},
  author    = {Good, Jack  and
               Majmudar, Jimit  and
               Dupuy, Christophe  and
               Wang, Jixuan  and
               Peris, Charith  and
               Chung, Clement  and
               Zemel, Richard  and
               Gupta, Rahul},
  editor    = {Wang, Mingxuan  and
               Zitouni, Imed},
  booktitle = {Proceedings of the 2023 Conference on Empirical Methods in Natural Language Processing: Industry Track},
  month     = dec,
  year      = {2023},
  address   = {Singapore},
  publisher = {Association for Computational Linguistics},
  url       = {https://aclanthology.org/2023.emnlp-industry.32},
  doi       = {10.18653/v1/2023.emnlp-industry.32},
  pages     = {331--342}
}

@inproceedings{9593126,
  author    = {Seo, Sejin and Ko, Seung-Woo and Park, Jihong and Kim, Seong-Lyun and Bennis, Mehdi},
  booktitle = {2021 IEEE 22nd International Workshop on Signal Processing Advances in Wireless Communications (SPAWC)},
  title     = {Communication-Efficient and Personalized Federated Lottery Ticket Learning},
  year      = {2021},
  volume    = {},
  number    = {},
  pages     = {581-585},
  doi       = {10.1109/SPAWC51858.2021.9593126}
}

@inproceedings{9708944,
  author    = {Li, Ang and Sun, Jingwei and Wang, Binghui and Duan, Lin and Li, Sicheng and Chen, Yiran and Li, Hai},
  booktitle = {2021 IEEE/ACM Symposium on Edge Computing (SEC)},
  title     = {LotteryFL: Empower Edge Intelligence with Personalized and Communication-Efficient Federated Learning},
  year      = {2021},
  volume    = {},
  number    = {},
  pages     = {68-79},
  doi       = {10.1145/3453142.3492909}
}

@misc{GDPR,
  title  = {Regulation (EU) 2016/679 of the European Parliament and of the Council},
  author = {GDPR},
  year   = {2016},
  url    = {http://data.europa.eu/eli/reg/2016/679/2016-05-04}
}

@misc{CCPA,
  title  = {California Consumer Privacy Act (CCPA)},
  year   = {2023},
  author = {CCPA},
  url    = {https://oag.ca.gov/privacy/ccpa}
}

@article{zhuang2023foundation,
  title   = {When foundation model meets federated learning: Motivations, challenges, and future directions},
  author  = {Zhuang, Weiming and Chen, Chen and Lyu, Lingjuan},
  journal = {arXiv preprint arXiv:2306.15546},
  tags    = {resources, surveys},
  url     = {https://doi.org/10.48550/arXiv.2306.15546},
  year    = {2023}
}

@inproceedings{yu2023federated,
  title     = {Federated Foundation Models: Privacy-Preserving and Collaborative Learning for Large Models},
  author    = {Yu, Sixing  and
               Munoz, Juan Pablo  and
               Jannesari, Ali},
  editor    = {Calzolari, Nicoletta  and
               Kan, Min-Yen  and
               Hoste, Veronique  and
               Lenci, Alessandro  and
               Sakti, Sakriani  and
               Xue, Nianwen},
  booktitle = {Proceedings of the 2024 Joint International Conference on Computational Linguistics, Language Resources and Evaluation (LREC-COLING 2024)},
  month     = may,
  tags      = {resources, surveys},
  year      = {2024},
  address   = {Torino, Italia},
  venue     = {LREC-COLING},
  url       = {https://aclanthology.org/2024.lrec-main.630},
  pages     = {7174--7184}
}

@article{gunasekar2023textbook,
  title         = {Textbooks Are All You Need},
  author        = {Suriya Gunasekar and Yi Zhang and Jyoti Aneja and Caio Cesar Teodoro Mendes and Allie Del Giorno and Sivakanth Gopi and Mojan Javaheripi and Piero Conti Kauffmann and Gustavo Henrique de Rosa and Olli Saarikivi and Adil Salim and Shital Shah and Harkirat Behl and Xin Wang and Sebastien Bubeck and Ronen Eldan and Adam Tauman Kalai and Yin Tat Lee and Yuanzhi Li},
  year          = {2023},
  eprint        = {2306.11644},
  archiveprefix = {arXiv},
  primaryclass  = {cs.CL},
  url           = {https://arxiv.org/abs/2306.11644},
  journal       = {arXiv preprint arXiv:2306.11644}
}

@inproceedings{woisetschlager2024survey,
  title     = {A Survey on Efficient Federated Learning Methods for Foundation Model Training},
  author    = {Woisetschl{\"a}ger, Herbert and Isenko, Alexander and Wang, Shiqiang and Mayer, Ruben and Jacobsen, Hans-Arno},
  booktitle = {Proceedings of the Thirty-Second International Joint Conference on
               Artificial Intelligence, {IJCAI-24}},
  publisher = {International Joint Conferences on Artificial Intelligence Organization},
  year      = {2024},
  month     = {aug},
  venue     = {IJCAI},
  tags      = {efficiency, resources, surveys},
  url       = {https://www.ijcai.org/proceedings/2024/0919.pdf}
}

@article{xu2024survey,
  title         = {A Survey of Resource-efficient LLM and Multimodal Foundation Models},
  author        = {Mengwei Xu and Wangsong Yin and Dongqi Cai and Rongjie Yi and Daliang Xu and Qipeng Wang and Bingyang Wu and Yihao Zhao and Chen Yang and Shihe Wang and Qiyang Zhang and Zhenyan Lu and Li Zhang and Shangguang Wang and Yuanchun Li and Yunxin Liu and Xin Jin and Xuanzhe Liu},
  year          = {2024},
  eprint        = {2401.08092},
  archiveprefix = {arXiv},
  primaryclass  = {cs.LG},
  url           = {https://arxiv.org/abs/2401.08092},
  journal       = {arXiv preprint arXiv:2401.08092}
}

@article{yang2023federated,
  title     = {Federated Continual Learning via Knowledge Fusion: A Survey},
  issn      = {2326-3865},
  url       = {http://dx.doi.org/10.1109/TKDE.2024.3363240},
  doi       = {10.1109/tkde.2024.3363240},
  journal   = {IEEE Transactions on Knowledge and Data Engineering},
  publisher = {Institute of Electrical and Electronics Engineers (IEEE)},
  author    = {Yang, Xin and Yu, Hao and Gao, Xin and Wang, Hao and Zhang, Junbo and Li, Tianrui},
  year      = {2024},
  pages     = {1--20}
}

@article{6790628,
  author  = {Hansen, Nikolaus and Ostermeier, Andreas},
  journal = {Evolutionary Computation},
  title   = {Completely Derandomized Self-Adaptation in Evolution Strategies},
  year    = {2001},
  volume  = {9},
  number  = {2},
  pages   = {159-195},
  doi     = {10.1162/106365601750190398}
}

@article{li2023visual,
  title   = {Visual Prompt Based Personalized Federated Learning},
  author  = {Li, Guanghao and Wu, Wansen and Sun, Yan and Shen, Li and Wu, Baoyuan and Tao, Dacheng},
  journal = {Transactions on Machine Learning Research},
  year    = {2024},
  month   = {feb},
  venue   = {TMLR},
  tags    = {efficiency, additive tuning, prompt tuning, visual prompt tuning},
  github  = {https://github.com/hkgdifyu/pFedPT},
  url     = {https://openreview.net/forum?id=dUVejidXO7}
}

@inproceedings{10205077,
  author    = {Feng, Chun-Mei and Li, Bangjun and Xu, Xinxing and Liu, Yong and Fu, Huazhu and Zuo, Wangmeng},
  booktitle = {2023 IEEE/CVF Conference on Computer Vision and Pattern Recognition (CVPR)},
  title     = {Learning Federated Visual Prompt in Null Space for MRI Reconstruction},
  year      = {2023},
  month     = {jun},
  volume    = {},
  venue     = {CVPR},
  number    = {},
  pages     = {8064-8073},
  doi       = {10.1109/CVPR52729.2023.00779},
  url       = {https://ieeexplore.ieee.org/document/10205077},
  tags      = {efficiency, additive tuning, application, domain specific, healthcare, MRI Reconstruction, prompt tuning, visual prompt tuning},
  github    = {https://github.com/chunmeifeng/FedPR}
}

@article{lincy2020early,
  title   = {Early detection of type-2 diabetes using federated learning},
  author  = {Lincy, M and Kowshalya, A Meena},
  journal = {International Journal of Scientific Research in Science, Engineering and Technology},
  volume  = {12},
  pages   = {257--267},
  url     = {https://api.semanticscholar.org/CorpusID:234514776},
  year    = {2020}
}

@article{PANDYA2023102987,
  title   = {Federated learning for smart cities: A comprehensive survey},
  journal = {Sustainable Energy Technologies and Assessments},
  volume  = {55},
  pages   = {102987},
  year    = {2023},
  issn    = {2213-1388},
  doi     = {https://doi.org/10.1016/j.seta.2022.102987},
  url     = {https://www.sciencedirect.com/science/article/pii/S2213138822010359},
  author  = {Sharnil Pandya and Gautam Srivastava and Rutvij Jhaveri and M. Rajasekhara Babu and Sweta Bhattacharya and Praveen Kumar Reddy Maddikunta and Spyridon Mastorakis and Md. Jalil Piran and Thippa Reddy Gadekallu}
}

@article{ramu2022federated,
  title    = {Federated learning enabled digital twins for smart cities: Concepts, recent advances, and future directions},
  journal  = {Sustainable Cities and Society},
  volume   = {79},
  pages    = {103663},
  year     = {2022},
  issn     = {2210-6707},
  doi      = {https://doi.org/10.1016/j.scs.2021.103663},
  url      = {https://www.sciencedirect.com/science/article/pii/S2210670721009264},
  author   = {Swarna Priya Ramu and Parimala Boopalan and Quoc-Viet Pham and Praveen Kumar Reddy Maddikunta and Thien Huynh-The and Mamoun Alazab and Thanh Thi Nguyen and Thippa Reddy Gadekallu}
}

@article{joshi2022federated,
  author     = {Joshi, Madhura and Pal, Ankit and Sankarasubbu, Malaikannan},
  title      = {Federated Learning for Healthcare Domain - Pipeline, Applications and Challenges},
  year       = {2022},
  issue_date = {October 2022},
  publisher  = {Association for Computing Machinery},
  address    = {New York, NY, USA},
  volume     = {3},
  number     = {4},
  url        = {https://doi.org/10.1145/3533708},
  doi        = {10.1145/3533708},
  journal    = {ACM Trans. Comput. Healthcare},
  month      = {nov},
  articleno  = {40},
  numpages   = {36}
}

@article{rieke2020future,
  title     = {The future of digital health with federated learning},
  author    = {Rieke, Nicola and Hancox, Jonny and Li, Wenqi and Milletari, Fausto and Roth, Holger R and Albarqouni, Shadi and Bakas, Spyridon and Galtier, Mathieu N and Landman, Bennett A and Maier-Hein, Klaus and others},
  journal   = {NPJ digital medicine},
  volume    = {3},
  number    = {1},
  pages     = {119},
  url       = {https://doi.org/10.1038/s41746-020-00323-1},
  year      = {2020},
  publisher = {Nature Publishing Group UK London}
}

@article{chatterjee2023use,
  title         = {Use of Federated Learning and Blockchain towards Securing Financial Services},
  author        = {Pushpita Chatterjee and Debashis Das and Danda B Rawat},
  year          = {2023},
  eprint        = {2303.12944},
  archiveprefix = {arXiv},
  primaryclass  = {cs.CR},
  url           = {https://arxiv.org/abs/2303.12944},
  journal       = {arXiv preprint arXiv:2303.12944}
}

@article{liu2023efficient,
  author         = {Liu, Tao and Wang, Zhi and He, Hui and Shi, Wei and Lin, Liangliang and An, Ran and Li, Chenhao},
  title          = {Efficient and Secure Federated Learning for Financial Applications},
  journal        = {Applied Sciences},
  volume         = {13},
  year           = {2023},
  number         = {10},
  article-number = {5877},
  url            = {https://www.mdpi.com/2076-3417/13/10/5877},
  issn           = {2076-3417},
  doi            = {10.3390/app13105877}
}

@inproceedings{jia2022visual,
  author    = {Jia, Menglin
               and Tang, Luming
               and Chen, Bor-Chun
               and Cardie, Claire
               and Belongie, Serge
               and Hariharan, Bharath
               and Lim, Ser-Nam},
  editor    = {Avidan, Shai
               and Brostow, Gabriel
               and Ciss{\'e}, Moustapha
               and Farinella, Giovanni Maria
               and Hassner, Tal},
  title     = {Visual Prompt Tuning},
  booktitle = {Computer Vision -- ECCV 2022},
  year      = {2022},
  publisher = {Springer Nature Switzerland},
  address   = {Cham},
  pages     = {709--727},
  url       = {https://doi.org/10.1007/978-3-031-19827-4_41},
  isbn      = {978-3-031-19827-4}
}

@article{10.1145/3560815,
  author     = {Liu, Pengfei and Yuan, Weizhe and Fu, Jinlan and Jiang, Zhengbao and Hayashi, Hiroaki and Neubig, Graham},
  title      = {Pre-train, Prompt, and Predict: A Systematic Survey of Prompting Methods in Natural Language Processing},
  year       = {2023},
  issue_date = {September 2023},
  publisher  = {Association for Computing Machinery},
  address    = {New York, NY, USA},
  volume     = {55},
  number     = {9},
  issn       = {0360-0300},
  url        = {https://doi.org/10.1145/3560815},
  doi        = {10.1145/3560815},
  journal    = {ACM Comput. Surv.},
  month      = {jan},
  articleno  = {195},
  numpages   = {35}
}

@inproceedings{qiu2024textdriven,
  title     = {Federated Text-driven Prompt Generation for Vision-Language Models},
  author    = {Chen Qiu and Xingyu Li and Chaithanya Kumar Mummadi and Madan Ravi Ganesh and Zhenzhen Li and Lu Peng and Wan-Yi Lin},
  booktitle = {The Twelfth International Conference on Learning Representations},
  year      = {2024},
  month     = {may},
  venue     = {ICLR},
  tags      = {efficiency, additive tuning, prompt tuning, textual-visual prompt tuning},
  url       = {https://openreview.net/forum?id=NW31gAylIm}
}

@inproceedings{zhao2024breaking,
  title     = {Breaking Physical and Linguistic Borders: Multilingual Federated Prompt Tuning for Low-Resource Languages},
  author    = {Wanru Zhao and Yihong Chen and Royson Lee and Xinchi Qiu and Yan Gao and Hongxiang Fan and Nicholas Donald Lane},
  booktitle = {The Twelfth International Conference on Learning Representations},
  month     = {may},
  year      = {2024},
  tags      = {application, Multilingualism},
  venue     = {ICLR},
  url       = {https://openreview.net/forum?id=zzqn5G9fjn},
  tags      = {efficiency, additive tuning,  prompt tuning, textual prompt tuning, application, low resource languages, Multilingualism},
  github    = {https://github.com/Ryan0v0/multilingual_borders}
}

@article{su2023federated,
  title   = {Federated Adaptive Prompt Tuning for Multi-Domain Collaborative Learning},
  volume  = {38},
  url     = {https://ojs.aaai.org/index.php/AAAI/article/view/29434},
  doi     = {10.1609/aaai.v38i13.29434},
  number  = {13},
  journal = {Proceedings of the AAAI Conference on Artificial Intelligence},
  author  = {Su, Shangchao and Yang, Mingzhao and Li, Bin and Xue, Xiangyang},
  year    = {2024},
  month   = {mar},
  venue   = {AAAI},
  github  = {https://github.com/leondada/FedAPT},
  tags    = {efficiency, additive tuning, prompt tuning, textual-visual prompt tuning, adaptability, domain-centric adaptation, multi-domain adaptation},
  pages   = {15117-15125}
}

@article{feng2023adapterbased,
  author   = {Feng, Xiachong and Feng, Xiaocheng and Du, Xiyuan and Kan, Min-Yen and Qin, Bing},
  journal  = {IEEE/ACM Transactions on Audio, Speech, and Language Processing},
  title    = {Adapter-based Selective Knowledge Distillation for Federated Multi-domain Meeting Summarization},
  year     = {2024},
  volume   = {},
  url      = {https://ieeexplore.ieee.org/document/10557150},
  number   = {},
  venue    = {TASLP},
  month    = {jun},
  pages    = {1-15},
  tags     = {efficiency, knowledge distillation, additive tuning, adapter tuning, adaptability, domain-centric adaptation, multi-domain adaptation},
  doi      = {10.1109/TASLP.2024.3414313}
}

@inproceedings{dong-etal-2023-tunable,
  title     = {Tunable Soft Prompts are Messengers in Federated Learning},
  author    = {Dong, Chenhe  and
               Xie, Yuexiang  and
               Ding, Bolin  and
               Shen, Ying  and
               Li, Yaliang},
  editor    = {Bouamor, Houda  and
               Pino, Juan  and
               Bali, Kalika},
  booktitle = {Findings of the Association for Computational Linguistics: EMNLP 2023},
  month     = dec,
  year      = {2023},
  address   = {Singapore},
  publisher = {Association for Computational Linguistics},
  url       = {https://aclanthology.org/2023.findings-emnlp.976},
  doi       = {10.18653/v1/2023.findings-emnlp.976},
  pages     = {14665--14675},
  tags      = {efficiency, additive tuning, prompt tuning, textual prompt tuning},
  venue     = {EMNLP},
  github    = {https://github.com/alibaba/FederatedScope/tree/fedsp/federatedscope/nlp/fedsp}
}

@article{10666083,
  author  = {Wu, Panlong and Li, Kangshuo and Wang, Ting and Dong, Yanjie and Leung, Victor C. M. and Wang, Fangxin},
  journal = {IEEE Transactions on Mobile Computing},
  venue   = {IEEE TMC},
  title   = {FedFMSL: Federated Learning of Foundation Models With Sparsely Activated LoRA},
  year    = {2024},
  month   = dec,
  volume  = {23},
  number  = {12},
  pages   = {15167-15181},
  tags    = {efficiency, reparameterization-based, LoRA, sparsification},
  url     = {https://ieeexplore.ieee.org/abstract/document/10666083},
  doi     = {10.1109/TMC.2024.3454634}
}

@article{10398264,
  author   = {Huang, Xumin and Li, Peichun and Du, Hongyang and Kang, Jiawen and Niyato, Dusit and Kim, Dong In and Wu, Yuan},
  journal  = {IEEE Network},
  title    = {Federated Learning-Empowered AI-Generated Content in Wireless Networks},
  year     = {2024},
  volume   = {},
  number   = {},
  pages    = {1-1},
  doi      = {10.1109/MNET.2024.3353377}
}

@inproceedings{10.1145/3570361.3592505,
  author    = {Cai, Dongqi and Wu, Yaozong and Wang, Shangguang and Lin, Felix Xiaozhu and Xu, Mengwei},
  title     = {Efficient Federated Learning for Modern NLP},
  year      = {2023},
  isbn      = {9781450399906},
  publisher = {Association for Computing Machinery},
  address   = {New York, NY, USA},
  url       = {https://doi.org/10.1145/3570361.3592505},
  doi       = {10.1145/3570361.3592505},
  booktitle = {Proceedings of the 29th Annual International Conference on Mobile Computing and Networking},
  articleno = {37},
  numpages  = {16},
  series    = {ACM MobiCom '23}
}

@inproceedings{10389738,
  author    = {Jia, Junteng and Li, Ke and Malek, Mani and Malik, Kshitiz and Mahadeokar, Jay and Kalinli, Ozlem and Seide, Frank},
  booktitle = {2023 IEEE Automatic Speech Recognition and Understanding Workshop (ASRU)},
  title     = {Joint Federated Learning and Personalization for on-Device ASR},
  year      = {2023},
  volume    = {},
  number    = {},
  venue     = {ASRU},
  month     = {dec},
  pages     = {1-8},
  url       = {https://ieeexplore.ieee.org/document/10389738},
  tags      = {personalization, ASR, efficiency, additive tuning,  adapter tuning, application, speech},
  doi       = {10.1109/ASRU57964.2023.10389738}
}

@inproceedings{10389620,
  author    = {Azam, Sheikh Shams and Likhomanenko, Tatiana and Pelikan, Martin and Silovsky, Jan "Honza"},
  booktitle = {2023 IEEE Automatic Speech Recognition and Understanding Workshop (ASRU)},
  title     = {Importance of Smoothness Induced by Optimizers in Fl4Asr: Towards Understanding Federated Learning for End-To-End ASR},
  year      = {2023},
  volume    = {},
  number    = {},
  pages     = {1-8},
  venue     = {ASRU},
  month     = {dec},
  tags      = {personalization, ASR, application, speech},
  url       = {https://ieeexplore.ieee.org/document/10389620},
  doi       = {10.1109/ASRU57964.2023.10389620}
}

@inproceedings{du2024communicationefficient,
  author    = {Du, Yichao and Zhang, Zhirui and Yue, Linan and Huang, Xu and Zhang, Yuqing and Xu, Tong and Xu, Linli and Chen, Enhong},
  booktitle = {ICASSP 2024 - 2024 IEEE International Conference on Acoustics, Speech and Signal Processing (ICASSP)},
  title     = {Communication-Efficient Personalized Federated Learning for Speech-to-Text Tasks},
  year      = {2024},
  venue     = {ICASSP},
  volume    = {},
  url       = {https://ieeexplore.ieee.org/document/10447662},
  number    = {},
  tags      = {efficiency, personalization, speech, ASR, reparameterization-based, LoRA, application, speech},
  month     = {mar},
  pages     = {10001-10005},
  doi       = {10.1109/ICASSP48485.2024.10447662}
}

@inproceedings{anonymous2025selective,
  title     = {Selective Aggregation for Low-Rank Adaptation in Federated Learning},
  author    = {Pengxin Guo and Shuang Zeng and Yanran Wang and Huijie Fan and Feifei Wang and Liangqiong Qu},
  venue     = {ICLR},
  month     = {apr},
  booktitle = {The Thirteenth International Conference on Learning Representations},
  year      = {2025},
  github    = {https://github.com/Pengxin-Guo/FedSA-LoRA},
  tags      = {efficiency, personalization, reparameterization-based, LoRA},
  url       = {https://openreview.net/forum?id=iX3uESGdsO}
}

@inproceedings{cho2024heterogeneous,
  title     = {Heterogeneous {L}o{RA} for Federated Fine-tuning of On-Device Foundation Models},
  author    = {Cho, Yae Jee  and
               Liu, Luyang  and
               Xu, Zheng  and
               Fahrezi, Aldi  and
               Joshi, Gauri},
  editor    = {Al-Onaizan, Yaser  and
               Bansal, Mohit  and
               Chen, Yun-Nung},
  booktitle = {Proceedings of the 2024 Conference on Empirical Methods in Natural Language Processing},
  month     = {nov},
  venue     = {EMNLP},
  year      = {2024},
  address   = {Miami, Florida, USA},
  publisher = {Association for Computational Linguistics},
  url       = {https://aclanthology.org/2024.emnlp-main.717},
  tags      = {efficiency, reparameterization-based, LoRA, heterogeneous resource},
  pages     = {12903--12913}
}

@article{jiang2023lowparameter,
  title         = {Low-Parameter Federated Learning with Large Language Models},
  author        = {Jingang Jiang and Xiangyang Liu and Chenyou Fan},
  year          = {2023},
  eprint        = {2307.13896},
  archiveprefix = {arXiv},
  tags          = {efficiency, reparameterization-based, LoRA},
  primaryclass  = {cs.DC},
  url           = {https://arxiv.org/abs/2307.13896},
  journal       = {arXiv preprint arXiv:2307.13896}
}

@inproceedings{9706703,
  author    = {Yao, Chun-Han and Gong, Boqing and Qi, Hang and Cui, Yin and Zhu, Yukun and Yang, Ming-Hsuan},
  booktitle = {2022 IEEE/CVF Winter Conference on Applications of Computer Vision (WACV)},
  title     = {Federated Multi-Target Domain Adaptation},
  year      = {2022},
  volume    = {},
  number    = {},
  pages     = {1081-1090},
  doi       = {10.1109/WACV51458.2022.00115}
}

@inproceedings{diao2021heterofl,
  title     = {Hetero{\{}FL{\}}: Computation and Communication Efficient Federated Learning for Heterogeneous Clients},
  author    = {Enmao Diao and Jie Ding and Vahid Tarokh},
  booktitle = {International Conference on Learning Representations},
  year      = {2021},
  url       = {https://openreview.net/forum?id=TNkPBBYFkXg}
}

@article{9762360,
  author   = {Jiang, Yuang and Wang, Shiqiang and Valls, Victor and Ko, Bong Jun and Lee, Wei-Han and Leung, Kin K. and Tassiulas, Leandros},
  journal  = {IEEE Transactions on Neural Networks and Learning Systems},
  title    = {Model Pruning Enables Efficient Federated Learning on Edge Devices},
  year     = {2023},
  volume   = {34},
  number   = {12},
  pages    = {10374-10386},
  doi      = {10.1109/TNNLS.2022.3166101}
}

@article{bubeck2023sparks,
  title         = {Sparks of Artificial General Intelligence: Early experiments with GPT-4},
  author        = {S\'ebastien Bubeck and Varun Chandrasekaran and Ronen Eldan and Johannes Gehrke and Eric Horvitz and Ece Kamar and Peter Lee and Yin Tat Lee and Yuanzhi Li and Scott Lundberg and Harsha Nori and Hamid Palangi and Marco Tulio Ribeiro and Yi Zhang},
  year          = {2023},
  eprint        = {2303.12712},
  archiveprefix = {arXiv},
  primaryclass  = {cs.CL},
  url           = {https://arxiv.org/abs/2303.12712},
  journal       = {arXiv preprint arXiv:2303.12712}
}

@article{li2024position,
  title         = {Position Paper: Assessing Robustness, Privacy, and Fairness in Federated Learning Integrated with Foundation Models},
  author        = {Xi Li and Jiaqi Wang},
  year          = {2024},
  eprint        = {2402.01857},
  archiveprefix = {arXiv},
  primaryclass  = {cs.LG},
  url           = {https://arxiv.org/abs/2402.01857},
  journal       = {arXiv preprint arXiv:2402.01857}
}

@inproceedings{NEURIPS2019_60a6c400,
  author    = {Zhu, Ligeng and Liu, Zhijian and Han, Song},
  booktitle = {Advances in Neural Information Processing Systems},
  editor    = {H. Wallach and H. Larochelle and A. Beygelzimer and F. d\textquotesingle Alch\'{e}-Buc and E. Fox and R. Garnett},
  pages     = {},
  publisher = {Curran Associates, Inc.},
  title     = {Deep Leakage from Gradients},
  url       = {https://proceedings.neurips.cc/paper\_files/paper/2019/file/60a6c4002cc7b29142def8871531281a-Paper.pdf},
  volume    = {32},
  year      = {2019}
}

@inproceedings{NEURIPS2020_c4ede56b,
  author    = {Geiping, Jonas and Bauermeister, Hartmut and Dr\"{o}ge, Hannah and Moeller, Michael},
  booktitle = {Advances in Neural Information Processing Systems},
  editor    = {H. Larochelle and M. Ranzato and R. Hadsell and M.F. Balcan and H. Lin},
  pages     = {16937--16947},
  publisher = {Curran Associates, Inc.},
  title     = {Inverting Gradients - How easy is it to break privacy in federated learning?},
  url       = {https://proceedings.neurips.cc/paper\_files/paper/2020/file/c4ede56bbd98819ae6112b20ac6bf145-Paper.pdf},
  volume    = {33},
  year      = {2020}
}

@inproceedings{xu2024fwdllm,
  author    = {Mengwei Xu and Dongqi Cai and Yaozong Wu and Xiang Li and Shangguang Wang},
  title     = {{FwdLLM}: Efficient Federated Finetuning of Large Language Models with Perturbed Inferences},
  booktitle = {2024 USENIX Annual Technical Conference (USENIX ATC 24)},
  year      = {2024},
  isbn      = {978-1-939133-41-0},
  address   = {Santa Clara, CA},
  github    = {https://github.com/UbiquitousLearning/FwdLLM},
  pages     = {579--596},
  tags      = {efficiency, zeroth-order optimization},
  venue     = {ATC},
  url       = {https://www.usenix.org/conference/atc24/presentation/xu-mengwei},
  publisher = {USENIX Association},
  month     = {jul}
}

@inproceedings{ling2024convergence,
  author    = {Ling, Zhenqing and Chen, Daoyuan and Yao, Liuyi and Li, Yaliang and Shen, Ying},
  title     = {On the Convergence of Zeroth-Order Federated Tuning for Large Language Models},
  year      = {2024},
  isbn      = {9798400704901},
  publisher = {Association for Computing Machinery},
  address   = {New York, NY, USA},
  url       = {https://doi.org/10.1145/3637528.3671865},
  doi       = {10.1145/3637528.3671865},
  booktitle = {Proceedings of the 30th ACM SIGKDD Conference on Knowledge Discovery and Data Mining},
  pages     = {1827–1838},
  numpages  = {12},
  tags      = {efficiency, zeroth-order optimization},
  github    = {https://github.com/alibaba/FederatedScope/tree/FedMeZO},
  month     = {aug},
  venue     = {KDD}
}

@article{feng2023does,
  title         = {Does Federated Learning Really Need Backpropagation?},
  author        = {Haozhe Feng and Tianyu Pang and Chao Du and Wei Chen and Shuicheng Yan and Min Lin},
  year          = {2023},
  eprint        = {2301.12195},
  archiveprefix = {arXiv},
  primaryclass  = {cs.LG},
  url           = {https://arxiv.org/abs/2301.12195},
  journal       = {arXiv preprint arXiv:2301.12195}
}

@article{maritan2023fedzen,
  title         = {FedZeN: Towards superlinear zeroth-order federated learning via incremental Hessian estimation},
  author        = {Alessio Maritan and Subhrakanti Dey and Luca Schenato},
  year          = {2023},
  eprint        = {2309.17174},
  archiveprefix = {arXiv},
  primaryclass  = {cs.LG},
  url           = {https://arxiv.org/abs/2309.17174},
  journal       = {arXiv preprint arXiv:2309.17174}
}

@inproceedings{NEURIPS2020_a1d4c20b,
  author    = {He, Chaoyang and Annavaram, Murali and Avestimehr, Salman},
  booktitle = {Advances in Neural Information Processing Systems},
  editor    = {H. Larochelle and M. Ranzato and R. Hadsell and M.F. Balcan and H. Lin},
  pages     = {14068--14080},
  publisher = {Curran Associates, Inc.},
  title     = {Group Knowledge Transfer: Federated Learning of Large CNNs at the Edge},
  url       = {https://proceedings.neurips.cc/paper\_files/paper/2020/file/a1d4c20b182ad7137ab3606f0e3fc8a4-Paper.pdf},
  volume    = {33},
  year      = {2020}
}

@article{10.1145/3510033,
  author     = {Tian, Yuanyishu and Wan, Yao and Lyu, Lingjuan and Yao, Dezhong and Jin, Hai and Sun, Lichao},
  title      = {FedBERT: When Federated Learning Meets Pre-training},
  year       = {2022},
  issue_date = {August 2022},
  publisher  = {Association for Computing Machinery},
  address    = {New York, NY, USA},
  volume     = {13},
  number     = {4},
  issn       = {2157-6904},
  url        = {https://doi.org/10.1145/3510033},
  doi        = {10.1145/3510033},
  journal    = {ACM Trans. Intell. Syst. Technol.},
  month      = {aug},
  articleno  = {66},
  numpages   = {26}
}

@inproceedings{pmlr-v108-reisizadeh20a,
  title     = {FedPAQ: A Communication-Efficient Federated Learning Method with Periodic Averaging and Quantization},
  author    = {Reisizadeh, Amirhossein and Mokhtari, Aryan and Hassani, Hamed and Jadbabaie, Ali and Pedarsani, Ramtin},
  booktitle = {Proceedings of the Twenty Third International Conference on Artificial Intelligence and Statistics},
  pages     = {2021--2031},
  year      = {2020},
  editor    = {Chiappa, Silvia and Calandra, Roberto},
  volume    = {108},
  series    = {Proceedings of Machine Learning Research},
  month     = {26--28 Aug},
  publisher = {PMLR},
  url       = {https://proceedings.mlr.press/v108/reisizadeh20a.html}
}

@inproceedings{anonymous-acl24-arr,
  title     = {Personalized Federated Learning for Text Classification with Gradient-Free Prompt Tuning},
  author    = {Rui, Wang and Tong, Yu and Ruiyi, Zhang and Sungchul, Kim and Ryan A., Rossi and Handong, Zhao
               and Junda, Wu and Subrata, Mitra and Lina, Yao and Ricardo, Henao},
  booktitle = {Proceedings of the 2024 Conference of the North American Chapter of the Association for Computational Linguistics: Human Language Technologies},
  year      = {2024},
  venue     = {NAACL},
  month     = {jun},
  tags      = {efficiency, additive tuning, prompt tuning, trustworthiness, ip protection, black-box tuning, textual prompt tuning, zeroth-order optimization},
  url       = {https://aclanthology.org/2024.findings-naacl.286.pdf}
}

@article{10.1145/3494834.3500240,
  author     = {Bonawitz, Kallista and Kairouz, Peter and McMahan, Brendan and Ramage, Daniel},
  title      = {Federated Learning and Privacy: Building privacy-preserving systems for machine learning and data science on decentralized data},
  year       = {2021},
  issue_date = {September-October 2021},
  publisher  = {Association for Computing Machinery},
  address    = {New York, NY, USA},
  volume     = {19},
  number     = {5},
  issn       = {1542-7730},
  url        = {https://doi.org/10.1145/3494834.3500240},
  doi        = {10.1145/3494834.3500240},
  journal    = {Queue},
  month      = {nov},
  pages      = {87-114},
  numpages   = {28}
}

@article{10018261,
  author   = {Li, Shenghui and Ngai, Edith C. H. and Voigt, Thiemo},
  journal  = {IEEE Transactions on Big Data},
  title    = {An Experimental Study of Byzantine-Robust Aggregation Schemes in Federated Learning},
  year     = {2024},
  volume   = {10},
  number   = {6},
  pages    = {975-988},
  url      = {https://ieeexplore.ieee.org/document/10018261},
  doi      = {10.1109/TBDATA.2023.3237397}
}

@article{kang2024grounding,
  author     = {Kang, Yan and Fan, Tao and Gu, Hanlin and Zhang, Xiaojin and Fan, Lixin and Yang, Qiang},
  title      = {Grounding Foundation Models through Federated Transfer Learning: A General Framework},
  year       = {2025},
  issue_date = {August 2025},
  publisher  = {Association for Computing Machinery},
  address    = {New York, NY, USA},
  volume     = {16},
  number     = {4},
  issn       = {2157-6904},
  url        = {https://doi.org/10.1145/3742788},
  doi        = {10.1145/3742788},
  venue      = {ACM TIST},
  tags       = {knowledge distillation},
  journal    = {ACM Trans. Intell. Syst. Technol.},
  month      = aug,
  articleno  = {96},
  numpages   = {54}
}

@inproceedings{9603498,
  author    = {Tekgul, Buse G. A. and Xia, Yuxi and Marchal, Samuel and Asokan, N.},
  booktitle = {2021 40th International Symposium on Reliable Distributed Systems (SRDS)},
  title     = {WAFFLE: Watermarking in Federated Learning},
  year      = {2021},
  volume    = {},
  number    = {},
  pages     = {310-320},
  doi       = {10.1109/SRDS53918.2021.00038}
}

@inproceedings{217591,
  author    = {Yossi Adi and Carsten Baum and Moustapha Cisse and Benny Pinkas and Joseph Keshet},
  title     = {Turning Your Weakness Into a Strength: Watermarking Deep Neural Networks by Backdooring},
  booktitle = {27th USENIX Security Symposium (USENIX Security 18)},
  year      = {2018},
  isbn      = {978-1-939133-04-5},
  address   = {Baltimore, MD},
  pages     = {1615--1631},
  url       = {https://www.usenix.org/conference/usenixsecurity18/presentation/adi},
  publisher = {USENIX Association},
  month     = aug
}

@inproceedings{yu2023who,
  title     = {Who Leaked the Model? Tracking {IP} Infringers in Accountable Federated Learning},
  author    = {Shuyang Yu and Junyuan Hong and Yi Zeng and Fei Wang and Ruoxi Jia and Jiayu Zhou},
  booktitle = {NeurIPS 2023 Workshop on Regulatable ML},
  year      = {2023},
  url       = {https://openreview.net/forum?id=qhP1aHHyeA}
}

@inproceedings{NEURIPS2022_35b5c175,
  author    = {Gupta, Samyak and Huang, Yangsibo and Zhong, Zexuan and Gao, Tianyu and Li, Kai and Chen, Danqi},
  booktitle = {Advances in Neural Information Processing Systems},
  editor    = {S. Koyejo and S. Mohamed and A. Agarwal and D. Belgrave and K. Cho and A. Oh},
  pages     = {8130--8143},
  publisher = {Curran Associates, Inc.},
  title     = {Recovering Private Text in Federated Learning of Language Models},
  url       = {https://proceedings.neurips.cc/paper_files/paper/2022/file/35b5c175e139bff5f22a5361270fce87-Paper-Conference.pdf},
  volume    = {35},
  venue     = {NeurIPS},
  month     = {nov},
  tags      = {trustworthiness, differential privacy, privacy preservation, privacy attack},
  github    = {https://github.com/Princeton-SysML/FILM},
  year      = {2022}
}

@inproceedings{zhang-etal-2024-revisiting-data,
  title     = {Revisiting Data Reconstruction Attacks on Real-world Dataset for Federated Natural Language Understanding},
  author    = {Zhang, Zhuo  and
               Huang, Jintao  and
               Hu, Xiangjing  and
               Zhang, Jingyuan  and
               Zhang, Yating  and
               Wang, Hui  and
               Yu, Yue  and
               Wang, Qifan  and
               Qu, Lizhen  and
               Xu, Zenglin},
  editor    = {Calzolari, Nicoletta  and
               Kan, Min-Yen  and
               Hoste, Veronique  and
               Lenci, Alessandro  and
               Sakti, Sakriani  and
               Xue, Nianwen},
  booktitle = {Proceedings of the 2024 Joint International Conference on Computational Linguistics, Language Resources and Evaluation (LREC-COLING 2024)},
  month     = may,
  venue     = {LREC-COLING},
  year      = {2024},
  address   = {Torino, Italia},
  tags      = {trustworthiness, differential privacy, privacy preservation, privacy attack},
  publisher = {ELRA and ICCL},
  github    = {https://github.com/SMILELab-FL/FedAttack},
  url       = {https://aclanthology.org/2024.lrec-main.1227},
  pages     = {14080--14091}
}

@inproceedings{NEURIPS2024_97d008f7,
  author    = {Wen, Yuxin and Marchyok, Leo and Hong, Sanghyun and Geiping, Jonas and Goldstein, Tom and Carlini, Nicholas},
  booktitle = {Advances in Neural Information Processing Systems},
  editor    = {A. Globerson and L. Mackey and D. Belgrave and A. Fan and U. Paquet and J. Tomczak and C. Zhang},
  pages     = {83374--83396},
  publisher = {Curran Associates, Inc.},
  tags      = {trustworthiness, differential privacy, privacy preservation, privacy attack},
  venue     = {NeurIPS},
  month     = {sep},
  title     = {Privacy Backdoors: Enhancing Membership Inference through Poisoning Pre-trained Models},
  url       = {https://proceedings.neurips.cc/paper_files/paper/2024/file/97d008f7873b8dd55cb6dd343fc4386f-Paper-Conference.pdf},
  volume    = {37},
  year      = {2024}
}

@inproceedings{chu2023panning,
  title     = {Panning for Gold in Federated Learning: Targeted Text Extraction under Arbitrarily Large-Scale Aggregation},
  author    = {Hong-Min Chu and Jonas Geiping and Liam H Fowl and Micah Goldblum and Tom Goldstein},
  booktitle = {The Eleventh International Conference on Learning Representations },
  year      = {2023},
  month     = mar,
  tags      = {trustworthiness, differential privacy, privacy preservation, privacy attack},
  venue     = {ICLR},
  url       = {https://openreview.net/forum?id=A9WQaxYsfx}
}

@inproceedings{pmlr-v238-vu24a,
  title     = { Analysis of Privacy Leakage in Federated Large Language Models },
  author    = {Vu, Minh and Nguyen, Truc and Jeter, Tre' and T. Thai, My},
  booktitle = {Proceedings of The 27th International Conference on Artificial Intelligence and Statistics},
  pages     = {1423--1431},
  year      = {2024},
  editor    = {Dasgupta, Sanjoy and Mandt, Stephan and Li, Yingzhen},
  volume    = {238},
  series    = {Proceedings of Machine Learning Research},
  month     = {may},
  venue     = {AISTATS},
  github    = {https://github.com/vunhatminh/FL_Attacks},
  publisher = {PMLR},
  tags      = {trustworthiness, differential privacy, privacy preservation, privacy attack},
  url       = {https://proceedings.mlr.press/v238/vu24a.html}
}

@inproceedings{xu-etal-2023-federated,
  title     = {Federated Learning of Gboard Language Models with Differential Privacy},
  author    = {Xu, Zheng  and
               Zhang, Yanxiang  and
               Andrew, Galen  and
               Choquette, Christopher  and
               Kairouz, Peter  and
               Mcmahan, Brendan  and
               Rosenstock, Jesse  and
               Zhang, Yuanbo},
  editor    = {Sitaram, Sunayana  and
               Beigman Klebanov, Beata  and
               Williams, Jason D},
  booktitle = {Proceedings of the 61st Annual Meeting of the Association for Computational Linguistics (Volume 5: Industry Track)},
  month     = jul,
  year      = {2023},
  venue     = {ACL},
  tags      = {trustworthiness, differential privacy, privacy preservation, privacy-preserving techniques},
  address   = {Toronto, Canada},
  publisher = {Association for Computational Linguistics},
  url       = {https://aclanthology.org/2023.acl-industry.60},
  doi       = {10.18653/v1/2023.acl-industry.60},
  pages     = {629--639}
}

@inproceedings{sun2024improving,
  title     = {Improving Lo{RA} in Privacy-preserving Federated Learning},
  author    = {Youbang Sun and Zitao Li and Yaliang Li and Bolin Ding},
  booktitle = {The Twelfth International Conference on Learning Representations},
  venue     = {ICLR},
  year      = {2024},
  month     = {may},
  tags      = {efficiency, LoRA, reparameterization-based, trustworthiness, differential privacy, privacy preservation, privacy-preserving techniques},
  url       = {https://openreview.net/forum?id=NLPzL6HWNl}
}

@article{liu2023differentially,
  author    = {Liu, Xiao-Yang and Zhu, Rongyi and Zha, Daochen and Gao, Jiechao and Zhong, Shan and White, Matt and Qiu, Meikang},
  title     = {Differentially Private Low-Rank Adaptation of Large Language Model Using Federated Learning},
  year      = {2024},
  publisher = {Association for Computing Machinery},
  address   = {New York, NY, USA},
  issn      = {2158-656X},
  doi       = {10.1145/3682068},
  journal   = {ACM Trans. Manage. Inf. Syst.},
  month     = aug,
  url       = {https://dl.acm.org/doi/abs/10.1145/3682068},
  venue     = {ACM TMIS},
  tags      = {efficiency, LoRA, reparameterization-based, trustworthiness, differential privacy, privacy preservation, privacy-preserving techniques}
}

@article{babakniya2023revisiting,
  title   = {Revisiting Sparsity Hunting in Federated Learning: Why does Sparsity Consensus Matter?},
  author  = {Sara Babakniya and Souvik Kundu and Saurav Prakash and Yue Niu and Salman Avestimehr},
  journal = {Transactions on Machine Learning Research},
  issn    = {2835-8856},
  year    = {2023},
  url     = {https://openreview.net/forum?id=iHyhdpsnyi},
  note    = {}
}

@article{9069945,
  author   = {Wei, Kang and Li, Jun and Ding, Ming and Ma, Chuan and Yang, Howard H. and Farokhi, Farhad and Jin, Shi and Quek, Tony Q. S. and Vincent Poor, H.},
  journal  = {IEEE Transactions on Information Forensics and Security},
  title    = {Federated Learning With Differential Privacy: Algorithms and Performance Analysis},
  year     = {2020},
  volume   = {15},
  number   = {},
  pages    = {3454-3469},
  doi      = {10.1109/TIFS.2020.2988575}
}

@article{shin2023fedsplitx,
  title         = {FedSplitX: Federated Split Learning for Computationally-Constrained Heterogeneous Clients},
  author        = {Jiyun Shin and Jinhyun Ahn and Honggu Kang and Joonhyuk Kang},
  year          = {2023},
  eprint        = {2310.14579},
  archiveprefix = {arXiv},
  primaryclass  = {cs.LG},
  url           = {https://arxiv.org/abs/2310.14579},
  journal       = {arXiv preprint arXiv:2310.14579}
}

@article{Thapa_MahawagaArachchige_Camtepe_Sun_2022,
  title   = {SplitFed: When Federated Learning Meets Split Learning},
  volume  = {36},
  url     = {https://ojs.aaai.org/index.php/AAAI/article/view/20825},
  doi     = {10.1609/aaai.v36i8.20825},
  number  = {8},
  journal = {Proceedings of the AAAI Conference on Artificial Intelligence},
  author  = {Thapa, Chandra and Mahawaga Arachchige, Pathum Chamikara and Camtepe, Seyit and Sun, Lichao},
  year    = {2022},
  month   = {Jun.},
  pages   = {8485-8493}
}

@article{9660377,
  author  = {Shah, Suhail Mohmad and Lau, Vincent K. N.},
  journal = {IEEE Transactions on Neural Networks and Learning Systems},
  title   = {Model Compression for Communication Efficient Federated Learning},
  year    = {2023},
  volume  = {34},
  number  = {9},
  pages   = {5937-5951},
  doi     = {10.1109/TNNLS.2021.3131614}
}

@inproceedings{reddi2020adaptive,
  title     = {Adaptive Federated Optimization},
  author    = {Sashank J. Reddi and Zachary Charles and Manzil Zaheer and Zachary Garrett and Keith Rush and Jakub Kone{\v{c}}n{\'y} and Sanjiv Kumar and Hugh Brendan McMahan},
  booktitle = {International Conference on Learning Representations},
  year      = {2021},
  url       = {https://openreview.net/forum?id=LkFG3lB13U5}
}

@inproceedings{MLSYS2020_1f5fe839,
  author    = {Li, Tian and Sahu, Anit Kumar and Zaheer, Manzil and Sanjabi, Maziar and Talwalkar, Ameet and Smith, Virginia},
  booktitle = {Proceedings of Machine Learning and Systems},
  editor    = {I. Dhillon and D. Papailiopoulos and V. Sze},
  pages     = {429--450},
  title     = {Federated Optimization in Heterogeneous Networks},
  url       = {https://proceedings.mlsys.org/paper\_files/paper/2020/file/1f5fe83998a09396ebe6477d9475ba0c-Paper.pdf},
  volume    = {2},
  year      = {2020}
}

@inproceedings{acar2021federated,
  title     = {Federated Learning Based on Dynamic Regularization},
  author    = {Durmus Alp Emre Acar and Yue Zhao and Ramon Matas and Matthew Mattina and Paul Whatmough and Venkatesh Saligrama},
  booktitle = {International Conference on Learning Representations},
  year      = {2021},
  url       = {https://openreview.net/forum?id=B7v4QMR6Z9w}
}

@article{wu2023aigenerated,
  title         = {AI-Generated Content (AIGC): A Survey},
  author        = {Jiayang Wu and Wensheng Gan and Zefeng Chen and Shicheng Wan and Hong Lin},
  year          = {2023},
  eprint        = {2304.06632},
  archiveprefix = {arXiv},
  primaryclass  = {cs.AI},
  url           = {https://arxiv.org/abs/2304.06632},
  journal       = {arXiv preprint arXiv:2304.06632}
}

@article{10398474,
  author   = {Xu, Minrui and Du, Hongyang and Niyato, Dusit and Kang, Jiawen and Xiong, Zehui and Mao, Shiwen and Han, Zhu and Jamalipour, Abbas and Kim, Dong In and Shen, Xuemin and Leung, Victor C. M. and Poor, H. Vincent},
  journal  = {IEEE Communications Surveys \& Tutorials},
  title    = {Unleashing the Power of Edge-Cloud Generative AI in Mobile Networks: A Survey of AIGC Services},
  year     = {2024},
  volume   = {},
  number   = {},
  pages    = {1-1},
  doi      = {10.1109/COMST.2024.3353265}
}

@inproceedings{ye2024openfedllm,
  author    = {Ye, Rui and Wang, Wenhao and Chai, Jingyi and Li, Dihan and Li, Zexi and Xu, Yinda and Du, Yaxin and Wang, Yanfeng and Chen, Siheng},
  title     = {OpenFedLLM: Training Large Language Models on Decentralized Private Data via Federated Learning},
  year      = {2024},
  isbn      = {9798400704901},
  publisher = {Association for Computing Machinery},
  address   = {New York, NY, USA},
  url       = {https://doi.org/10.1145/3637528.3671582},
  doi       = {10.1145/3637528.3671582},
  booktitle = {Proceedings of the 30th ACM SIGKDD Conference on Knowledge Discovery and Data Mining},
  pages     = {6137–6147},
  numpages  = {11},
  tags      = {resource, code, library},
  month     = {aug},
  venue     = {KDD}
}

@article{kuang2023federatedscopellm,
  title         = {FederatedScope-LLM: A Comprehensive Package for Fine-tuning Large Language Models in Federated Learning},
  author        = {Weirui Kuang and Bingchen Qian and Zitao Li and Daoyuan Chen and Dawei Gao and Xuchen Pan and Yuexiang Xie and Yaliang Li and Bolin Ding and Jingren Zhou},
  year          = {2023},
  eprint        = {2309.00363},
  archiveprefix = {arXiv},
  primaryclass  = {cs.LG},
  url           = {https://arxiv.org/abs/2309.00363},
  journal       = {arXiv preprint arXiv:2309.00363}
}

@article{fan2023fatellm,
  title         = {FATE-LLM: A Industrial Grade Federated Learning Framework for Large Language Models},
  author        = {Tao Fan and Yan Kang and Guoqiang Ma and Weijing Chen and Wenbin Wei and Lixin Fan and Qiang Yang},
  year          = {2023},
  eprint        = {2310.10049},
  archiveprefix = {arXiv},
  primaryclass  = {cs.LG},
  url           = {https://arxiv.org/abs/2310.10049},
  journal       = {arXiv preprint arXiv:2310.10049}
}

@misc{fedllm,
  title  = {FedLLM},
  author = {FedML},
  year   = {2023},
  url    = {https://blog.fedml.ai/releasing-fedllm-build-your-own-large-language-models-on-proprietary-data-using-the-fedml-platform/}
}

@misc{Alpaca-LoRA,
  title  = {Alpaca-LoRA},
  author = {Eric Wang},
  year   = {2023},
  url    = {https://github.com/tloen/alpaca-lora}
}

@inproceedings{zhao2024llmbased,
  title     = {A Federated Framework for {LLM}-based Recommendation},
  author    = {Zhao, Jujia  and Wang, Wenjie  and Xu, Chen  and Ng, See-Kiong  and Chua, Tat-Seng},
  booktitle = {Findings of the Association for Computational Linguistics: NAACL 2025},
  venue     = {NAACL},
  month     = {apr},
  year      = {2025},
  github    = {https://github.com/Polaris-JZ/FELLRec},
  publisher = {Association for Computational Linguistics},
  url       = {https://aclanthology.org/2025.findings-naacl.155/},
  doi       = {10.18653/v1/2025.findings-naacl.155},
  pages     = {2852--2865},
  tags      = {application, Recommendation Systems},
  isbn      = {979-8-89176-195-7}
}

@article{zhang2024transfr,
  title         = {TransFR: Transferable Federated Recommendation with Pre-trained Language Models},
  author        = {Honglei Zhang and He Liu and Haoxuan Li and Yidong Li},
  year          = {2024},
  eprint        = {2402.01124},
  archiveprefix = {arXiv},
  primaryclass  = {cs.IR},
  url           = {https://arxiv.org/abs/2402.01124},
  journal       = {arXiv preprint arXiv:2402.01124},
  tags          = {application, Recommendation Systems}
}

@article{roth2024empowering,
  title         = {Empowering Federated Learning for Massive Models with NVIDIA FLARE},
  author        = {Holger R. Roth and Ziyue Xu and Yuan-Ting Hsieh and Adithya Renduchintala and Isaac Yang and Zhihong Zhang and Yuhong Wen and Sean Yang and Kevin Lu and Kristopher Kersten and Camir Ricketts and Daguang Xu and Chester Chen and Yan Cheng and Andrew Feng},
  year          = {2024},
  eprint        = {2402.07792},
  archiveprefix = {arXiv},
  primaryclass  = {cs.LG},
  url           = {https://arxiv.org/abs/2402.07792},
  journal       = {arXiv preprint arXiv:2402.07792}
}

@article{yang2023interpretable,
  title         = {Towards Interpretable Mental Health Analysis with Large Language Models},
  author        = {Kailai Yang and Shaoxiong Ji and Tianlin Zhang and Qianqian Xie and Ziyan Kuang and Sophia Ananiadou},
  year          = {2023},
  eprint        = {2304.03347},
  archiveprefix = {arXiv},
  primaryclass  = {cs.CL},
  url           = {https://arxiv.org/abs/2304.03347},
  journal       = {arXiv preprint arXiv:2304.03347}
}

@article{doi:10.1126/sciadv.adg7865,
  author  = {Vijil Chenthamarakshan  and Samuel C. Hoffman  and C. David Owen  and Petra Lukacik  and Claire Strain-Damerell  and Daren Fearon  and Tika R. Malla  and Anthony Tumber  and Christopher J. Schofield  and Helen M.E. Duyvesteyn  and Wanwisa Dejnirattisai  and Loic Carrique  and Thomas S. Walter  and Gavin R. Screaton  and Tetiana Matviiuk  and Aleksandra Mojsilovic  and Jason Crain  and Martin A. Walsh  and David I. Stuart  and Payel Das },
  title   = {Accelerating drug target inhibitor discovery with a deep generative foundation model},
  journal = {Science Advances},
  volume  = {9},
  number  = {25},
  pages   = {eadg7865},
  year    = {2023},
  doi     = {10.1126/sciadv.adg7865},
  url     = {https://www.science.org/doi/abs/10.1126/sciadv.adg7865},
  eprint  = {https://www.science.org/doi/pdf/10.1126/sciadv.adg7865}
}

@article{panagoulias2024evaluating,
  title         = {Evaluating LLM -- Generated Multimodal Diagnosis from Medical Images and Symptom Analysis},
  author        = {Dimitrios P. Panagoulias and Maria Virvou and George A. Tsihrintzis},
  year          = {2024},
  eprint        = {2402.01730},
  archiveprefix = {arXiv},
  primaryclass  = {cs.CL},
  url           = {https://arxiv.org/abs/2402.01730},
  journal       = {arXiv preprint arXiv:2402.01730}
}

@article{chen2023federated,
  title         = {Federated Large Language Model: A Position Paper},
  author        = {Chaochao Chen and Xiaohua Feng and Jun Zhou and Jianwei Yin and Xiaolin Zheng},
  year          = {2023},
  eprint        = {2307.08925},
  archiveprefix = {arXiv},
  primaryclass  = {cs.LG},
  url           = {https://arxiv.org/abs/2307.08925},
  journal       = {arXiv preprint arXiv:2307.08925}
}

@inproceedings{chu2024send,
  author    = {Chu, Yun-Wei and Han, Dong-Jun and Brinton, Christopher G.},
  title     = {Only Send What You Need: Learning to Communicate Efficiently in Federated Multilingual Machine Translation},
  year      = {2024},
  month     = {may},
  isbn      = {9798400701726},
  publisher = {Association for Computing Machinery},
  address   = {New York, NY, USA},
  url       = {https://doi.org/10.1145/3589335.3651931},
  doi       = {10.1145/3589335.3651931},
  booktitle = {Companion Proceedings of the ACM on Web Conference 2024},
  pages     = {1548-1557},
  numpages  = {10},
  venue     = {WWW},
  tags      = {Application, Machine Translation, Multilingualism, efficiency, selective tuning, sparsification}
}

@inproceedings{zhu2024promotingdatamodelprivacy,
  title     = {Promoting Data and Model Privacy in Federated Learning through Quantized {L}o{RA}},
  author    = {JianHao, Zhu  and
               Lv, Changze  and
               Wang, Xiaohua  and
               Wu, Muling  and
               Liu, Wenhao  and
               Li, Tianlong  and
               Ling, Zixuan  and
               Zhang, Cenyuan  and
               Zheng, Xiaoqing  and
               Huang, Xuanjing},
  booktitle = {Findings of the Association for Computational Linguistics: EMNLP 2024},
  year      = {2024},
  month     = nov,
  venue     = {EMNLP},
  tags      = {efficiency, sparsification, LoRA, reparameterization-based, trustworthiness, quantization, privacy preservation, privacy-preserving techniques},
  publisher = {Association for Computational Linguistics},
  url       = {https://aclanthology.org/2024.findings-emnlp.615/},
  doi       = {10.18653/v1/2024.findings-emnlp.615},
  pages     = {10501--10512}
}

@inproceedings{pmlr-v119-rothchild20a,
  title     = {{F}etch{SGD}: Communication-Efficient Federated Learning with Sketching},
  author    = {Rothchild, Daniel and Panda, Ashwinee and Ullah, Enayat and Ivkin, Nikita and Stoica, Ion and Braverman, Vladimir and Gonzalez, Joseph and Arora, Raman},
  booktitle = {Proceedings of the 37th International Conference on Machine Learning},
  pages     = {8253--8265},
  year      = {2020},
  editor    = {III, Hal Daumé and Singh, Aarti},
  volume    = {119},
  series    = {Proceedings of Machine Learning Research},
  publisher = {PMLR},
  venue     = {ICML},
  month     = {jul},
  github    = {https://github.com/kiddyboots216/CommEfficient},
  tags      = {efficiency, sparsification},
  url       = {https://proceedings.mlr.press/v119/rothchild20a.html}
}

@inproceedings{anonymous-naacl24-arr,
  title     = {Generalizable Multilingual Hate Speech Detection on Low Resource Indian Languages using Fair Selection in Federated Learning},
  author    = {Akshay, Singh and Rahul, Thakur},
  booktitle = {Proceedings of the 2024 Conference of the North American Chapter of the Association for Computational Linguistics: Human Language Technologies},
  year      = {2024},
  venue     = {NAACL},
  month     = {jun},
  url       = {https://2024.naacl.org/program/accepted_papers/},
  tags      = {Application, Multilingualism}
}

@inproceedings{NEURIPS2020_92d1e1eb,
  author    = {Baevski, Alexei and Zhou, Yuhao and Mohamed, Abdelrahman and Auli, Michael},
  booktitle = {Advances in Neural Information Processing Systems},
  editor    = {H. Larochelle and M. Ranzato and R. Hadsell and M.F. Balcan and H. Lin},
  pages     = {12449--12460},
  publisher = {Curran Associates, Inc.},
  title     = {wav2vec 2.0: A Framework for Self-Supervised Learning of Speech Representations},
  url       = {https://proceedings.neurips.cc/paper_files/paper/2020/file/92d1e1eb1cd6f9fba3227870bb6d7f07-Paper.pdf},
  volume    = {33},
  year      = {2020}
}

@inproceedings{pmlr-v202-radford23a,
  title     = {Robust Speech Recognition via Large-Scale Weak Supervision},
  author    = {Radford, Alec and Kim, Jong Wook and Xu, Tao and Brockman, Greg and Mcleavey, Christine and Sutskever, Ilya},
  booktitle = {Proceedings of the 40th International Conference on Machine Learning},
  pages     = {28492--28518},
  year      = {2023},
  editor    = {Krause, Andreas and Brunskill, Emma and Cho, Kyunghyun and Engelhardt, Barbara and Sabato, Sivan and Scarlett, Jonathan},
  volume    = {202},
  series    = {Proceedings of Machine Learning Research},
  month     = {23--29 Jul},
  publisher = {PMLR},
  url       = {https://proceedings.mlr.press/v202/radford23a.html}
}

@inproceedings{zeng2024federated,
  author    = {Zeng, Huimin and Yue, Zhenrui and Jiang, Qian and Wang, Dong},
  booktitle = {2024 IEEE International Conference on Big Data (BigData)},
  title     = {Federated Recommendation via Hybrid Retrieval Augmented Generation},
  year      = {2024},
  venue     = {IEEE BigData},
  month     = {dec},
  volume    = {},
  number    = {},
  pages     = {8078-8087},
  github    = {https://github.com/huiminzeng/GPT-FedRec},
  url       = {https://ieeexplore.ieee.org/document/10825302},
  tags      = {application, Recommendation Systems},
  doi       = {10.1109/BigData62323.2024.10825302}
}

@article{BOBADILLA2013109,
  title    = {Recommender systems survey},
  journal  = {Knowledge-Based Systems},
  volume   = {46},
  pages    = {109-132},
  year     = {2013},
  issn     = {0950-7051},
  doi      = {https://doi.org/10.1016/j.knosys.2013.03.012},
  url      = {https://www.sciencedirect.com/science/article/pii/S0950705113001044},
  author   = {J. Bobadilla and F. Ortega and A. Hernando and A. Gutierrez}
}

@article{ammaduddin2019federated,
  title         = {Federated Collaborative Filtering for Privacy-Preserving Personalized Recommendation System},
  author        = {Muhammad Ammad-Ud-Din and Elena Ivannikova and Suleiman A. Khan and Were Oyomno and Qiang Fu and Kuan Eeik Tan and Adrian Flanagan},
  year          = {2019},
  eprint        = {1901.09888},
  archiveprefix = {arXiv},
  primaryclass  = {cs.IR},
  url           = {https://arxiv.org/abs/1901.09888},
  journal       = {arXiv preprint arXiv:1901.09888}
}

@article{10.1145/3578361,
  author     = {Zhang, Honglei and Luo, Fangyuan and Wu, Jun and He, Xiangnan and Li, Yidong},
  title      = {LightFR: Lightweight Federated Recommendation with Privacy-preserving Matrix Factorization},
  year       = {2023},
  issue_date = {October 2023},
  publisher  = {Association for Computing Machinery},
  address    = {New York, NY, USA},
  volume     = {41},
  number     = {4},
  issn       = {1046-8188},
  url        = {https://doi.org/10.1145/3578361},
  doi        = {10.1145/3578361},
  journal    = {ACM Trans. Inf. Syst.},
  month      = {mar},
  articleno  = {90},
  numpages   = {28}
}

@article{gao2023chatrec,
  title         = {Chat-REC: Towards Interactive and Explainable LLMs-Augmented Recommender System},
  author        = {Yunfan Gao and Tao Sheng and Youlin Xiang and Yun Xiong and Haofen Wang and Jiawei Zhang},
  year          = {2023},
  eprint        = {2303.14524},
  archiveprefix = {arXiv},
  primaryclass  = {cs.IR},
  url           = {https://arxiv.org/abs/2303.14524},
  journal       = {arXiv preprint arXiv:2303.14524}
}

@article{wu2023survey,
  title         = {A Survey on Large Language Models for Recommendation},
  author        = {Likang Wu and Zhi Zheng and Zhaopeng Qiu and Hao Wang and Hongchao Gu and Tingjia Shen and Chuan Qin and Chen Zhu and Hengshu Zhu and Qi Liu and Hui Xiong and Enhong Chen},
  year          = {2023},
  eprint        = {2305.19860},
  archiveprefix = {arXiv},
  primaryclass  = {cs.IR},
  url           = {https://arxiv.org/abs/2305.19860},
  journal       = {arXiv preprint arXiv:2305.19860}
}

@article{zhang2023recommendation,
  title         = {Recommendation as Instruction Following: A Large Language Model Empowered Recommendation Approach},
  author        = {Junjie Zhang and Ruobing Xie and Yupeng Hou and Wayne Xin Zhao and Leyu Lin and Ji-Rong Wen},
  year          = {2023},
  eprint        = {2305.07001},
  archiveprefix = {arXiv},
  primaryclass  = {cs.IR},
  url           = {https://arxiv.org/abs/2305.07001},
  journal       = {arXiv preprint arXiv:2305.07001}
}

@inproceedings{yu2023multimodal,
  title     = {Multimodal Federated Learning via Contrastive Representation Ensemble},
  author    = {Qiying Yu and Yang Liu and Yimu Wang and Ke Xu and Jingjing Liu},
  booktitle = {The Eleventh International Conference on Learning Representations },
  year      = {2023},
  url       = {https://openreview.net/forum?id=Hnk1WRMAYqg}
}

@article{liu2019roberta,
  title         = {RoBERTa: A Robustly Optimized BERT Pretraining Approach},
  author        = {Yinhan Liu and Myle Ott and Naman Goyal and Jingfei Du and Mandar Joshi and Danqi Chen and Omer Levy and Mike Lewis and Luke Zettlemoyer and Veselin Stoyanov},
  year          = {2019},
  eprint        = {1907.11692},
  archiveprefix = {arXiv},
  primaryclass  = {cs.CL},
  url           = {https://arxiv.org/abs/1907.11692},
  journal       = {arXiv preprint arXiv:1907.11692}
}

@inproceedings{bai2024federated,
  title     = {Federated Fine-tuning of Large Language Models under Heterogeneous Language Tasks and Client Resources},
  author    = {Jiamu Bai and Daoyuan Chen and Bingchen Qian and Liuyi Yao and Yaliang Li},
  booktitle = {The Thirty-eighth Annual Conference on Neural Information Processing Systems},
  year      = {2024},
  venue     = {NeurIPS},
  month     = sep,
  github    = {https://github.com/alibaba/FederatedScope/tree/FlexLoRA},
  tags      = {efficiency, LoRA, reparameterization-based, heterogeneous resource},
  url       = {https://openreview.net/forum?id=gkOzoHBXUw}
}

@article{wu2020visual,
  title         = {Visual Transformers: Token-based Image Representation and Processing for Computer Vision},
  author        = {Bichen Wu and Chenfeng Xu and Xiaoliang Dai and Alvin Wan and Peizhao Zhang and Zhicheng Yan and Masayoshi Tomizuka and Joseph Gonzalez and Kurt Keutzer and Peter Vajda},
  year          = {2020},
  eprint        = {2006.03677},
  archiveprefix = {arXiv},
  primaryclass  = {cs.CV},
  url           = {https://arxiv.org/abs/2006.03677},
  journal       = {arXiv preprint arXiv:2006.03677}
}

@inproceedings{NEURIPS2021_50525975,
  author    = {Li, Junnan and Selvaraju, Ramprasaath and Gotmare, Akhilesh and Joty, Shafiq and Xiong, Caiming and Hoi, Steven Chu Hong},
  booktitle = {Advances in Neural Information Processing Systems},
  editor    = {M. Ranzato and A. Beygelzimer and Y. Dauphin and P.S. Liang and J. Wortman Vaughan},
  pages     = {9694--9705},
  publisher = {Curran Associates, Inc.},
  title     = {Align before Fuse: Vision and Language Representation Learning with Momentum Distillation},
  url       = {https://proceedings.neurips.cc/paper_files/paper/2021/file/505259756244493872b7709a8a01b536-Paper.pdf},
  volume    = {34},
  year      = {2021}
}

@article{sanh2020distilbert,
  title         = {DistilBERT, a distilled version of BERT: smaller, faster, cheaper and lighter},
  author        = {Victor Sanh and Lysandre Debut and Julien Chaumond and Thomas Wolf},
  year          = {2020},
  eprint        = {1910.01108},
  archiveprefix = {arXiv},
  primaryclass  = {cs.CL},
  url           = {https://arxiv.org/abs/1910.01108},
  journal       = {arXiv preprint arXiv:1910.01108}
}

@article{tang2020multilingual,
  title         = {Multilingual Translation with Extensible Multilingual Pretraining and Finetuning},
  author        = {Yuqing Tang and Chau Tran and Xian Li and Peng-Jen Chen and Naman Goyal and Vishrav Chaudhary and Jiatao Gu and Angela Fan},
  year          = {2020},
  eprint        = {2008.00401},
  archiveprefix = {arXiv},
  primaryclass  = {cs.CL},
  url           = {https://arxiv.org/abs/2008.00401},
  journal       = {arXiv preprint arXiv:2008.00401}
}

@article{radford2019rewon,
  title   = {Rewon child, david luan, dario amodei, and ilya sutskever. 2019},
  author  = {Radford, Alec and Wu, Jeffrey},
  journal = {Language models are unsupervised multitask learners. OpenAI blog},
  url     = {https://d4mucfpksywv.cloudfront.net/better-language-models/language_models_are_unsupervised_multitask_learners.pdf},
  volume  = {1},
  number  = {8},
  pages   = {9},
  year    = {2019}
}

@article{pires2019multilingual,
  title         = {How multilingual is Multilingual BERT?},
  author        = {Telmo Pires and Eva Schlinger and Dan Garrette},
  year          = {2019},
  eprint        = {1906.01502},
  archiveprefix = {arXiv},
  primaryclass  = {cs.CL},
  url           = {https://arxiv.org/abs/1906.01502},
  journal       = {arXiv preprint arXiv:1906.01502}
}

@inproceedings{lester-etal-2021-power,
  title     = {The Power of Scale for Parameter-Efficient Prompt Tuning},
  author    = {Lester, Brian  and
               Al-Rfou, Rami  and
               Constant, Noah},
  booktitle = {Proceedings of the 2021 Conference on Empirical Methods in Natural Language Processing},
  month     = nov,
  year      = {2021},
  address   = {Online and Punta Cana, Dominican Republic},
  publisher = {Association for Computational Linguistics},
  url       = {https://aclanthology.org/2021.emnlp-main.243},
  doi       = {10.18653/v1/2021.emnlp-main.243},
  pages     = {3045--3059}
}

@inproceedings{weller-etal-2022-pretrained,
  title     = {Pretrained Models for Multilingual Federated Learning},
  author    = {Weller, Orion  and
               Marone, Marc  and
               Braverman, Vladimir  and
               Lawrie, Dawn  and
               Van Durme, Benjamin},
  booktitle = {Proceedings of the 2022 Conference of the North American Chapter of the Association for Computational Linguistics: Human Language Technologies},
  month     = jul,
  year      = {2022},
  address   = {Seattle, United States},
  publisher = {Association for Computational Linguistics},
  url       = {https://aclanthology.org/2022.naacl-main.101},
  doi       = {10.18653/v1/2022.naacl-main.101},
  github    = {https://github.com/orionw/Multilingual-Federated-Learning},
  venue     = {NAACL},
  tags      = {Application, Multilingualism},
  pages     = {1413--1421}
}

@inproceedings{devlin-etal-2019-bert,
  title     = {{BERT}: Pre-training of Deep Bidirectional Transformers for Language Understanding},
  author    = {Devlin, Jacob  and
               Chang, Ming-Wei  and
               Lee, Kenton  and
               Toutanova, Kristina},
  booktitle = {Proceedings of the 2019 Conference of the North {A}merican Chapter of the Association for Computational Linguistics: Human Language Technologies, Volume 1 (Long and Short Papers)},
  month     = jun,
  year      = {2019},
  address   = {Minneapolis, Minnesota},
  publisher = {Association for Computational Linguistics},
  url       = {https://aclanthology.org/N19-1423},
  doi       = {10.18653/v1/N19-1423},
  pages     = {4171--4186}
}

@inproceedings{ben-zaken-etal-2022-bitfit,
  title     = {{B}it{F}it: Simple Parameter-efficient Fine-tuning for Transformer-based Masked Language-models},
  author    = {Ben Zaken, Elad  and
               Goldberg, Yoav  and
               Ravfogel, Shauli},
  booktitle = {Proceedings of the 60th Annual Meeting of the Association for Computational Linguistics (Volume 2: Short Papers)},
  month     = may,
  year      = {2022},
  address   = {Dublin, Ireland},
  publisher = {Association for Computational Linguistics},
  url       = {https://aclanthology.org/2022.acl-short.1},
  doi       = {10.18653/v1/2022.acl-short.1},
  pages     = {1--9}
}

@inproceedings{aghajanyan-etal-2021-intrinsic,
  title     = {Intrinsic Dimensionality Explains the Effectiveness of Language Model Fine-Tuning},
  author    = {Aghajanyan, Armen  and
               Gupta, Sonal  and
               Zettlemoyer, Luke},
  booktitle = {Proceedings of the 59th Annual Meeting of the Association for Computational Linguistics and the 11th International Joint Conference on Natural Language Processing (Volume 1: Long Papers)},
  month     = aug,
  year      = {2021},
  address   = {Online},
  publisher = {Association for Computational Linguistics},
  url       = {https://aclanthology.org/2021.acl-long.568},
  doi       = {10.18653/v1/2021.acl-long.568},
  pages     = {7319--7328}
}

@inproceedings{wu-dredze-2020-languages,
  title     = {Are All Languages Created Equal in Multilingual {BERT}?},
  author    = {Wu, Shijie  and
               Dredze, Mark},
  booktitle = {Proceedings of the 5th Workshop on Representation Learning for NLP},
  month     = jul,
  year      = {2020},
  address   = {Online},
  publisher = {Association for Computational Linguistics},
  url       = {https://aclanthology.org/2020.repl4nlp-1.16},
  doi       = {10.18653/v1/2020.repl4nlp-1.16},
  pages     = {120--130}
}

@inproceedings{conneau-etal-2020-unsupervised,
  title     = {Unsupervised Cross-lingual Representation Learning at Scale},
  author    = {Conneau, Alexis  and
               Khandelwal, Kartikay  and
               Goyal, Naman  and
               Chaudhary, Vishrav  and
               Wenzek, Guillaume  and
               Guzm{\'a}n, Francisco  and
               Grave, Edouard  and
               Ott, Myle  and
               Zettlemoyer, Luke  and
               Stoyanov, Veselin},
  booktitle = {Proceedings of the 58th Annual Meeting of the Association for Computational Linguistics},
  month     = jul,
  year      = {2020},
  address   = {Online},
  publisher = {Association for Computational Linguistics},
  url       = {https://aclanthology.org/2020.acl-main.747},
  doi       = {10.18653/v1/2020.acl-main.747},
  pages     = {8440--8451}
}

@inproceedings{pmlr-v209-manoel23a,
  title     = {Federated Multilingual Models for Medical Transcript Analysis},
  author    = {Manoel, Andrea and Garcia, Mirian del Carmen Hipolito and Baumel, Tal and Su, Shize and Chen, Jialei and Sim, Robert and Miller, Dan and Karmon, Danny and Dimitriadis, Dimitrios},
  booktitle = {Proceedings of the Conference on Health, Inference, and Learning},
  pages     = {147--162},
  year      = {2023},
  editor    = {Mortazavi, Bobak J. and Sarker, Tasmie and Beam, Andrew and Ho, Joyce C.},
  volume    = {209},
  series    = {Proceedings of Machine Learning Research},
  month     = {jun},
  publisher = {PMLR},
  url       = {https://proceedings.mlr.press/v209/manoel23a.html},
  tags      = {Application, Multilingualism, Domain, Domain Specific, Healthcare},
  venue     = {CHIL}
}

@article{yang2024dualpersonalizing,
  title         = {Dual-Personalizing Adapter for Federated Foundation Models},
  author        = {Yiyuan Yang and Guodong Long and Tao Shen and Jing Jiang and Michael Blumenstein},
  year          = {2024},
  month         = sep,
  archiveprefix = {arXiv},
  venue         = {NeurIPS},
  primaryclass  = {cs.LG},
  url           = {https://openreview.net/forum?id=nkwPiBSw1f},
  tags          = {efficiency, reparameterization-based, personalization},
  journal       = {arXiv preprint arXiv:2403.19211}
}

@article{10.1162/tacl_a_00065,
  author  = {Johnson, Melvin and Schuster, Mike and Le, Quoc
             V. and Krikun, Maxim and Wu, Yonghui and Chen, Zhifeng and Thorat, Nikhil and Viegas, Fernanda and Wattenberg, Martin and Corrado, Greg and Hughes, Macduff and Dean, Jeffrey},
  title   = {{Google\'s Multilingual Neural Machine Translation System: Enabling
             Zero-Shot Translation}},
  journal = {Transactions of the Association for Computational Linguistics},
  volume  = {5},
  pages   = {339-351},
  year    = {2017},
  month   = {10},
  issn    = {2307-387X},
  doi     = {10.1162/tacl_a_00065},
  url     = {https://doi.org/10.1162/tacl\_a\_00065},
  eprint  = {https://direct.mit.edu/tacl/article-pdf/doi/10.1162/tacl\_a\_00065/1567476/tacl\_a\_00065.pdf}
}

@article{bai2024diprompt,
  title   = {DiPrompT: Disentangled Prompt Tuning for Multiple Latent Domain Generalization in Federated Learning},
  author  = {Sikai Bai and Jie Zhang and Shuaicheng Li and Song Guo and Jingcai Guo and Jun Hou and Tao Han and Xiaocheng Lu},
  year    = {2024},
  venue   = {CVPR},
  month   = {jun},
  url     = {https://arxiv.org/abs/2403.08506},
  journal = {arXiv preprint arXiv:2403.08506},
  tags    = {efficiency, additive tuning, prompt tuning, textual-visual prompt tuning, adaptability, domain-centric adaptation, multi-domain adaptation}
}

@article{liu2024fedfms,
  title         = {FedFMS: Exploring Federated Foundation Models for Medical Image Segmentation},
  author        = {Yuxi Liu and Guibo Luo and Yuesheng Zhu},
  year          = {2024},
  eprint        = {2403.05408},
  archiveprefix = {arXiv},
  venue         = {MICCAI},
  primaryclass  = {eess.IV},
  github        = {https://github.com/LIU-YUXI/FedFMS},
  url           = {https://arxiv.org/abs/2403.05408},
  journal       = {arXiv preprint arXiv:2403.05408},
  tags          = {application, domain specific, healthcare}
}

@article{deng2023unlocking,
  title   = {Unlocking the Potential of Prompt-Tuning in Bridging Generalized and Personalized Federated Learning},
  author  = {Deng, Wenlong and Thrampoulidis, Christos and Li, Xiaoxiao},
  journal = {CVPR},
  pages   = {},
  venue   = {CVPR},
  month   = {jun},
  github  = {https://github.com/ubc-tea/SGPT},
  tags    = {efficiency, additive tuning,  prompt tuning, visual prompt tuning},
  url     = {https://arxiv.org/abs/2310.18285},
  year    = {2024}
}

@article{nguyen2024flora,
  title         = {FLoRA: Enhancing Vision-Language Models with Parameter-Efficient Federated Learning},
  author        = {Duy Phuong Nguyen and J. Pablo Munoz and Ali Jannesari},
  year          = {2024},
  eprint        = {2404.15182},
  archiveprefix = {arXiv},
  tags          = {efficiency, LoRA, reparameterization-based},
  primaryclass  = {cs.LG},
  url           = {https://arxiv.org/abs/2404.15182},
  journal       = {arXiv preprint arXiv:2404.15182}
}

@article{singh2019detailed,
  title         = {Detailed comparison of communication efficiency of split learning and federated learning},
  author        = {Abhishek Singh and Praneeth Vepakomma and Otkrist Gupta and Ramesh Raskar},
  year          = {2019},
  eprint        = {1909.09145},
  archiveprefix = {arXiv},
  primaryclass  = {cs.LG},
  url           = {https://arxiv.org/abs/1909.09145},
  journal       = {arXiv preprint arXiv:1909.09145}
}

@inproceedings{10.24963/ijcai.2023/519,
  author    = {Zheng, Fei and Chen, Chaochao and Lyu, Lingjuan and Yao, Binhui},
  title     = {Reducing communication for split learning by randomized top-k sparsification},
  year      = {2023},
  isbn      = {978-1-956792-03-4},
  url       = {https://doi.org/10.24963/ijcai.2023/519},
  doi       = {10.24963/ijcai.2023/519},
  booktitle = {Proceedings of the Thirty-Second International Joint Conference on Artificial Intelligence},
  articleno = {519},
  numpages  = {9},
  series    = {IJCAI '23},
  venue     = {IJCAI}
}

@article{spall1992multivariate,
  author  = {Spall, J.C.},
  journal = {IEEE Transactions on Automatic Control},
  title   = {Multivariate stochastic approximation using a simultaneous perturbation gradient approximation},
  year    = {1992},
  volume  = {37},
  number  = {3},
  pages   = {332-341},
  url     = {https://ieeexplore.ieee.org/document/119632}
}

@article{9186148,
  author   = {Liu, Sijia and Chen, Pin-Yu and Kailkhura, Bhavya and Zhang, Gaoyuan and Hero III, Alfred O. and Varshney, Pramod K.},
  journal  = {IEEE Signal Processing Magazine},
  title    = {A Primer on Zeroth-Order Optimization in Signal Processing and Machine Learning: Principals, Recent Advances, and Applications},
  year     = {2020},
  volume   = {37},
  number   = {5},
  url      = {https://ieeexplore.ieee.org/abstract/document/9186148},
  pages    = {43-54},
  doi      = {10.1109/MSP.2020.3003837}
}

@article{duchi2015optimal,
  title     = {Optimal rates for zero-order convex optimization: The power of two function evaluations},
  author    = {Duchi, John C and Jordan, Michael I and Wainwright, Martin J and Wibisono, Andre},
  journal   = {IEEE Transactions on Information Theory},
  volume    = {61},
  number    = {5},
  pages     = {2788--2806},
  year      = {2015},
  publisher = {IEEE}
}

@inproceedings{fan-etal-2025-fedcot,
  title     = {{F}ed{C}o{T}: Federated Chain-of-Thought Distillation for Large Language Models},
  author    = {Fan, Tao  and Chen, Weijing  and Kang, Yan  and Ma, Guoqiang  and Gu, Hanlin  and Song, Yuanfeng  and Fan, Lixin  and Yang, Qiang},
  editor    = {Christodoulopoulos, Christos  and Chakraborty, Tanmoy  and Rose, Carolyn  and Peng, Violet},
  booktitle = {Findings of the Association for Computational Linguistics: EMNLP 2025},
  month     = nov,
  year      = {2025},
  tags      = {knowledge distillation},
  address   = {Suzhou, China},
  publisher = {Association for Computational Linguistics},
  url       = {https://aclanthology.org/2025.findings-emnlp.454/},
  venue     = {EMNLP},
  github    = {https://github.com/FederatedAI/FATE-LLM},
  doi       = {10.18653/v1/2025.findings-emnlp.454},
  pages     = {8546--8557},
  isbn      = {979-8-89176-335-7}
}

@article{wu2022communication,
  title     = {Communication-efficient federated learning via knowledge distillation},
  author    = {Wu, Chuhan and Wu, Fangzhao and Lyu, Lingjuan and Huang, Yongfeng and Xie, Xing},
  journal   = {Nature communications},
  venue     = {Nature communications},
  volume    = {13},
  number    = {1},
  pages     = {2032},
  year      = {2022},
  tags      = {knowledge distillation},
  month     = {apr},
  url       = {https://www.nature.com/articles/s41467-022-29763-x},
  github    = {https://github.com/wuch15/FedKD},
  publisher = {Nature Publishing Group UK London}
}

@inproceedings{9892845,
  author    = {Lit, Zhengyang and Sit, Shijing and Wang, Jianzong and Xiao, Jing},
  booktitle = {2022 International Joint Conference on Neural Networks (IJCNN)},
  title     = {Federated Split BERT for Heterogeneous Text Classification},
  year      = {2022},
  volume    = {},
  number    = {},
  pages     = {1-8},
  venue     = {IJCNN},
  doi       = {10.1109/IJCNN55064.2022.9892845},
  url       = {https://ieeexplore.ieee.org/document/9892845},
  tags      = {efficiency, split learning, heterogeneous resource}
}

@article{jere2020taxonomy,
  title     = {A taxonomy of attacks on federated learning},
  author    = {Jere, Malhar S and Farnan, Tyler and Koushanfar, Farinaz},
  journal   = {IEEE Security \& Privacy},
  volume    = {19},
  number    = {2},
  year      = {2020},
  publisher = {IEEE}
}

\end{document}